\documentclass[mono,a4paper]{riverk}

\usepackage{times}

\usepackage[utf8]{inputenc}
\usepackage{rivps}
\usepackage{titling}

 \usepackage{amsmath}

\usepackage{bibentry}

\usepackage[table]{xcolor}
\usepackage{tikz,bm}
\usetikzlibrary{shapes,arrows}
\usetikzlibrary{positioning,fit,calc}
\usetikzlibrary{arrows}
\usetikzlibrary{decorations.pathreplacing}

\usepackage{graphics} 
\usepackage{graphicx} 

\usepackage{caption}
\usepackage{subcaption}

\usepackage{framed}

\usepackage{makeidx}
\usepackage{hyperref}

\usepackage{overture}

\usepackage{hyperref}
\usepackage{url}

\usepackage{paralist}

\usepackage[]{contributions}

\usepackage[prefix={P},publications={Christiansen&12a,Christiansen&12b,Edwards&13,Christiansen&13a,Christiansen&14a,Christiansen&14b,Christiansen&14c,Christiansen&15a,Christiansen&15b}]{citeprefix}

\usepackage{pdfpages}

\usepackage{array}

\usepackage{float}
\graphicspath{{figures/}} 

\usepackage{enumitem}

\usepackage{comment}
\usepackage{todonotes}

\newcounter{myfootnote}

\def\publication#1{
\item[\cite{#1}] \bibentry{#1}
\noindent}

\makeindex

\hypersetup{pdfauthor={Martin Peter Christiansen}}
\hypersetup{pdftitle={Co-modelling of Agricultural Robotic Systems}}
\hypersetup{pdfsubject={Ph.D Dissertation}}
\hypersetup{breaklinks, colorlinks, linkcolor=black, citecolor=black, filecolor=black, urlcolor=black}
\hypersetup{bookmarksopen=true, bookmarksnumbered=true}

\title{Co-modelling of Agricultural Robotic Systems}
\author{Martin Peter Christiansen}
\date{June, 2015}

\nobibliography*
\begin{document}



\pagestyle{empty}
\pagenumbering{alph}


\vspace*{2 mm}

\begin{center}
\linethickness{2mm}
\line(1,0){335}\\
\linethickness{0.8mm}
\line(1,0){335}
\end{center}

\vspace*{1 mm}

\begin{center}
{\noindent    \huge\bfseries
\thetitle}

\end{center}

\begin{center}
\linethickness{0.8mm}
\line(1,0){335}\\
\linethickness{2mm}
\line(1,0){335}
\end{center}

\vspace{10 mm}

\begin{center}
{\noindent    \huge
PhD Dissertation}

\vspace{5 mm}

{\noindent    \huge
\theauthor}

\vspace{5 mm}

{\noindent    \Large
\thedate}
\end{center}

\vspace{10 mm}


\vfill

\begin{flushleft}
  \begin{figure}[!h]
    \includegraphics[width=\textwidth]{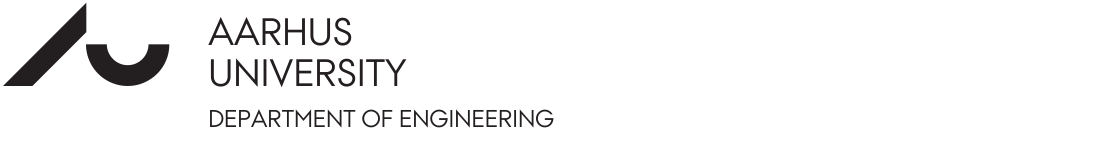}
  \end{figure}
\end{flushleft}


\cleardoublepage

\vspace*{10 mm}

\begin{center}
{\noindent    \huge\thetitle}

\vspace{5 mm}

\begin{flushleft}
  \begin{figure}[!h]
    \centering
    \includegraphics[width=0.5\textwidth]{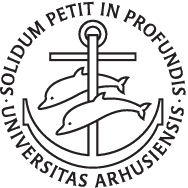}
  \end{figure}
\end{flushleft}

{\noindent    \large
A Dissertation \\
Presented to the Faculty of Science and Technology \\
of the University of Aarhus \\
in Partial Fulfilment of the Requirements for the \\
PhD Degree
}

\vspace{20 mm}

{\noindent    \large
by \\
\theauthor\\
June 17\textsuperscript{th}, 2015
}

\end{center}

\cleardoublepage


\rmfamily
\normalfont

\pagenumbering{roman}
\pagestyle{headings}


\chapter*{Abstract}

Automated and robotic ground-vehicle solutions are gradually becoming part of the agricultural industry, where they are used for performing tasks such as feeding, herding, planting, harvesting, and weed spraying. 
Agricultural machinery operates in both indoor and outdoor farm environments, resulting in changing operational conditions. 
Variation in the load transported by ground-vehicles is a common occurrence in the agricultural domain, in tasks such as animal feeding and field spraying.
The development of automated and robotic ground-vehicle solutions for conditions and scenarios in the agricultural domain is a complex task, which requires input from multiple engineering disciplines.
This PhD thesis proposes modelling and simulation for the research and development of automated and robotic ground-vehicle solutions for purposes such as component development, virtual prototype testing, and scenario evaluation. 
The collaboration of multiple engineering disciplines is achieved by combining multiple modelling and simulation tools from different engineering disciplines.
These combined models are known as co-models and their execution is referred to as co-simulation. 
The results of this thesis are a model-based development methodology for automated and robotic ground-vehicles utilised for a number of research and development cases. 
The co-models of the automated and robotic ground vehicles were created using the model-based development methodology, and they contribute to the future development support  in this research domain.
The thesis presents four contributions toward the exploration of a chosen design space for an automated or robotic ground vehicle. 
Solutions obtained using co-modelling and co-simulation are deployed to their ground-vehicle realisations, which ensures that all stages of development are covered.

\chapter*{Resume}

Robotkøretøjs-løsninger, som har til opgave at forbedre effektiviteten og produktiviteten, er gradvist ved at blive en del landbrugssektoren. 
Disse køretøjer bliver anvendt til driftsopgaver inden for husdyrshold som dyrefodring og hyrdeopgaver samt markarbejde såsom såning, og høst og ukrudtsbekæmpelse. 
Landbrugskøretøjerne har forskellige driftsbetingelser da de opererer både indenfor og udenfor. 
I mange driftsopgaver der udføres af et køretøj i landbrugssektoren, transporteres der en last der ændrer sig undervejs, 
så som dyrefodring hvor der bliver mindre foder efterhånden som der udfodres eller marksprøjtning hvor tank-indholdet løbende fordeles over marken. 
Udvikling af robotkøretøjs-løsninger til forholdene i landbruget er en kompliceret opgave, der kræver samspil imellem flere forskellige ingeniørdiscipliner.

Til forskning og udvikling af robotkøretøjer til landbruget foreslår denne ph.d.-afhandling modellering og simulering, som kan anvendes til komponentudvikling, test af virtuelle prototyper og evaluering af scenarier. 
Til samarbejdet imellem ingeniørdisciplinerne kombineres modellerings- og si\-mu\-le\-rings\-værk\-tøj\-er fra de forskellige discipliner. Disse kombinerede modeller er kendt som co-modeller og deres udførelse co-simulering. 

Resultatet af denne afhandling er en model-baseret udviklingsmetode til robotkøretøjs-løsninger til landbruget og en række co-modellerede forsknings- og udviklingsprojekter inden for dette domæne. 
De co-modellerede projekter er blevet til ved hjælp af den model-baserede udviklingsmetode som bidrager og yder support til effektiv videreudvikling af robotkøretøjer til landbruget. Der præsenteres fire forskellige metode-bidrag til udforskningen af et brugerdefineret designrum for co-modeller af et givet robotkøretøj. Løsninger fundet ved brug af co-modellering og co-simulering er blevet implementeret på de reelle køretøjer, så alle trinene fra koncept til endelig løsning er dækket.

\chapter*{Acknowledgments}

This PhD research project would not have been possible without the financial funding provided
by the Danish Ministry of Food, Agriculture, and Fisheries, which is gratefully acknowledged. 
A great number of people were important to the completion of this dissertation, 
as their support, guidance, and assistance made this research project possible. 
I would like to thank my supervisors, Senior Researcher Ole Green, 
Professor Peter Gorm Larsen, and Senior Researcher Rasmus Nyholm J\o{}rgensen. 
Each of these individuals has acted as my main supervisor at different stages of my PhD research project.
Their different approaches to supervision have had a significant impact on 
both my research and my personal development.
 
I would like to acknowledge current and former colleagues at the Department of Engineering, 
Aarhus University, who provided me with aid, support and knowledge. 
Special thanks are due to Kenneth Lausdahl, Kim Steen, Morten Stigaard Laursen, 
Morten Larsen, Jos\'e Antonio Esparza Isasa, Peter W. V. J\o{}rgensen, and 
Sune Wolf for their camaraderie and discussions on various scientific subjects. 
I also wish to thank S\o{}ren Hansen for introducing me to academic instruction and teaching,
Ole Balling for discussing the topic of vehicle dynamics, 
and Michael N\o{}rremark for discussions on automation in agriculture. 
Thanks are due to all of you for your assistance.
 
For their collaboration on joint work described in this thesis, 
I thank my co-authors: Rasmus Nyholm J\o{}rgensen, 
Peter Gorm Larsen, Ole Green, Gareth Edwards, Morten Larsen, 
Kim Bjerge, Dionysis Bochtis, and Claus Aage Gr\o{}n S\o{}rensen.

I wish to express my gratitude to Christian Kleijn, Paul Weustink, 
Frank Groen, and Marcel  Groothuis from the company, Controllab 
Products B.V., for providing me with the opportunity to visit them in Enschede, 
Netherlands, to gain further insight into the 20-sim modelling technology. 
I also wish to thank Tom Simonsen and Han Knutson from Conpleks Innovation in Struer, Denmark, 
for allowing me to work on the FixiRobo Mink-feeder project in my half-year leave period from my PhD studies.
 
On a more personal level, I would like to thank my girlfriend, Mette Fredsted Gram, 
for all her support, love, encouragement, and understanding throughout my PhD studies. 
She provided continual motivation and encouragement 
for me to challenge myself academically,
along with support
when the research work became too overwhelming. 
Thanks are also due to my sister and brother, Anja Elisabeth S\o{}ndergaard Christiansen 
and Holger Erik Christiansen, for their support and encouragement.

\begin{quotation}
\vspace{1cm}
\hfill \textsl{\theauthor}

\hfill Aarhus, June 2015
\end{quotation}

\chapter*{Abbreviations}
The abbreviations used in this thesis are listed below.
\begin{table}[!ht]
	\normalsize
	\raggedright
	\begin{tabular}{l l l}
	  \textbf{Abbreviation} & \textbf{Full Term}\\
	  ASuBot & Aarhus and Southern Denmark University robot\\
	  ACA & Automated co-model analysis \\
	  CT &  Continuous-time  \\
	  CG & Center of Gravity \\
	  DE & Discrete event  \\
	  DESTECS & Design support and tooling for embedded control software \\
	  DOF & Degrees of freedom \\
	  DSE & Design space exploration  \\
	  EKF & Extended kalman filter \\
	  FMI & Functional mock-up interface \\
	  GNSS & Global navigation satellite system \\
	  GPS & Global positioning system \\
	  ID & Identification data\\
	  IMU & Inertial measurement unit \\
	  ODE & Ordinary differential equation\\
	  RFID & Radio frequency identification \\
	  RSSI & Received signal strength indicator\\
	  ROS & Robot operating system\\
	  RTK-GPS & Real time kinematic GPS\\
	  SDP & Shared design parameter\\
	  VDM & Vienna development method \\
	  QR & Quick response \\
	  XTE & Cross track error\\
	\end{tabular}
\end{table}

\tableofcontents

\mainmatter

\part{Summary}\label{part1}

\chapter{Introduction}

\begin{figure}[th]%
		\centering
		\includegraphics[width=0.98\textwidth]{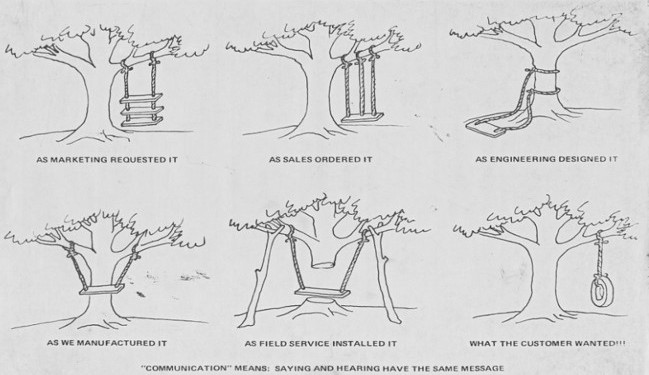}
		\caption[System Design]{1970's tree swing picture concerning communication\footnotemark}
		\label{fig:systemdesign}
	\end{figure}

The development of complex systems regularly involves many project stakeholders from different disciplinary backgrounds.
The project stakeholders are the group of individuals who are actively involved in the project and who may exert influence over the project development, objectives, and outcome.
These stakeholders typically have different points of view on the problem they are addressing, the system being developed, and the process by which it is being developed. 
In this context, a system is a group of interacting or independent components forming a coherent whole.
The development of such a system is, therefore, a highly interactive process involving overlapping problems, 
along with the collaboration of stakeholders designing interrelated components and making coupled decisions~\cite{Geryville&07}.
The well-known tree-swing illustration shown in Figure~\ref{fig:systemdesign} demonstrates the dangers and failures that can be encountered 
if stakeholders do not communicate with each other and their customers when developing a product.

Even different engineering disciplines, such as software, electrical, and mechanical engineering, 
have different perspectives or points of view on a system design. 
Further, the engineering disciplines have developed different concepts and theories to approach product and system development. 
In general, however, an engineer will try to determine the purpose behind a task, to help ensure that the development process is on target. 
Focusing on a single engineering discipline or skill area will not meet future needs for product development. 
Engineering disciplines should be perceived as overlapping and interconnected rather than constituting separate fields of knowledge~\cite{Paul&06}.

Industry demand exists for engineers who can collaborate with project stakeholders outside of their own discipline. 
Development involving multidisciplinary teams are intended to provide innovative new solutions and to improve the multidisciplinary thinking of developers. 
Multidisciplinary collaboration also allows product design to be understood from multiple viewpoints and 
provides team members with the ability to understand the significant design constraints affecting the other disciplines~\cite{Larsen&09c}.
\footnotetext{Source: \url{http://www.businessballs.com/treeswing.htm}. Last accessed: 08-06-2015.}
\subsection*{Mechatronics and Robotics}

Mechatronics is an engineering field that is heavily dependent on skills from multiple disciplines. 
The word mechatronics was coined by Tetsuro Mori in 1969, and is a combination of the words “mechanics” and “electronics”~\cite[Foreword]{Bradley&10}.  
Today, mechatronics combines the areas of control, computer, mechanical and electrical engineering~\cite{Bolton13,kamm96}. 
Mechatronics contains subfields such as automation, consumer products, machine vision, and robotics, where the former and latter are the focus here.
The term “robot” was first used in 1923 in a play entitled R.U.R~\cite{Capek13},\cite[Preface]{Murray&94} by Karel Capek. 
Here, R.U.R. represents Rossum’s Universal Robots. 

\begin{figure}[tb]%
	\centering
	\begin{subfigure}[b]{0.31\textwidth}
                		\includegraphics[width=\textwidth]{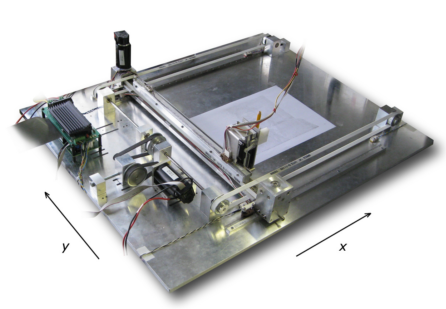}
                		\caption[xyplotter]{Cartesian plotter\footnotemark}
                		\label{fig:yxplotter}
        		\end{subfigure}\hfill
        		\begin{subfigure}[b]{0.31\textwidth}
                		\includegraphics[width=\textwidth]{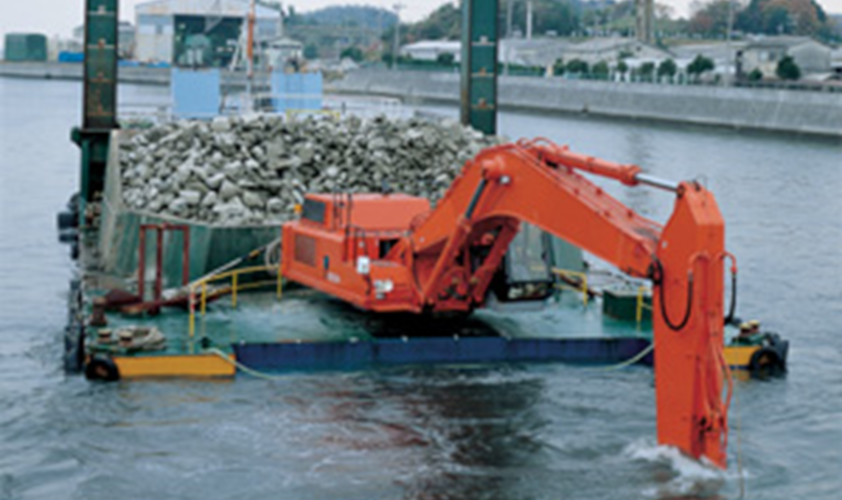}
                		\caption{Dredging Excavator\footnotemark}
                		\label{fig:execavator}
        		\end{subfigure}\hfill
        		\begin{subfigure}[b]{0.27\textwidth}
                		\includegraphics[width=\textwidth]{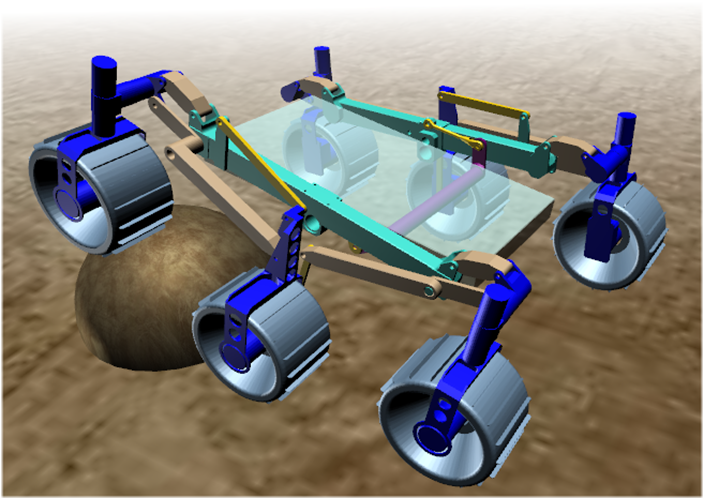}
                		\caption{Planetary Rover\footnotemark}
                		\label{fig:rover}
        		\end{subfigure}
        		\caption{Multidisciplinary automated and robotic engineering system examples from various industry areas: prototyping~(a); 
        		marine conservation~(b); and space exploration~(c).}
	\label{fig:exmultisystems}%
\end{figure}

The areas of automation and robotics tend to overlap, with robotics systems having the ability to perform similar tasks that may differ in terms of objects, distances and other variables.
Automation and robotics engineering projects can be found in most areas of industry and related academic areas. 
The examples in Figure~\ref{fig:exmultisystems} illustrate the broad application of automation and robotics.
They show use in prototype production (Cartesian plotter), material dredging from the seabed, and exploration of the Mars surface.
Robotics systems are diverse in their applications, but all such systems share three similarities, in the form of electronic components, mechanical constructs and software code.

\setcounter{myfootnote}{\value{footnote}}
\addtocounter{myfootnote}{-2}
\footnotetext[\value{myfootnote}]{Source: Picture by Marcel A. Groothuis~\cite{Groothuis&08}.}
\addtocounter{myfootnote}{1}
\footnotetext[\value{myfootnote}]{Source:~\cite{Fitzgerald&14b}.}
\addtocounter{myfootnote}{1}
\footnotetext[\value{myfootnote}]{Source:~\cite{Fitzgerald&14b}.}

To narrow the scope of this project on multidisciplinary automation and robotics development, 
it was decided to focus this thesis on applications in the mobile agricultural ground-vehicle domain. 
The intention is to accommodate the multidisciplinary view when developing an agricultural ground-vehicle system. 
Note that the concepts presented in this PhD thesis may be applicable to other industry areas involving multidisciplinary mechatronics development.

\section{Automated and Robotic Agricultural Ground Vehicles}
\label{sec:agro_intro}

The introduction of robotics and automated precision agriculture machinery provides a means of improving efficiency and productivity.
Research and development to improve machine efficiency and production output has been on-going for several decades~\cite{Kazmi&11,Li&09,OConnor&96,Perez-Ruiz&12,Young&83},
and such development efforts have resulted in various commercial products for industry~\cite{Dantoni&12,Grisson&09}.
The automated and robotic vehicles use localisation systems such as  Global navigation satellite system (GNSS) for automated steering in the environment and perform tasks such as field spraying and animal feeding.
Partially and fully automated systems have been developed for most phases of agricultural operation, from feeding to herding, planting to harvesting, and for packaging and boxing.

\begin{figure}[tb]%
	\centering
	\begin{subfigure}[b]{0.52\textwidth}
                		\includegraphics[width=\textwidth]{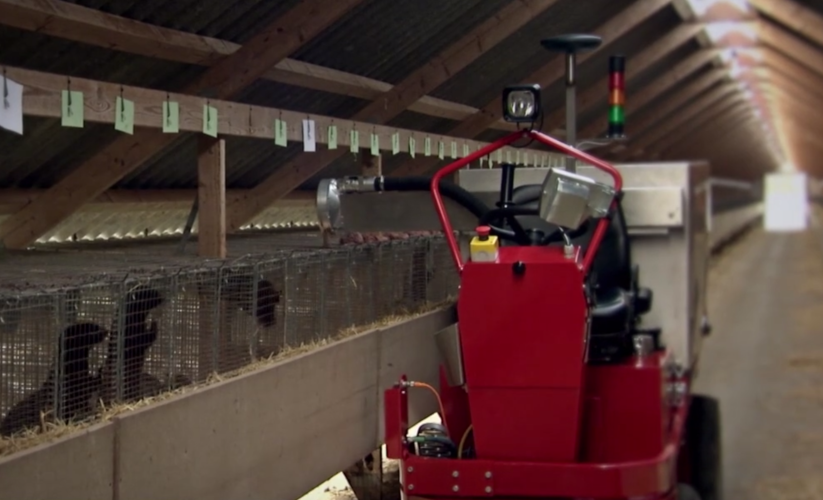}
                		\caption[MinkFeeding]{Robotic mink-feeding system\footnotemark~mounted on a manually operated vehicle.}
                		\label{fig:minkfeedingvehicle}
        		\end{subfigure}\hfill
        		\begin{subfigure}[b]{0.43\textwidth}
                		\includegraphics[width=\textwidth]{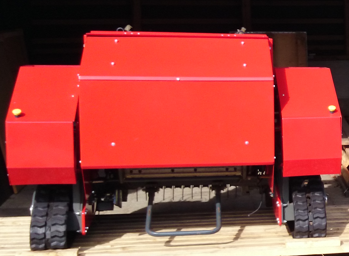}
                		\caption[grasscollect]{Grass-collector robot\footnotemark~from the Grassbots EU FP7 project.}
                		\label{fig:grasscollector}
        		\end{subfigure}
	\caption{Examples of robotics systems in the agricultural industry.}
       \label{fig:example robots}
\end{figure}

Today, one can encounter automated and robotic agricultural systems that are either add-ons to manually operated vehicle systems or fully robotics-focused redesigns. 
Figure~\ref{fig:example robots} illustrates two examples of robotics systems intended for the agricultural industry.
The use of fully robotic platforms is a direct leap towards robotic and autonomous automation, which is intended to provide new approaches to overall design and utilisation.
In contrast, the add-on approach to development of an already operational mobile vehicle aims to provide incremental system improvements by gradually adding automation functionality. 
The concept here is that greater functionality can be obtained by adding control and software intelligence, rather than by improving on well-known mechanical designs. 
For example, agricultural machines such as tractors are used with several various implements for different operational tasks, 
yielding highly modular systems that may require online changes of their control parameters.
Implementing controllers for these modular systems is a complex task, as they are used for a number of specific purposes.

\setcounter{myfootnote}{\value{footnote}}
\addtocounter{myfootnote}{-1}
\footnotetext[\value{myfootnote}]{Source: Picture of a Minkpapir A/S vehicle solution, Conpleks Innovation.}
\addtocounter{myfootnote}{1}
\footnotetext[\value{myfootnote}]{Source: Picture of a Kongskilde Industries A/S vehicle solution, Conpleks Innovation.}

\subsection*{Development and Testing}
Like the tree-swing illustration in Figure~\ref{fig:systemdesign}, different stakeholders can have different points of view
on the automated or robotic agricultural system they are developing.
\begin{figure}[tb]%
	\centering
	\begin{subfigure}[b]{0.25\textwidth}
                		\includegraphics[width=\textwidth]{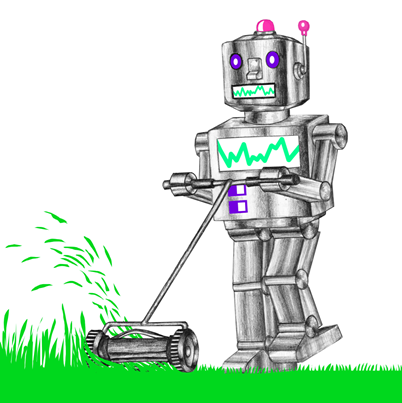}
        		\end{subfigure}\qquad
        		\begin{subfigure}[b]{0.375\textwidth}
        			\includegraphics[width=\textwidth]{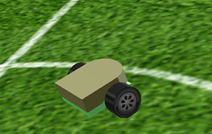}
        		\end{subfigure}
	\caption[grassscut]{Examples of different stakeholder perceptions\footnotemark~of robots for cutting grass.}
       \label{fig:grass_cutters}
\end{figure}
Even the overall envisaged concept of a specific agricultural robotic system can differ between stakeholders in a development team that is communicating internally. 
When a project team is assigned the task of developing a robot for grass cutting, the different members might envision solutions such as those illustrated in Figure~\ref{fig:grass_cutters}.
The above example is an extreme case, but it illustrates some of the problems stakeholders can encounter when they are collaborating on developing a system.
Visual imagery can provide a means of establishing an improved common understanding between different stakeholders in a project, and also provide clarification of the intended direction of the project.

\setcounter{myfootnote}{\value{footnote}}
\footnotetext[\value{myfootnote}]{Source: Robot illustration by Mette Fredsted Gram.}

One of the main obstacles developers must overcome is comparison between different system setups.
Various testing methods for development and evaluation have been proposed in the literature for
sensors~\cite{Cole&04,Gomez-Gil&11} and control system operation~\cite{DSFISODIS12188--2,RoviraMas&08}.
Reproducible scenarios are difficult to obtain in an agricultural setting, because of the semi-controlled outdoor environment,
which is in contrast with the controlled surroundings of the manufacturing industry.
Geographical conditions influence operation and vary in response to weather and terrain conditions~\cite{Eaton&08,Hall&01}.
Prior history from weather and farm operation influences the terrain and soil conditions~\cite{Molari&12,Schjonning&08}.
However, the focus of the developers is to produce the desired machinery response in these semi-controllable conditions.

\begin{figure}[tb]%
	\centering
	\begin{subfigure}[b]{0.45\textwidth}
        		\includegraphics[width=\textwidth]{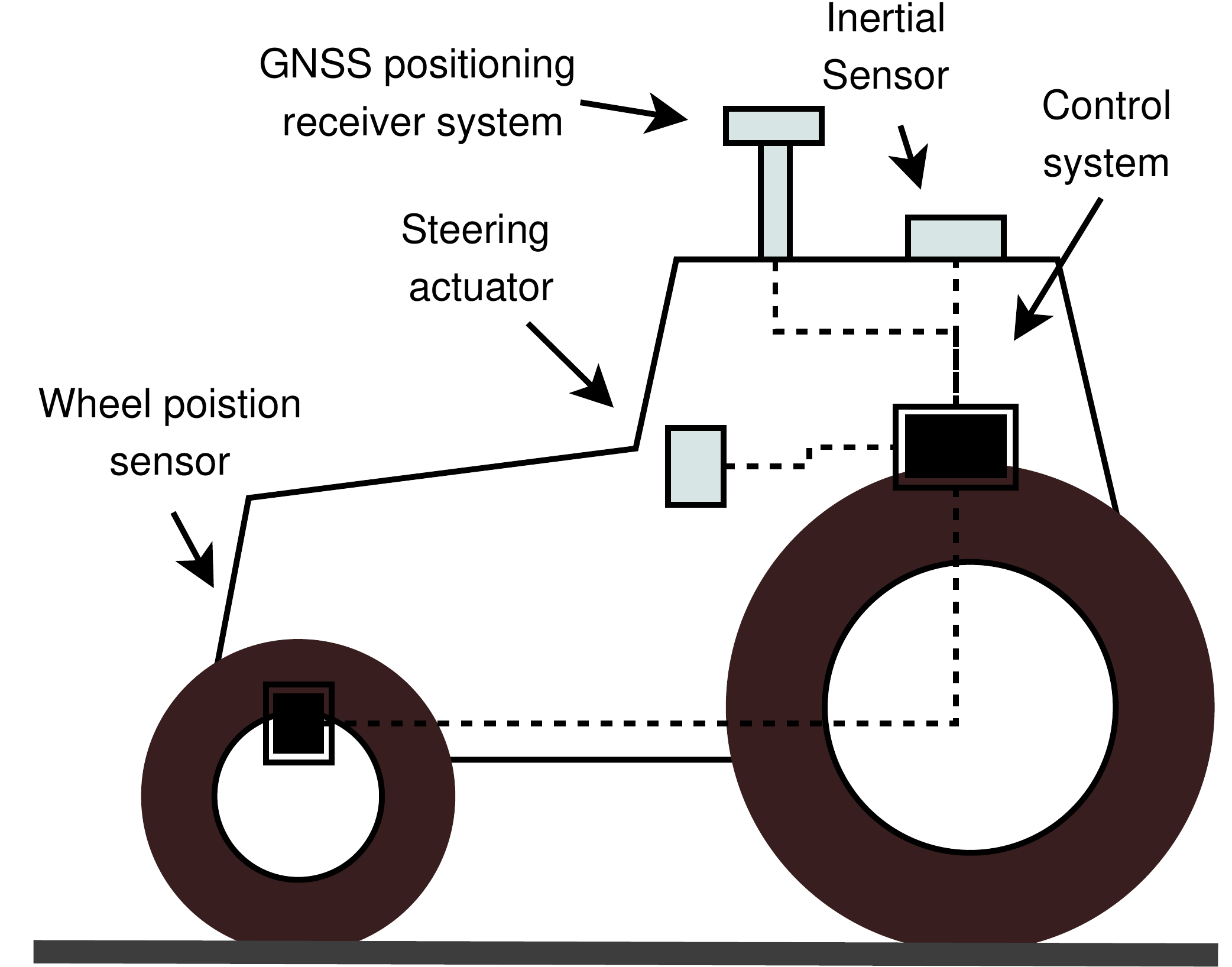}
        		\caption{Sketch of an autonomous tractor with actuator, sensors and control system}
        		\label{fig:auto_tractor_closed_coupled}
      \end{subfigure}
        		\hfill
        		\vline 
        		\vline
        		\hfill
       \begin{subfigure}[b]{0.5\textwidth}
                	\includegraphics[width=\textwidth]{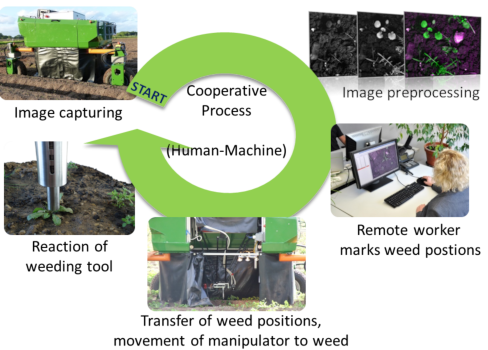}
                	\caption{Example of robotic weed-control development process\footnotemark.}
                	\label{fig:webrobo}
       \end{subfigure}
	\caption{Illustration of two  automated or robotic agricultural ground-vehicle systems, 
			where coupling between the control components differ.}
       \label{fig:agro_system_design}
\end{figure}

Robotic and automated systems can perform operations that are both open- or closed-loop in nature.
Compared to open-loop control, a closed-loop control utilises feedback to compare the actual output to the desired response.
Many control tasks in agriculture are closed-loop in nature, as the input/outputs of the system are tightly coupled; making it a challenge to develop components independently.
An example of  closed-loop control is the GNSS-based steering system for a tractor  illustrated in Figure~\ref{fig:auto_tractor_closed_coupled}, 
where the movement in the environment is managed by the controller, based on sensory feedback.
Evaluating different control strategies for the intended automated or robotic ground vehicle requires field testing to compare their effectiveness.
Because of the semi-controllable outdoor environment testing conditions, the control responses differ between field runs, making direct comparison a challenge.

\footnotetext{Source: Figure by Fabian Sellmann~\cite{Sellmann&14}.}
Robotic weed control in farming is based on sensory input from vision systems~\cite{Ruckelshausen07,Sellmann&14}. 
In Figure~\ref{fig:webrobo}, a remote worker marks the positions of weeds in the captured images and transmits them to the weeding tool.
In this case, a vision sensor is used for detection meaning that the overall process is open-loop. 
The remote worker's marking of weed positions is stored as a base reference and used to evaluate updates to the automated weed-detection algorithms. 
The illustrated robotic weed-control development process allows for decoupled development between the sensory input processing and the actuation system intended to handle the treatment of weeds. 
The system decoupling and remote worker setup allows the system to be developed gradually with an increasing level of automation. 
Having a similar decoupled development process for the control of closed-loop automated or robotic agricultural ground-vehicle systems would allow developers to develop subtasks independently.

\section{Modelling and Simulation}

A model provides the developer with a tool to experiment with system design parameters and configurations. 
The concepts of modelling and simulation are present in one form or another in all engineering disciplines.
Essentially, modelling and simulation are used to represent a specific system with a certain level of fidelity.
They also provide a means of representing the physical world for purposes such as component development, 
prototype testing (virtual prototype), and evaluation of dangerous or cost-demanding scenarios.
Visualisation of simulation results can be a means to address the problem illustrated in Figure~\ref{fig:grass_cutters}.

Engineers and other developers can use modelling and simulation to divide a system into subparts that can be analysed separately and, thus, allow for decoupled development of components. 
This approach to development is similar to the method presented for robot weed-removal development, in which subcomponents are developed independently.
By dividing the system into subcomponents one can achieve a similar development advantage to that illustrated in Figure~\ref{fig:webrobo}.

Modelling and simulation have been utilised in automated and robotic agricultural ground-vehicle development. 
In~\cite{Staranowicz&11}, the authors provide a comparison between commercial and open-source robotic simulation software
and tools ranging from MATLAB/Simulink to the Robot operating system (ROS).
A 2D kinematic model of the tractor and implement was used in~\cite{Backman&12} to test nonlinear model predictive control, 
while a dynamic model of a tractor system was developed in~\cite{Feng&05}, based on measurements from Real time kinematic 
GPS\footnote{Global positioning system (GPS) is one of the satellite system types under the international common GNSS.} (RTK-GPS) and wheel-encoders. 
In~\cite{Karkee&11} the authors developed a model implementing back- and front-wheel cornering stiffness and turned model parameters, based on sample data from RTK-GPS.
In all these cases, a single tool was used to perform the modelling and simulation of a specific project. 
The single tool approach creates a need for project engineers and other developers to collaborate using this specific tool.

\subsection*{Collaborative modelling and co-operative simulation}
Domain-specific modelling software tends to focus on a subset of the engineering disciplines. 
Modelling and simulation across multiple disciplines and domains represent a design challenge in the development of a single tool.
Collaborative modelling (co-modelling) combines separate domain-specific models to create a full model of the intended system, 
by collaboratively exchanging information between the tools. 
Co-modelling allows system components to be developed using different development tools 
and then run simultaneously using co-operative simulation (co-simulation)~\cite{Broenink&10,Nicolescu&07b}.
The exchanged information concerns simulation parameters, control signals, or system events.

The Crescendo co-simulation technology provides a model-based approach to the engineering of embedded and robotic systems~\cite{Fitzgerald&14c}. 
The Crescendo~tool is an open-source tool originally developed in the EU FP7 DESTECS research project.
Crescendo models are built in order to support various forms of analysis, including simulation. 
The Crescendo technology supports models where the controller and plant or environment are modelled using different specialised tools. 
Co-simulation is intended to allow for multidisciplinary modelling with input from domain experts from the different disciplines.

\begin{figure}[!hbt]%
	\centering
	\includegraphics[width=0.8\textwidth]{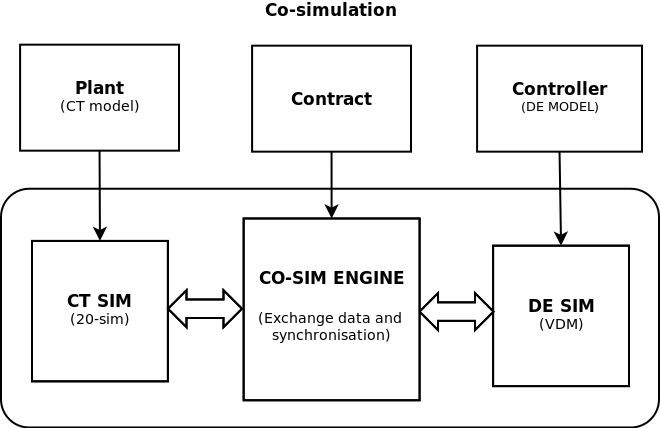}
	\caption{Crescendo co-simulation engine and synchronisation of the CT and DE simulation.}
	\label{fig:crescendo}
\end{figure}

The Crescendo tool uses a combination of discrete-event (DE) modelling of a digital controller and Continuous-time (CT) modelling of the plant/environment for co-simulation. 
The Overture tool~\cite{Jorgensen&13} and Vienna development method (VDM) formalism models the DE controllers, and the 20-sim tool~\cite{Kleijn13,Kleijn06} models the CT components. 
20-sim is a modelling and simulation tool that can model complex multi-domain dynamic systems, such as combined mechanical, electrical, and hydraulic systems. 
VDM Real Time (VDM-RT)~\cite{Verhoef&06} is the VDM dialect used for DE models in Crescendo, and has the capability to describe real-time, asynchronous, object-oriented features. 
VDM and 20-sim are well-established formalisms with stable tool support and a record of industry use.

The Crescendo co-simulation engine coordinates the 20-sim and VDM simulation by implementing a protocol for time-step synchronisation between the tools. 
Crescendo binds the domain models together using the Crescendo contract and is responsible for information exchange between the tools. 
The contract contains the parameters and variables CT and DE developers must be aware of when developing a combined model (a co-model).

The developers can use co-simulation to explore the solution design space, so as to determine viable candidate solutions.
This kind of viable candidate search using co-simulation is known as Design space exploration (DSE).
Crescendo provides a feature called Automated co-model analysis (ACA), with the means to perform automated DSE of a co-model~\cite{Pierce&12}.  
ACA provides the ability to test different system configurations, by running all combinations chosen by the user.
Such alternative system configurations can, for example, involve different actuators, controllers, filters, platforms, and sensor combinations in the design space the developers intend to explore.

The current state-of-art for co-modelling and co-simulation focuses on embedded and robotic system design in general. 
Automated and the robotic ground vehicle has a number of design challenges that is not addressed in the current co-model co-simulations, 
such as localisation, operation in semi-controllable outdoor environments and load transportation.

\section{Motivation}
\label{sec:movtivation}
Co-simulation performed in the agricultural domain has been documented previously in the literature~\cite{Pawlowski&09, Wang&14}.
However, this research field is still in the early stages of development and further research is required to achieve a systematic development methodology. 
Such a methodology would allow developers to move from project start-up through co-modelling to finish with a product that can be deployed on a system realisation. 

Model development is normally constrained by resources such as time and money.
The developers’ primary goal is to achieve a system model that is viable for controller development.  
One should remember that modelling is not an attempt to replicate the full reality into a model, 
but rather an attempt to focus on the parts relevant to the developers’ current case~\cite{Eykhoff74}.
To support development using co-modelling and co-simulation, 
there is a need for general modelling building blocks that can be reused to kickstart development projects.

The ability to divide the development of new automated or robotics systems into subtasks that can be developed independently will increase efficiency,
by allowing independent subtasks to be completed in parallel by a development team. 
Here, co-modelling and co-simulation can facilitate development of different parts of the system independently 
and allow developers from different disciplines to collaborate.

Using co-modelling and co-simulation for the development of automated and robotic agricultural ground-vehicle systems 
can allow developers to address the problem of comparing different system setups that are intended to operate in a semi-controllable environment. 
Ground vehicles are dependent on solutions for variable sensory conditions, such as GNSS and visual input based on landmarks,
when the ground vehicle is automatically moving between indoor and outdoor operational conditions on a farm. 
Scenarios in which the load transported by the ground vehicles varies also occurs in agricultural operational task, such as animal feeding and field spraying.
Addressing changing operational conditions for automated and robotic agricultural ground vehicles is an ongoing area of research that has been been fully developed.
Here co-simulation has the capacity to allow developers to explore these changing operational conditions and to compare alternative solutions.

\section{Research Objectives}
 \label{sec:research_objective}

This PhD project deals with the problems encountered during the design and deployment of an automated or robotic mobile ground vehicle in the agricultural domain.
The overall aim of the project is to have co-models and a methodology that support the design and deployment of robotic and automated agricultural vehicles. 
The focus is on co-models, combining DE models of the control elements with CT modelling of the physical elements and the surrounding environment. 
We believe that better system configurations can be selected by utilising a co-simulated model of an automated or robotic agricultural ground vehicle.
The intended approach is to model the significant factors influencing a specific scenario. 
For instance, steering performance has a higher impact than motor vibrations on the vehicle ground movement. 
Part of the intended development process is to determine the influential factors for a given model.
The project hypothesis can be divided into the following two statements:

\begin{itemize}
	\item \textbf{Collaborative models can support multidisciplinary collaboration and system development.}\\
				\\
				Model-based development can support collaboration between different engineering disciplines throughout the development process. 
				A model can provide insights into the multidisciplinary development design of an automated or robotic agricultural ground vehicle.
				Collaborative models of different vehicle solutions can also be used to understand the controller interactions with the vehicle and its dynamics.
				The candidate solutions found using co-modelling and co-simulation can be deployed in the system realisation. 

	\item \textbf{A collaborative model of a robotic or automated agricultural ground-vehicle can be utilised to explore alternative design configurations.}\\
				\\
				Co-simulation can be used to test and evaluate developer-defined virtual prototype solutions of robotic or automated agricultural ground-vehicles. 
				The design space can be rather large and real-world testing can be a costly and time-consuming task. 
				The goal of prototype testing in co-simulation mode is to diminish the amount of prototype solutions that require testing in the real world. 
				Developers can also use the co-model to obtain an overview of a design space they have defined, to allow them to select viable candidate solutions.
\end{itemize}
The two components of the hypothesis should not be regarded as separate entities, but rather as different aspects of the PhD research objectives.
The hypothesis division is intended to allow for improved evaluation of the contributions of this PhD thesis, based on the evaluation criteria described in section~\ref{sec:evalcrit} below. 

\section{Research Methods}

To facilitate understanding of the approach and results of this PhD project, a description of the applied  research methods is given in this section.

The approach adopted in this PhD project focuses on co-model design and the application of co-simulation in the development of robotic and automated agricultural vehicles. 
A robotics engineering perspective on the development of co-models for the agricultural domain is assumed. 
This research focuses on designing co-models using input from a combination of external case studies, 
domain specialists, literature surveys, and collaboration with companies in the agricultural industry. 
Additional input is gained through observations and analysis of academic and industrial case studies in the DESTECS project.

In the early research stages, time was spent on identifying relevant academic and industrial development cases that were deemed to benefit from co-modelling and co-simulation. 
The selected development cases are used as case studies in this PhD project and should be regarded as a significant part of this study's contribution to the wider field.
For each case study, the existing literature for multi disciplines is consulted to identify similar problems with related solutions. 
Model elements found in different domains are evaluated and coupled into a working co-model. 
Models from the domains of mechanical vehicle modelling, software structuring, control theory, and optimisation are adapted and extended to fit the demands of the project. 
The experience obtained from these case studies is collected to derive a methodology for the development of co-models for automated and robotic agricultural ground vehicles.

\section{Evaluation Criteria}
\label{sec:evalcrit}

The individual contributions made by this PhD project are numbered sequentially in Chapters~\ref{chap:agro}~and~\ref{chap:dse}. 
There is a total of 9 contributions, each of which is evaluated using the evaluation criteria described below:

\begin{description}
	\item[Multi-disciplinary collaboration support] Improved support for intercommunication between different types of developers and other stakeholders in a project concerning agricultural robotic or automated vehicle development. 
	This evaluation criterion focuses on contributions that have value for project collaboration or that allow stakeholders the ability to grasp concepts that are not inherent in their own disciplines.
	
	\item[Model Deployment] Evaluation of the ability to deploy a component from a co-modelling scenario to a system realisation.
		The intent is to verify that a solution is applicable in an actual system setting.
		
	\item[Determination of candidate solutions] The use of co-simulation to determine prototype and parameter solutions automatically in both the DE and CT domains.
	
	\item[Support for modelling of different vehicle solutions] Different types of vehicle solution exist in the agricultural industry; 
	the ability to model these different types will aid in expanding the co-modelling base.
	
	\item[Virtual prototype development support] The ability to support system development using virtual prototypes based on co-modelling and co-simulation. 
		The virtual prototypes are used to evaluate the impact of changes and to test new system designs.
\end{description}

The evaluations of the research contributions are described in \autoref{sec:evaluation}, 
where the level of fulfilment of the individual criteria is assessed. 
\begin{figure}[!hpt]
%

\usetikzlibrary{shapes}
\usetikzlibrary{positioning,fit,calc}

\ifdefined\cref\else\def\cref#1{[C1]}\fi
\newcommand{\D}{9} 
\newcommand{\U}{4} 

\newdimen\R 
\R=3.5cm 
\newdimen\L 
\L=4cm

\newcommand{\A}{-360/\D} 
\newcommand{\AO}{\A + 120} 
 
\def\labelformat#1{\tiny #1}

\def\makeweb{
\tikzset{spiderwebpath/.style={opacity=0.2}}
\tikzset{measurepath/.style={ultra thick,opacity=0.5,rounded corners=1pt}}
\tikzset{mcol1/.style={color={rgb:red,237;green,45;blue,46}}}
\tikzset{mcol2/.style={color={rgb:red,0;green,140;blue,71}}}
\tikzset{mcol3/.style={color={rgb:red,24;green,89;blue,169}}}
\tikzset{mcol4/.style={color={rgb:red,102;green,44;blue,145}}}
\tikzset{mcol5/.style={color=olive}}
  \path (0:0cm) coordinate (O); 

  \foreach \X in {1,...,\D}{
    \draw[spiderwebpath] (\X*\A:0) -- (\X*\A:\R);
  }

  \foreach \Y in {0,...,\U}{
    \foreach \X in {1,...,\D}{
      \path (\X*\AO :\Y*\R/\U) coordinate (D\X-\Y);
      \fill[spiderwebpath] (D\X-\Y) circle (4pt);
    }

    \draw [spiderwebpath] (0:\Y*\R/\U) \foreach \X in {1,...,\D}{
        -- (\X*\A :\Y*\R/\U)
    } -- cycle;
  }
     \fill[white] (0,0) circle (4.03pt);
     \fill[spiderwebpath] (0,0) circle (4pt);

  
  \path (1*\AO :\L) node (L1) {\labelformat{ \cref{contribution:methodology}}};
  \path (2*\AO :\L) node (L2) {\labelformat{ \cref{contribution:legosim}}};
  \path (3*\AO :\L) node (L3) {\labelformat{ \cref{contribution:legodeploy}}};
  \path (4*\AO :\L) node (L4) {\labelformat{ \cref{contribution:modelcase2}}};
  \path (5*\AO :\L) node (L5) {\labelformat{ \cref{contribution:feeding}}};
  \path (6*\AO :\L) node (L8) {\labelformat{ \cref{contribution:visualisation}}};
  \path (7*\AO :\L) node (L6) {\labelformat{ \cref{contribution:patent}}};
  \path (8*\AO :\L) node (L9) {\labelformat{ \cref{contribution:DSEvehicle}}};
  \path (9*\AO :\L) node (L7) {\labelformat{ \cref{contribution:DSEANA}}};
}

   \def\labelformat#1{\large #1}
   \def\subfixspiderscale{0.4}

\def\makeContributionCompareWeb{
\begin{subfigure}[t]{0.33\textwidth}
\centering
\begin{tikzpicture}[scale=\subfixspiderscale, transform shape]
\makeweb
  \fill [mcol1,measurepath]
    (D1-4) -- 
    (D2-2) --
    (D3-2) --
    (D4-3) --
    (D5-2) --
    (D6-2) --
    (D7-0) -- 
    (D8-2) -- 
    (D9-2) --  
   cycle;
\end{tikzpicture}
\caption{\scriptsize Multi-disciplinary collaboration support}
\label{fig:multidisciplinary}
\end{subfigure}~
\begin{subfigure}[t]{0.33\textwidth}
\centering
\begin{tikzpicture}[scale=\subfixspiderscale, transform shape]
\makeweb
  \fill [mcol3,measurepath]
    (D1-3) --
    (D2-3) --
    (D3-1) --
    (D4-4) --
    (D5-4) -- 
    (D6-0) --
    (D7-0) -- 
    (D8-0) -- 
    (D9-0) -- 
   cycle;
\end{tikzpicture}
\caption{\scriptsize Support for modelling of different vehicle solutions}
\label{fig:contrib:modelling}
\end{subfigure}
\begin{subfigure}[t]{0.3\textwidth}
\centering
\begin{tikzpicture}[scale=\subfixspiderscale, transform shape]
\makeweb
  \fill [mcol2,measurepath]
    (D1-2) --
    (D2-3) --
    (D3-4) --
    (D4-3) --
    (D5-0) --
    (D6-4) --
    (D7-1) -- 
    (D8-2) -- 
    (D9-0) -- 
    cycle;
\end{tikzpicture}
\caption{\scriptsize Model deployment}
\label{fig:deployment}
\end{subfigure}
\begin{subfigure}[t]{0.3\textwidth}
\centering
\begin{tikzpicture}[scale=\subfixspiderscale, transform shape]
\makeweb
  \fill [mcol4,measurepath]
    (D1-2) --
    (D2-2) --
    (D3-1) --
    (D4-4) --
    (D5-3) -- 
    (D6-4) --
    (D7-2) -- 
    (D8-3) -- 
    (D9-2) -- 
   cycle;
\end{tikzpicture}
\caption{\scriptsize Virtual prototype development support}
\label{fig:contrib:support}
\end{subfigure}
\begin{subfigure}[t]{0.33\textwidth}
\centering
\begin{tikzpicture}[scale=\subfixspiderscale, transform shape]
\makeweb
  \fill [mcol5,measurepath]
    (D1-0) --
    (D2-0) --
    (D3-0) --
    (D4-0) --
    (D5-0) --
    (D6-4) --
    (D7-4) -- 
    (D8-2) -- 
    (D9-3) -- 
   cycle;
\end{tikzpicture}
\caption{\scriptsize Determination of candidate solutions}
\label{fig:contrib:solution}
\end{subfigure}
%
%
%
\begin{subfigure}[t]{0.33\textwidth}
\centering
\begin{tikzpicture}[scale=\subfixspiderscale, transform shape]
\makeweb
 \fill [mcol1,measurepath]
    (D1-4) -- 
    (D2-2) --
    (D3-2) --
    (D4-3) --
    (D5-2) --
    (D6-2) --
    (D7-0) -- 
    (D8-2) -- 
    (D9-2) --  
   cycle;
  \fill [mcol3,measurepath]
    (D1-3) --
    (D2-3) --
    (D3-1) --
    (D4-4) --
    (D5-4) -- 
    (D6-0) --
    (D7-0) -- 
    (D8-0) -- 
    (D9-0) -- 
   cycle;
  \fill [mcol2,measurepath]
    (D1-2) --
    (D2-3) --
    (D3-4) --
    (D4-3) --
    (D5-0) --
    (D6-4) --
    (D7-1) -- 
    (D8-2) -- 
    (D9-0) -- 
    cycle;
  \fill [mcol4,measurepath]
    (D1-2) --
    (D2-2) --
    (D3-1) --
    (D4-4) --
    (D5-3) -- 
    (D6-4) --
    (D7-2) -- 
    (D8-3) -- 
    (D9-2) -- 
   cycle;
  \fill [mcol5,measurepath]
    (D1-0) --
    (D2-0) --
    (D3-0) --
    (D4-0) --
    (D5-0) --
    (D6-4) --
    (D7-4) -- 
    (D8-2) -- 
    (D9-3) -- 
   cycle;
\end{tikzpicture}
\caption{\scriptsize Combined Overview}
\label{fig:contrib:combined-overview}
\end{subfigure}
}

\def\makeHypothesisCompareWeb{
\begin{subfigure}[t]{0.49\textwidth}
\centering
\begin{tikzpicture}[scale=0.6, transform shape]
\makeweb
  \fill [mcol1,measurepath]
    (D1-4) -- 
    (D2-2) --
    (D3-2) --
    (D4-3) --
    (D5-2) --
    (D6-2) --
    (D7-0) -- 
    (D8-2) -- 
    (D9-2) --  
   cycle;
  \fill [mcol3,measurepath]
    (D1-3) --
    (D2-3) --
    (D3-1) --
    (D4-4) --
    (D5-4) -- 
    (D6-0) --
    (D7-0) -- 
    (D8-0) -- 
    (D9-0) -- 
   cycle;
  \fill [mcol2,measurepath]
    (D1-2) --
    (D2-3) --
    (D3-4) --
    (D4-3) --
    (D5-0) --
    (D6-4) --
    (D7-1) -- 
    (D8-2) -- 
    (D9-0) -- 
    cycle;
\end{tikzpicture}
\caption{\scriptsize Hypothesis - part 1}
\label{fig:hypo1_eval}
\end{subfigure}
\begin{subfigure}[t]{0.49\textwidth}
\centering
\begin{tikzpicture}[scale=0.6, transform shape]
\makeweb
    \fill [mcol2,measurepath]
    (D1-2) --
    (D2-3) --
    (D3-4) --
    (D4-3) --
    (D5-0) --
    (D6-4) --
    (D7-1) -- 
    (D8-2) -- 
    (D9-0) -- 
    cycle;
  \fill [mcol4,measurepath]
    (D1-2) --
    (D2-2) --
    (D3-1) --
    (D4-4) --
    (D5-3) -- 
    (D6-4) --
    (D7-2) -- 
    (D8-3) -- 
    (D9-2) -- 
   cycle;
  \fill [mcol5,measurepath]
    (D1-0) --
    (D2-0) --
    (D3-0) --
    (D4-0) --
    (D5-0) --
    (D6-4) --
    (D7-4) -- 
    (D8-2) -- 
    (D9-3) -- 
   cycle;
\end{tikzpicture}
\caption{\scriptsize Hypothesis - part 2}
\label{fig:hypo2_eval}
\end{subfigure}
}
	\centering
	\begin{tikzpicture}[scale=1.2*\subfixspiderscale, transform shape]
		\makeweb
		  \fill [mcol1,measurepath]
		    (D1-0) -- 
		    (D2-4) --
		    (D3-3) --
		    (D4-2) --
		    (D5-3) --
		    (D6-3) --
		    (D7-2) -- 
		    (D8-4) -- 
		    (D9-2) -- 
		   cycle;
	\end{tikzpicture}
	\caption{Sample comparison chart.}
	\label{fig:webexample}
\end{figure}
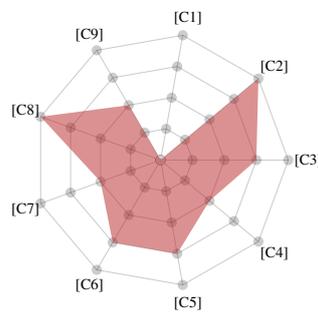
The extent to which the criteria are fulfilled is illustrated using a chart, as exemplified in \autoref{fig:webexample}, 
which shows the research contributions and the fulfilment of the criteria. 
We use the natural number range 0--4 as the value set used to judge the contribution relevance for each evaluation criterion.
Thus, the value “2” should be seen as 50\% of the maximum evaluation value. 

\section{Academic Work}
	This section presents the academic work produced during this PhD project, 
	which primarily focuses on the topics of co-simulation and DSE of automated agricultural vehicles.
	
	\subsection{Publications}
	The publications listed here are all included in Part~\ref{part2} of this thesis.
		\begin{itemize}
			\publication{Christiansen&12a}
			\publication{Christiansen&13a}
			\publication{Edwards&13}
		\end{itemize}
		
	\subsection{Submitted work}
		\begin{itemize}
			\publication{Christiansen&14c}
			Patent application, DP 14870, Filing number:~PA 2014 70803
			\publication{Christiansen&15a}
			\publication{Christiansen&15b}
		\end{itemize}
		
	\subsection{Other publications}
		The publications listed here have not been selected for inclusion in this thesis but are all available from their publishers.
		\begin{itemize}
			\publication{Christiansen&12b}
			\publication{Christiansen&14a}
			\publication{Christiansen&14b}
		\end{itemize}
			
\section{Outline and Reading Guide}
\def\mytotalpapers{six}
The dissertation is divided into two parts: Part~\ref{part1} contains an introduction and provides a summary of the research performed during the PhD project. 
This part also provides an overview of the research contributions on the basis of the publications that were produced during the project. 
Part~\ref{part2} contains a subset of publications by the author with co-authors, on which the research contributions are based. 

In addition to an introduction to the research field, Part~\ref{part1} provides an overview of the research performed, on the basis of the abovementioned publications.
For clear identification, contributions are numbered, e.g.,~\cref{contribution:methodology}, and framed: 
\begin{framed} Contribution 1: Description \end{framed}

The purpose of Part~\ref{part1} is to provide an overview of the publications produced during this project and their contributions to this topic, 
while also introducing relevant background material and related work. 
It was decided to rewrite formulas differently to their representation in the papers included in Part~\ref{part2}, to provided consistent parameter terminology throughout Part~\ref{part1}.
Part I introduces a total of \mytotalpapers\/  publications, three of which have been published and three of which have been submitted.
To allow these publications to be distinguished from other references, they are prefixed with ``\getciteprefix'',~e.g.,~\cite{Christiansen&12a}.

Part~\ref{part1} is structured as follows: 
Chapter 1 contains a short introduction to the PhD thesis. 
Chapter 2 presents the agricultural automated and robotic co-modelling cases and the developed co-modelling methodology,
while Chapter 3 introduces work regarding DSE.
Chapter 4 concludes Part~\ref{part1} of this thesis by summarising the work produced within the PhD project and discusses the contributions made. 
The contributions are  evaluated based on the set criteria and compared to similar or related work.
The conclusion also contains a discussion of how the contributions meet the research aim and hypothesis, and outlines possible future work.

Part II presents a selection of scientific papers written by the author of this PhD thesis in collaboration with others. 
Each chapter presents a publication and starts by listing the bibliography entry for the publication.
This is followed by the publication in its original form.

\chapter{Modelling and Simulation of Automated and Robotic Agricultural Vehicles}
\label{chap:agro}
	\vspace{-20pt}
	
	This chapter presents the research and resultant contributions regarding the co-model based development of 
	automated and robotic agricultural ground-vehicles and the derived extended agricultural development methodology. 
	The extended methodology is dependent on co-simulation to evaluate the co-models and to allow for model deployment. 
	This chapter begins with a literature survey of the state-of-art in co-modelling development methodologies.
	The chapter then moves on to the extended agricultural development methodology, derived from the development of co-models for agricultural ground vehicles. 
	The extended agricultural development methodology is derived based on key case studies co-modelled in this PhD project, which we present in this chapter. 
	We conclude the chapter with an overview of the projects in which the co-modelling technology has been applied and deployed to the actual ground-vehicle realisations.
	
	The publications \cite{Christiansen&12a,Christiansen&13a,Christiansen&15b,Christiansen&15a, Edwards&13} are related to this chapter. 
	A total of five contributions are derived from these publications, which are framed in each section discussing a specific publication.
	\section{Approaches to Co-modelling and Co-simulation}
	\label{sec:lit_model}

	The approach to mechatronic system design using modelling and co-simulation is intended to improve cross-discipline design dialogue and to reduce development cost and time. 
	This modelling approach provides designers with the ability to explore candidate solutions virtually using co-simulation from a possible candidate design space.
	In indoor industrial robotics design applications, research indicates that a model-driven co-design approach incorporating co-simulation improves cross-disciplinary design dialogue~\cite{Broenink&12,Broenink&10c}.
	For example, co-simulation has been used in the development of several robotics manipulator systems, where no development methodology seems to be defined and the focuses are on the structure of the specific model. 
	All of these models use a combination of Matlab Simulink and Adams tools, where the Simulink model is the controller and ADAMS operates the robot body and environment dynamics~\cite{Brezina&11,Cheraghpour&11,Tian&06}. 
	A tomato-harvesting robot manipulator using the ADAMS/Simulink combination and with a similar development focus has also been designed~\cite{Jun&12}.
	
	In the CODIS framework, a design methodology has been used in which the gradual definition of the 
	simulation interface functionality between continuous/discrete model components drove the development.  
	A single case study of an optical network on a chip was used to verify the model components~\cite{Gheorge09}.
	Based on published literature, the MODELISAR project focus was on developing the Functional Mock-up Interface (FMI)
	and defining a tool independent standard for model exchange and co-simulation~\cite{Blochwitz&12, Blochwitz&11}.
	Later, a methodology for use of FMI for development in the automotive industry was published, which was related to the product life-cycle~\cite{LaLande14}. 
	In the DESTECS project, the design approach could be DE-first, CT-first, or contract-first. 
	This flexibility allowed the developers to either begin modelling the DE or CT side or to begin defining the co-model based on what should be shared knowledge between the disciplines.

	These observations clearly indicate that, in order to address design challenges, a design methodology for the development of automated and robotic agricultural ground vehicles should be capable of:	
	\begin{itemize}
		\item \textit{Ensuring a methodical movement from idea/problem to actual implementation of the automated or robotics ground-vehicle. 
					Co-modelling should be a stepping-stone to bridge the gap between project initialisation and the final product.}
		\item \textit{Providing the project stakeholders with guidelines to aid in the selection of the relevant aspects of the project that is being co-modelled. 
					Providing the development team with improved capabilities to address the various design and implementation 
					challenges in the multiple subsystems contained within an automated or robotic  agricultural ground-vehicle system.}
		\item \textit{Providing engineers with varying degrees of experience with the agriculture domain with an
					understanding of development cases and permitting inter-disciplinary collaboration with project stakeholders using co-modelling.
					The intention behind this approach is to allow the project stakeholders to obtain a combined overall project view and to gain confidence in the system design.}
	\end{itemize}
	
	\section{Extended Development Methodology}
	\label{sec:model_methodology}
\begin{figure}[bth]
		  \centering
		  \includegraphics[width=0.98\textwidth]{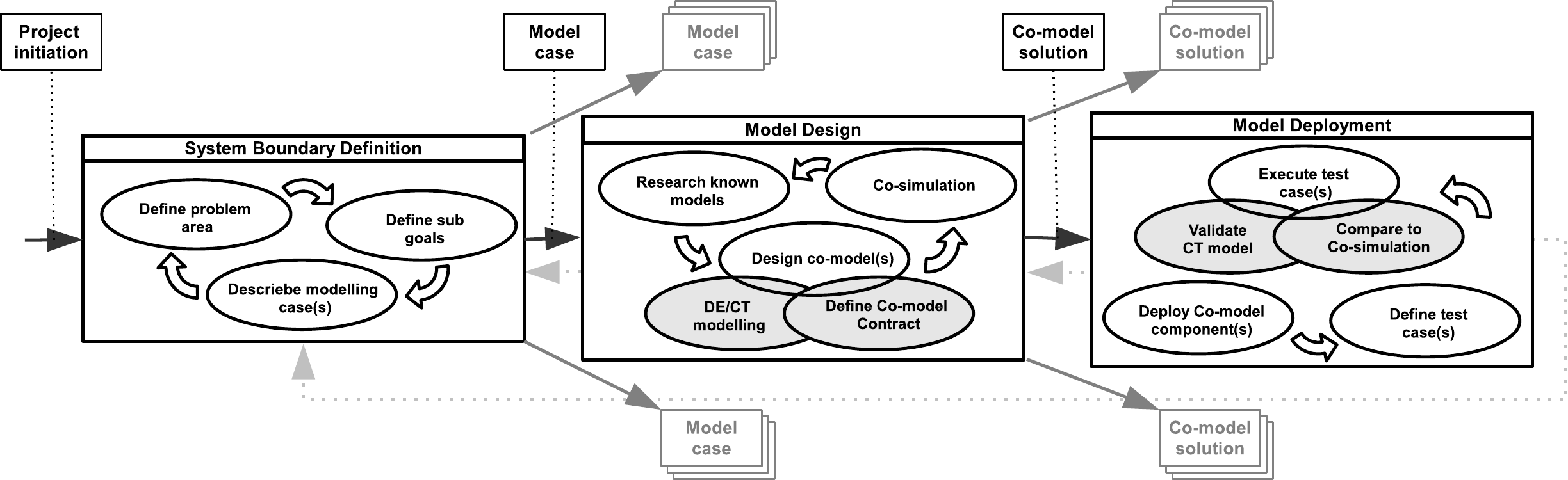}
		  \caption{Proposed co-modelling process workflow, from initial problem definitions to model deployment.	
			Depending on the project, each stage can result in one or more branches of the next stage.}
		  \label{fig:modelling-process}
	 \end{figure}

	The proposed co-model development methodology and process workflow are illustrated in Figure~\ref{fig:modelling-process}, 
	and are designed to encompass the full process from problem definition to deployment of a useable system.
	The co-modelling development methodology addresses model-based development for automated and robotics agricultural 
	ground vehicles operating under the changing operational conditions described in Sections~\ref{sec:agro_intro} and~\ref{sec:movtivation}.
	Therefore, one contribution of this thesis is:
	
	\contribution[methodology]{contribution:methodology}{An extended co-modelling and co-simulation methodology for 
				the development of automated and robotic agricultural ground-vehicle systems.}

	The work in~\cite{Fitzgerald&14c,Wolff13a} has partly inspired the proposed methodology used for the co-model development conducted in this study. 
 	Where these methods primarily focus on movement from controller requirements and environment assumptions to a complete system co-model, 
	we extend this approach to include deployment to the actual system realisation. 
	In this thesis, the new methodology is called an extended development methodology, as it extends previously described work. 
	
	The extended  methodology is intended to be used in the design of new product solutions and as an approach to updating currently available products such as tractors.
	The motivation for developing this extended methodology is to gradually move towards automated or robotic agricultural ground-vehicles and 
	the intended semi-controllable operational environment.
	The proposed extended methodology allows the development team to validate their co-model solution against a developed product solution.

	The extended development methodology is structured to produce the intended system gradually, with a finer level of granularity being achieved in each stage. 
	The development process incorporates both the model developers and the remaining project stakeholders, to ensure input from the related partners. 
	The development process divides the workflow into three main stages through which the co-model development iterates:
	\textit{System Boundary Definition}, \textit{Model Design} and \textit{Model Deployment}.  
	The main focus of the extended methodology is the given concrete objective of the developers for a particular system. 
	The extended methodology provides guidelines to aid in the selection of the relevant aspects being co-modelled.
	
	The extended development methodology is designed to provide agile development from initial concept to product solution. 
	Process iterations occur both internally and through movement between the development stages, based on the input from the project stakeholders. 
	In Figure~\ref{fig:modelling-process}, movement forward to the next main stage is indicated by black arrows and occurs when the project developers initiate the next stage in agreement with the stakeholders. 
	Each successive main stage can branch into multiple instances as different aspects of the system may be developed separately. 
	We do not claim that the output of the development process at each stage in our approach is necessarily a refinement of its predecessors. 
	If the project stakeholders discover the need to redefine a model case or co-model solution, backward movement in the development process to revisit an earlier stage can be initiated.

	Each main stage in the extended development methodology contains three substages that are cycled between during development of the intended system. 
	The developers should expect to run through the substages multiple times and advance based on the previous iteration. 
	The stakeholders should continually update the project timeframe based on the progress achieved using the extended methodology.
	Here, we provide a description of the three main stages of the extended development methodology:
	
	\begin{description}[leftmargin=*,itemindent=0cm]
	\item[System Boundary Definition:] 
		The initial decision in the development process is the definition of the boundaries of the problem area. 
		Boundaries can be defined based on current standards in the domain area, the demands of the stakeholders, and the limitations of the system. 
		The stakeholders decide their aim regarding the model, in terms of controller type and operating vehicle/environment. 
		
		Based on the problem area definition, the stakeholders then define the subgoals the system modelling must help realise. 
		For a grass-cutting robot, the subgoals might be to achieve in-field automated navigation, controlled grass-cutting operation, and field management in relation to fuel consumption. 
		For a manually driven ground vehicle used for animal feeding, the subgoals might be to achieve automated feeding operation and selection of robot arm solution for the feeding process.
		The stakeholders should rank the development subgoals based on their deemed importance to the deployment of the product.

		The task of the developers is to determine if the subgoals can be developed into separate co-modelling cases, or if several subgoals should be grouped into a single case. 		
		A co-modelling case focuses on specific parts of the system relevant to the selected subgoals and abstracts all other parts. 
		The development team achieves a form of decoupling that ensures a lower level of interdependence, by allowing subgoals to be developed in separate co-modelling cases.

	\item[Model Design:]
		New co-models should not be reinvented by the developers for each project. 
		In the design of viable model components that can effectively represent the modelling case, we recommend legacy domain knowledge from DE and CT modelling. 
		In the majority of cases, one can find related case studies from agriculture or other industry domains, aspects of which can then be implemented in the project in question. 
		The task is to select model components that the developers deem most likely to fit the demands of the co-model case.
		
		The co-model design is comprised of two parts: the co-model structuring and implementation of the CT/DE models. 
		The co-model structuring results in the co-model contract, which defines the communication between the models and is known to both DE and CT developers. 
		When the initial version of the co-model contact has been designed, the DE and CT developers can start implementing their individual co-model parts.
		A CT- or DE-first modelling-based approach can still be used, but co-modelling cannot commence until the co-model contract is defined.

		When a new revision of the co-model has been produced, co-simulation can be executed to analyse and verify the model response. 
		Different candidate solutions can also be compared and analysed using co-simulation, as they can be easily run for the same scenario. 
		The objective is to meet the goals set by the stakeholders, so that the solution can be deployed onto the actual platform realisation.

	\item[Model Deployment:] 
		The first step to deploying the co-model is porting the CT or DE components to the actual platform.
		CT deployment could be the implementation of actuators or mechanical constructs. 
		Co-simulation can be used to select a viable mechanical solution, for example an arm for animal feeding operations, which is implemented and tested on the actual platform. 
		DE deployment converts the model into software code that can be directly executed on the intended platform.

		To validate that the solution obtained from the co-model has been implemented correctly, the development team should define a number of test cases, 
		which are intended to validate that the  predictions from the simulation are sufficiently close to the actual realisation.

		The intended final internal state is the execution of the test case(s), which should verify that the co-model solution is implemented in the design of the automated or robotic ground vehicle. 		
		When the implementation of the co-model solution has been verified, the focus can move to other aspects of the project.
	\end{description}
	
	Development using this extended methodology should never be viewed as complete, as a system can be moved in a new direction or new uses can be found. 
	Lessons learned from the deployment can be used to initiate the development of the next generation of product solution.
	A deployed system solution can be adapted to related areas through further development. 
	For example, a robotic system that can feed pigs may be adapted to feed cows, sheep, or other farm animals.
	The light grey arrow from \textit{Model Deployment} to \textit{System Boundary Definition} in Figure~\ref{fig:modelling-process} indicates that a finished system can be used to initiate new systems development. 
	
	In the following sections, the main modelling cases that have been used to define the extended development methodology are described.
	
	\section{Case Study: The LEGO Micro-Tractor Platform}

	The lack of reported automated and robotic ground-vehicle co-models when this PhD project was initiated, made it clear that movement towards these model types needed to be conducted gradually. 
	Many of the parts for such systems are not realised, making it difficult to completely verify a co-model based on testing and analysis.
	LEGO Mindstorms is an example of a common framework that has been used in other scientific disciplines related to robotics, 
	e.g., robotic exploitation~\cite{Kovacs&11} and team intelligence~\cite{simonin&09}. 
	LEGO Mindstorms provides a proven, versatile framework for prototyping mechanical robotic systems that are programmed with a high degree of complexity. 
	It also provides a system that has the ability to add and remove functionalities, as well as to reconfigure its architecture.
	
	\subsection{Co-modelling}
	A co-model based on a LEGO Mindstorms NXT tractor (micro-tractor) was developed, to simplify the initial co-modelling design process.
	The micro-tractor is intended to be a representative scaled model of agricultural machinery used to test and demonstrate operations. 
	This prototype co-model of an automated vehicle is intended to provide an abstraction with key components in autonomous vehicle steering.
	Therefore, a contribution of this thesis is:
	
	\contribution[legosim]{contribution:legosim}{A kinematic co-model of an automated LEGO tractor platform.}
	
	 \begin{figure}[tb]
		  \centering
		  \includegraphics[width=0.5\textwidth]{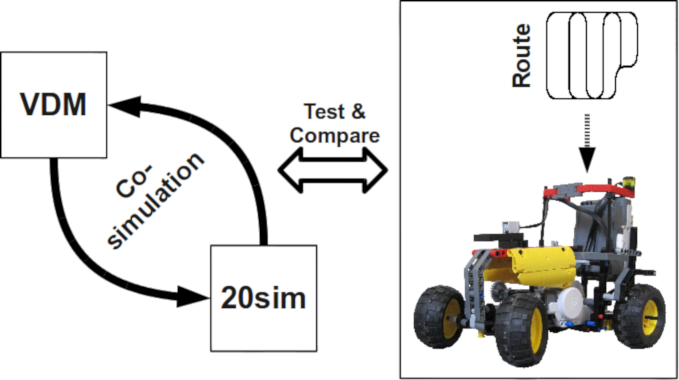}
		  \caption{Overview of the Lego©Mindstorm©NXT tractor and the co-modelling method. 20-sim models the vehicle/environment and VDM models the control part of the micro-tractor .
		 Comparisons are made between real and simulated systems to determine errors and shortcomings in the model.}
		  \label{fig:tractor-VDM}
	 \end{figure}
	 
	The DE controller models the NXT’s steering of the micro-tractor and is modelled using VDM-RT. 
	A pre-planned route is given to the autonomous system,  
	which is used when the micro-tractor commences its task in a given field area.
	The route is based on a collection of continuous curve elements. 
	Each continuous path element is either a line segment or circular arc with constant radius, containing a start and stop waypoint~\cite{Bevly09}. 
	The micro-tractor is aware of its current position and can use this information when following the route.
	A route-manager ensures that each route segment is performed in the order described in the route. 
	The VDM model uses invariants and pre- and post-conditions to ensure that only a viable route and route segments are commenced.

	A similar methodology is applied when modelling the CT components in 20-sim. 
	A combination of bond graphs and iconic diagrams are used to model the CT part of the micro-tractor co-model. 
	The bond-graph model motor and environmental dynamics and the iconic diagram are used for differential equations and sensor models.

	\begin{figure}[ht]
		\centering
		\includegraphics[width=0.95\textwidth]{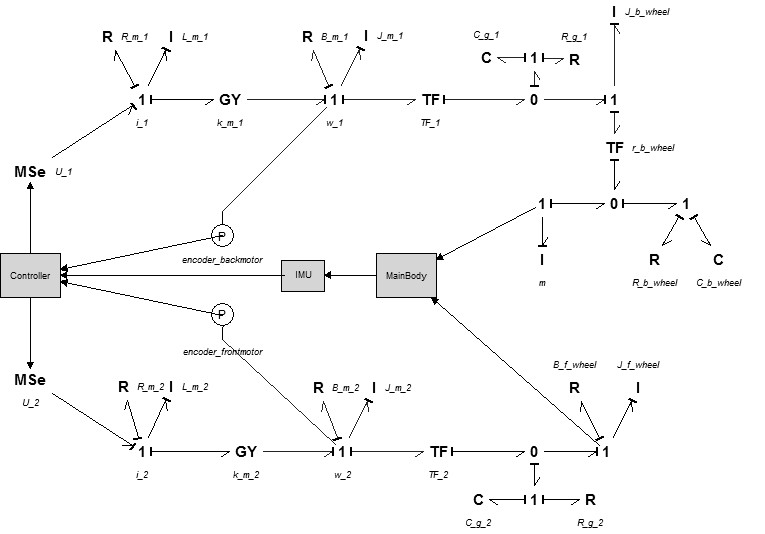}
		\caption{Micro-tractor model in 20-sim. A combination of bond graphs and iconic diagrams are used to model the micro-tractor.}
		\label{fig:20simmodelpaper1}
	\end{figure}
	
	The controller is connected to the DC-motor implementations, providing the interface for communication with the VDM model. 
	Interactions between the wheels and ground plane are only considered for a smooth surface to reduce the complexity of the model. 
	Only the longitudinal effects on the wheel are considered, since the tyre-road normal effort~\cite{Merzouki&07} is expected to be minimal. 
	
	To combine the dynamic effects of the back and front wheels, a first-order bicycle model was chosen.
	The resultant first-order bicycle model is a pure kinematic model~\cite{RoviraMas11} of the chassis movements, without regard for the forces acting on the body.
	The speed $u$ of the back wheels in combination with the rotational speed $\dot{\delta_f}$ of the front steering system are used as the model inputs.
	Then, the chassis movement is modelled as:
	
	\begin{equation}\label{eq:legodx}
	  \dot{x} = cos(\psi) u
	\end{equation}
	
	\begin{equation}\label{eq:legody}
	  \dot{y} = sin(\psi) u
	\end{equation}
	
	\begin{equation}\label{eq:legodtheta}
	 \dot{\psi} = \frac{tan(\delta_{f})}{L}u
	\end{equation}
	
	Equations~\eqref{eq:legodx},~\eqref{eq:legody},~and~\eqref{eq:legodtheta}  are used to calculate the vehicle rotation~$\psi$ and speed~$\dot{x},\dot{y}$ in the x-y direction in a global reference frame. 
	$L$ represents the distance between the front and back wheels  and $\delta_{f}$ is the orientation of the front wheels. 
	
	The experience obtained by modelling of the LEGO Mindstorms micro-tractor described above was used to refine the concept of the extended development methodology, 
	indicating that legacy models and related case studies should be researched before co-modelling is commenced.
	Even though the extended development methodology was updated based on the lessons learned from the modelling case study, 
	the structure illustrated in Figure~\ref{fig:modelling-process} was followed.

	\subsection{Testing operational management techniques}
	In order to quickly test operational management techniques, a test platform utilising a LEGO Mindstorms micro-tractor was developed, 
	allowing for easily replicable results that can be evaluated while interpreting collected data. 
	Compared to a Hardware-In-the-Loop solution, the micro-tractor allows for the evaluation of software components using actual sensory input. 
	This test platform is seen as an intermediary step between simulation and full-scale testing, rather than a replacement of either.
	
\begin{figure}[!ht]%
	\centering\hfill
	\begin{subfigure}[b]{0.6\textwidth}
                		\includegraphics[width=\textwidth]{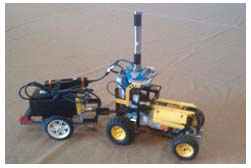}
                		\caption[legoplatform]{Micro-tractor with mounted iGPS sensor.}
                		\label{fig:lego-platform}
        		\end{subfigure}\hfill
        		\begin{subfigure}[b]{0.33\textwidth}
                		\includegraphics[width=\textwidth]{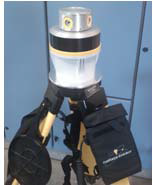}
                		\caption[igps]{iGPS beacon.}
                		\label{fig:lego-iGPS}
        		\end{subfigure}\hfil
        	\label{fig:lego-test-system}%
	\end{figure}

	The micro-tractor is equipped with a drawbar suitable for connecting implements. 
	In this research, an indoor GPS (iGPS) was used to test the accuracy of the micro-tractor position determination. 
	The iGPS system (Nikon Metrology, NV Europe) combined a transmitter sensor placed at the center of the rear axle of the micro-tractor 
	(Figure~\ref{fig:lego-platform}) with six beacon posts (Figure~\ref{fig:lego-iGPS}) located around the working area.
	
	In the related paper~\cite{Edwards&13}, the test platform was described in terms of its hardware and software components.
	The performance of the platform was demonstrated and tested in terms of its capability for supporting decision making on field operation planning using indoor environment simulations. 
	Further the micro-tractor was equipped with a drawbar suitable for connecting implements.
	Therefore, another contribution of this thesis is:
	
	\contribution[legodeploy]{contribution:legodeploy}{Deployment of a co-model solution to a LEGO tractor to allow for two-step gradual verification and movement from simulation to an actual system.}
	
\begin{figure}[ht]
	  \centering
	  \includegraphics[width=0.685\textwidth]{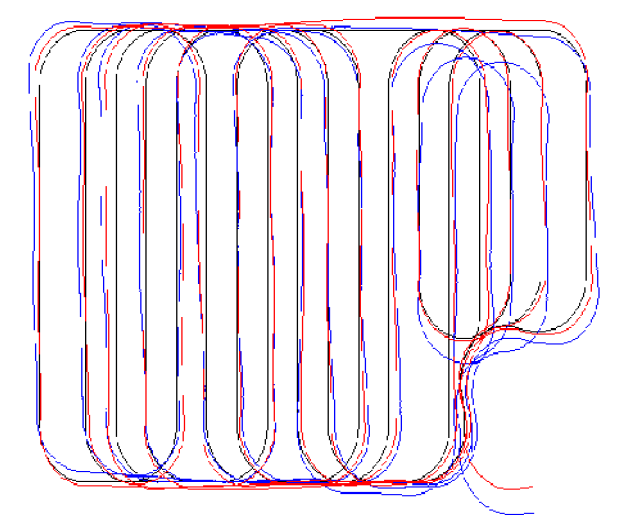}
	  \caption{Planned path (black line), path estimated by micro-tractor internal sensors (red line), and actual path recorded by  iGPS (blue line).}
	  \label{fig:legopath}
	\end{figure}

	A series of navigation accuracy tests were performed in a “virtual” field. 
	Figure~\ref{fig:legopath} presents the three paths (off-line planned, 
	on-line estimated and actual measured) on a “virtual” field for the case of a 0.250 m working width and 0$^\circ$ driving direction. 
	Based on the tests, for a basis driving distance of 71.43 m  (corresponding to a 1 $km$ full-scale distance), 
	including straight line driving (operating on a field-work track) and 180$^\circ$  maneuvering (headland turnings), 
	the average cross-track error between the estimated path and the iGPS path executed by the micro-tractor was 0.028 m, 
	corresponding to a 0.39 m full-scale Cross-track error (XTE).
	
	The lessons learned from the testing using the micro-tractor were used to define the model deployment development stage to achieve actual realisation using the extended methodology in Section~\ref{sec:model_methodology}. 
	Note that the goal is not be the co-model in itself, but rather the realisations one can derive from the model-based development methodology.  
	
	The micro-tractor system example encapsulates the basic measures necessary for developing a complete test platform for field operations, 
	where route plans, mission plans, and multiple-machinery cooperation strategies can be simulated and tested in the laboratory. 
	The execution of coverage plans was chosen to demonstrate the capabilities of the test platform to implement agricultural operation management techniques. 
	The demonstration examples also show that it is possible to evaluate coverage plan scenarios 
	involving various operational features (e.g., working widths, driving angles, and number of headland passes)
	in terms of various operational efficiency measures, e.g., the measured non-working travelled distance, overlapped or missed area, and the operational time.

	\section{Case Study: Common Agricultural Ground-vehicle Platforms.}
	\label{sec:case_study2}
	
	The LEGO case study demonstrated the need to develop a base co-model that takes more aspects of a normal agricultural vehicle platform into account. 
	In the current market, a significant number of agricultural vehicles are based on front- or back-axle-operated steering, like tractors or combine harvesters.
	Skid steering, also known as differential steering, four-wheel steering, and other steering schemes are becoming more common~\cite[Ch. 2]{Zhang&2013b}. 
	In this study, it was decided to focus the modelling development on front- or back-axle-operated steering, as this is still the most common platform 
	and developers can utilise this approach in automated or robotics projects.
	Therefore, another contribution of this thesis is:
	
	\contribution[modelcase2]{contribution:modelcase2}{A dynamic co-model of a front- or back-axle-steered automated agricultural ground-vehicle platform.}
	
\begin{figure}[bth]
		\centering
		\includegraphics[width=\textwidth]{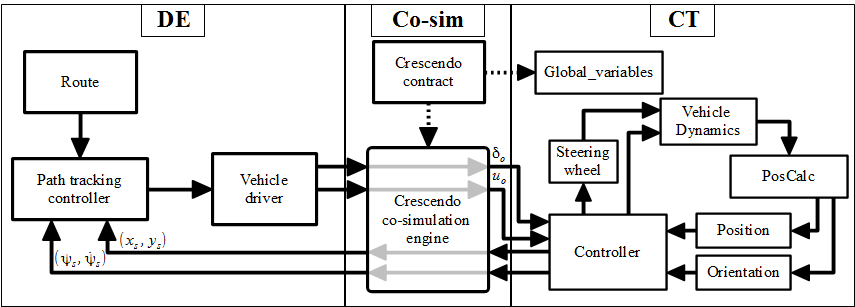}
		\caption{Communication between the CT and DE models of the envisioned 
				automated and robotic vehicle base co-model using the Crescendo co-simulation engine.}
		\label{fig:comodelstructure}
        \end{figure}

	The base co-model for a front- and back-steered vehicle is complementary to the extended development methodology presented in Section~\ref{sec:model_methodology}, designed to support future model-based development projects.
	The base co-model illustrated in Figure~\ref{fig:comodelstructure} is implemented in agricultural development cases where automated path following is of interest.
        
    	This co-model allows multiple views and permits the developers to focus on  six shared variables.
	This simplification is intended to allow DE and CT developers to focus on their own domains and only requires them to be aware of the Crescendo interface. 
	Both the DE and CT model parts are well-known components; the strength here is that a developer can utilise a development tool related to their domain~\cite{Backman&10,Backman&12,Bevly09,Karkee&10}.
	The idea is to have a base agricultural vehicle co-model structure that can be used in a number of different development cases to complement the extended development methodology.
	Other shared design parameters (\textbf{sdp}) or variable extensions may be implemented in a specific project, to allow for analysis of the concrete problem on which the developers are working.

	Two \textbf{controlled} ($u_o, \delta_o$) variables are used to operate the actuators for vehicle speed and steering axle angle. 
	In some cases, $u_o$ is assumed to be constant for specific co-simulation and is instead a \textbf{sdp}; 
	this simplifies both the DE and CT parts of the co-model. 
	With an update to the vehicle driver block on the DE side, the DE-model can be reused for differential and four wheeled vehicle steering. 
	In the case of a differentially steered vehicle, the controlled variables to the CT side would instead be the left- and right-side rotational speeds of the wheels.
	
	On the CT side of the co-model, all communication with the co-simulation engine is performed trough the controller interface, which contains local versions of \textbf{controlled} and  \textbf{monitored} variables.
	The \textbf{monitored} variables ($x_s$,$y_s$, $\phi_s$, $\dot{\phi}_s$) are sensor measurements that the DE controller receives, which concern the current vehicle position and rotational speed in the yaw plane. 
	Compared to the actual automated or robotic solution, one is not required to implement aspects of sensor fusion to obtain reliable vehicle state information for the DE side of the co-model. 
	Using abstractions of system components such as sensor-fusion, the developers on the DE side can focus directly on the ground-vehicle steering algorithm.
	The intention is not to model the full reality, but rather to provide a base model that allows us to analyse automated and robotic agricultural ground-vehicle design problems.
	
	\subsection{DE modelling}
	
		The DE part of the base co-model assumes that the developers intend the robot to follow a pre-planned or updated infield route.
		The DE-mode utilises a route based on a waypoint sequence and path tracing methods that can select the waypoints the ground vehicle must follow at a given instance in time.
		The model utilises the method given in Listing~\ref{lbl:route_waypointl:vdm} to retrieve the current and next waypoint  from the sequence.
		The  VDM variables \textit{current\_waypoint} and \textit{next\_waypoint} are nominated $P_{k}$ and $P_{k+1}$, respectively, where k represents the current element in the sequence.
		
		To ensure that the update method for the route is not called for an empty waypoint sequence, a precondition is set for the minimum length. 
		When running a co-simulation, the precondition functionality can be used to warn the developer if the condition is met and the update method is, therefore, being called under the wrong pretences.

	\begin{vdm_al} [\lstset{caption=VDM-RT method for updating to the next waypoint in the sequence contained in the route., label={lbl:route_waypointl:vdm}}]
public NextWayPoint: () ==> WayPoint
NextWayPoint() == (
	-- Goto next waypoint in sequence
	current_waypoint := (hd waypoints);
	-- Remove waypoint from sequence
	waypoints := (tl waypoints);
	-- Get next waypoint in sequenc
	next_waypoint := (hd waypoints); 
	return(current_waypoint);
)pre (len waypoints > 1);
	\end{vdm_al}
		
\begin{figure}[b]
			\centering
			\includegraphics[width=0.98\textwidth]{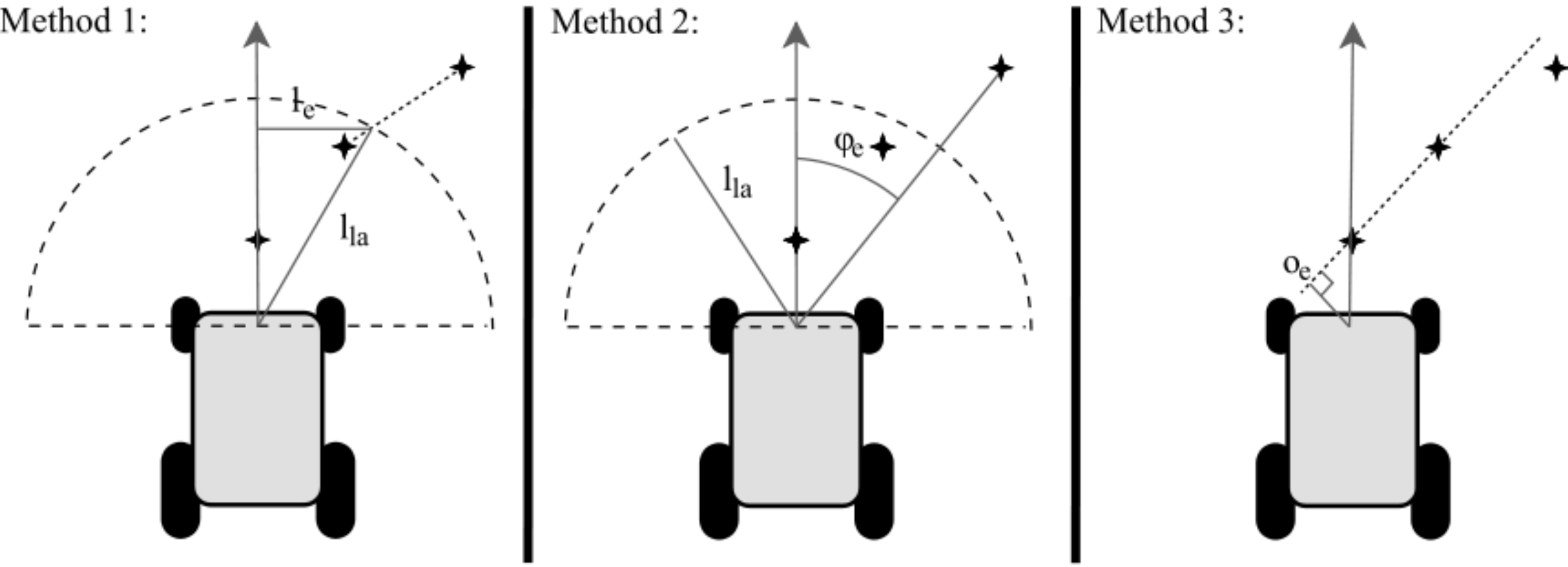}
			\caption{Three waypoint path-tracking methods: 
				Method~1: Heading error path-following based on look-a-head distance $l_{d}$ and a single waypoint.
				Method~2: Lateral error path-following based on $l_{d}$ and two waypoints.
				Method~3: Line segment path-following based on $l_{d}$ and two waypoints.}
			\label{fig:waypoint_follow}
		\end{figure}

		The three different path-tracking controller methods that have been implemented as 
		DE model components and utilised as single or mixed solutions to steer ground-vehicles are illustrated in Figure \ref{fig:waypoint_follow}.
		One should note that that the path tracking controller methods use either one or two waypoints to calculate the vehicle steering response.
		
		The look-a-head distance $l_{d}$ is used to determine both the current waypoints for the
		vehicle to follow and when the vehicle control should move to the next waypoint in the sequence. 
		The magnitude of $l_{d}$ is dependent on the vehicle type and the speed of the controlled vehicle.
		Different path-tracking methods are utilised depending on the operational task the vehicle must perform.
		Examples of the utilisation of different  path tracking controllers can be found in~\cite{Christiansen&13a},~\cite{Christiansen&15a},~and~\cite{Christiansen&15b}.
	
        \subsection{CT modelling}
        
\begin{figure}[htb]
	            		\centering
	            		\includegraphics[width=0.95\textwidth]{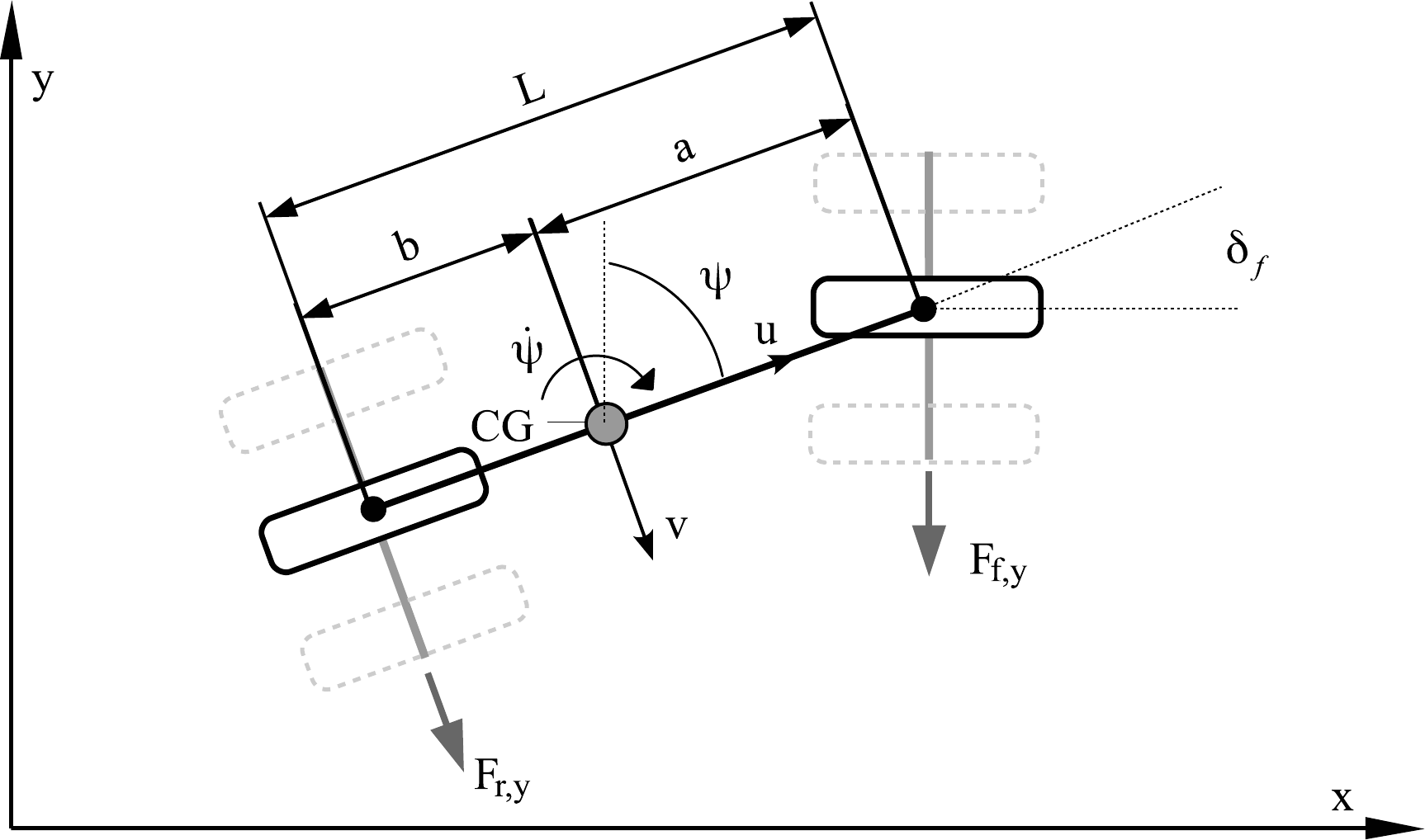} 
	            		\caption{Dynamic bicycle model of the vehicle part of the robot system.}
	            		\label{fig:bicyclemodel}
	            \end{figure}

	  	The CT part defining the vehicle is a non-linear model with three Degrees of freedom (DOF), 
	  	i.e., the longitudinal, lateral, and yaw directions, irrespective of the suspension and described in~\cite{Christiansen&15a}.
		The model of the vehicle utilises the bicycle approach, meaning that the lateral forces on 
		the left and right wheels are assumed to be equal and summed together.
	      This assumption holds for typical agricultural vehicle operation velocities~($<$7.5~m/s)~\cite{Karkee&10}.
	      The bicycle structure is also known as a half-vehicle~(Figure~\ref{fig:bicyclemodel}).
	      The model allows for yaw and lateral motion through adjustment of the front wheel angle $\delta_f$.
	
		The velocities $u$, $v$ are at the Center of gravity (CG) of the vehicle. 
		$L$ is the wheelbase, where $a$ is the longitudinal distance to the front wheel, and $b$ is the longitudinal distance to the rear wheel. 
		For a constant forward velocity, the vehicle motion is given by
		\begin{equation}
			m(\dot{v}+u\dot{\psi}) =F_{f,y}cos(\delta_{f})+F_{r,y}
			\label{eq:rotation1}
		\end{equation}
		where $r$ is the angular rate about the yaw axis.
		Similarly, the vehicle yaw motion is expressed by
		\begin{equation}
			I_{zz}\ddot{\psi} = aF_{f,y}-bF_{r,y}
			\label{eq:vehicle_yaw_motion}
		\end{equation}
		\noindent where $I_{zz}$ is the moment of inertia along the yaw axis.
		
		A similar solution can be used for a back-axle-steered vehicle solution, where steering wheel angle $\delta$ impacts the forces from the rear wheel. 
		The ability to change the co-model to back-axle steering allows the solution to be used for combine harvesters and similarly steered vehicles.
		
	\section{Case Study: Animal Feeding System}
	\label{sec:case_study3}
	The base co-model case study is intended to provide a base co-model that is applicable to a large number of agricultural cases. 
	However, this model cannot be applied in the study of aspects relevant to a special case.  
	For example, the base co-model cannot be used to model aspects of sensor-fusion or other operational vehicle dynamics directly, and 
	more specialised modelling solutions are needed to satisfy the requirements of these kinds of co-modelling cases. 
	A specialised co-model solution allows the development team to analyse unique aspects of the given system, 
	but this new co-model's levels of re-usability in other projects is also reduced. 
	In this research project, a robotic ground-vehicle for mink feeding was used to study of specialised co-modelling solutions, 
	and was developed using the extended development methodology presented in Section~\ref{fig:modelling-process}.
	Therefore, a further contribution of this thesis is:
	
	\contribution[feeding]{contribution:feeding}{A specialised co-modelling solution for automated and robotic agricultural ground-vehicle solutions related to mink feeding.}
	
\begin{figure}[tbh]
	\centering
           \includegraphics[width=\textwidth]{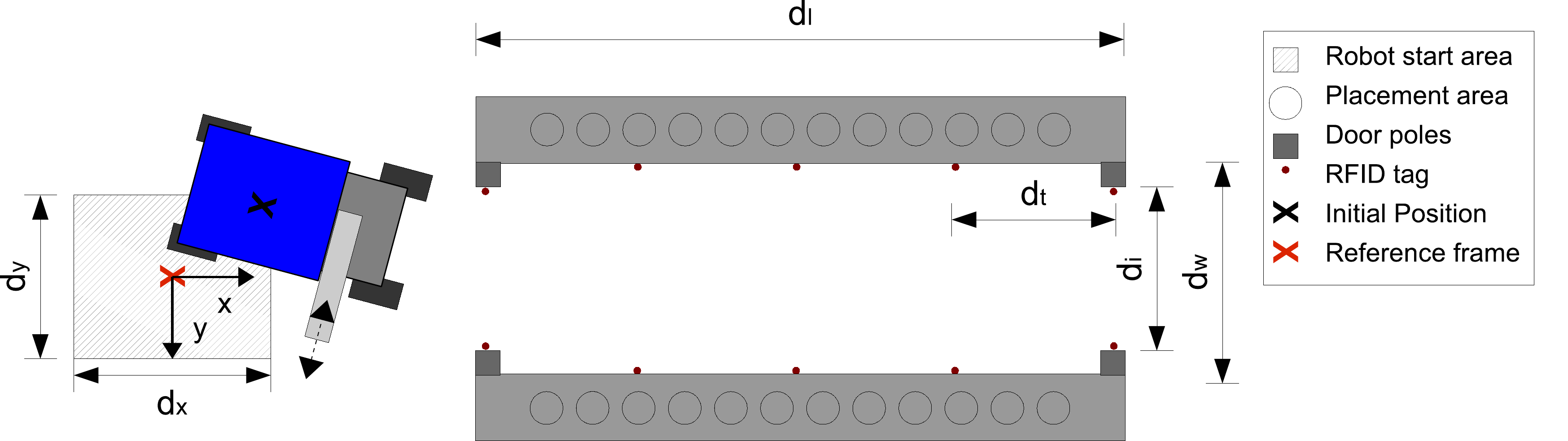}
           \caption{Sketch of the load-carrying robot  and the mink feeding area.}
           \label{fig:system_setup}
	\end{figure}

	The chosen robot is a four-wheeled vehicle with front-wheel steering and rear-wheel drive equipped with a differential gearing.
	The robot receives sensory input from a vision system, a Radio frequency identification (RFID) tag reader, Inertial measurement unit (IMU),
	and rotary encoders installed on the back wheels and front wheel kingpins.
	The vision system is used to detect the entrance to the feeding area, when the robot is still outside. 
	The RFID tags are placed along the rows of mink cages to act as fixed location reference points, as illustrated in Figure~\ref{fig:system_setup}.
	Fused sensory data are used to determine the current location and to enable the robot to perform its required actions in the environment.
	A feeder arm mounted on the robot is used to dispense the fodder on the cages at the predetermined locations. 
	When the robot moves into the feeding area, it stops to deploy the feeding arm and then begins the feeding procedure.
	
	Sensor fusion concepts and load distribution dynamics were co-modelled for this robotic vehicle solution.
	The robotic vehicle solution is described in additional detail in Section~\ref{sec:visualisation} and~\ref{sec:over-dse}, 
	where the base co-model and this co-model  are used for DSE. 
	The main focus here is the model extension of the base co-model using the extended methodology.  
	We here present the vehicle dynamics and sensor fusion parts of the co-model, which were developed based on the base co-model.
	
	\subsection{Vehicle body dynamics}
	
	The generated tyre forces interact with the robot to produce the output response in the environment. 
	To take the front steer angle into consideration, the CT-model rotates the front tyre forces into the coordinate system of the robot vehicle.

	\begin{figure*}[!bh]
	        \centering
		\begin{subfigure}[b]{0.40\textwidth}
			\includegraphics[width=\textwidth]{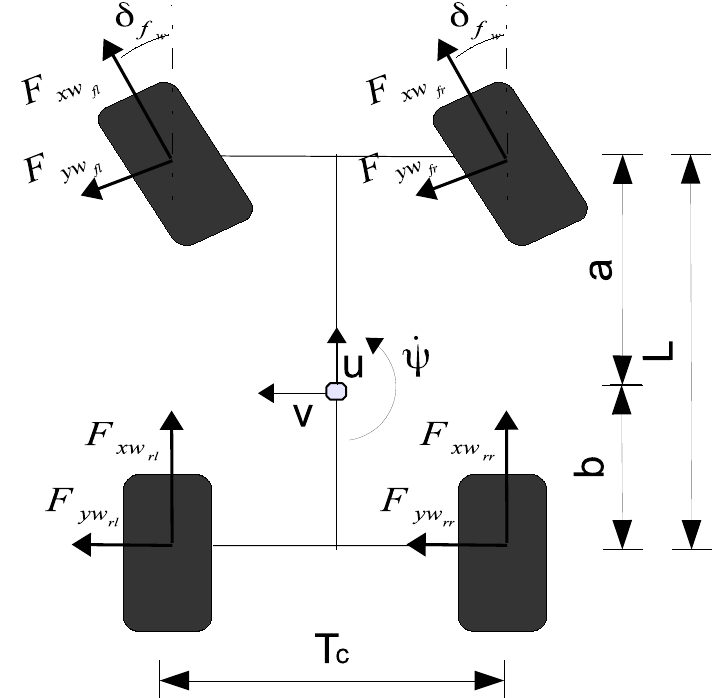}
			\caption{Forces in the X-Y plane.}
			\label{subfig:vehicle_model_xy}
		 \end{subfigure}%
		\quad
		 \begin{subfigure}[b]{0.40\textwidth}
			\includegraphics[width=\textwidth]{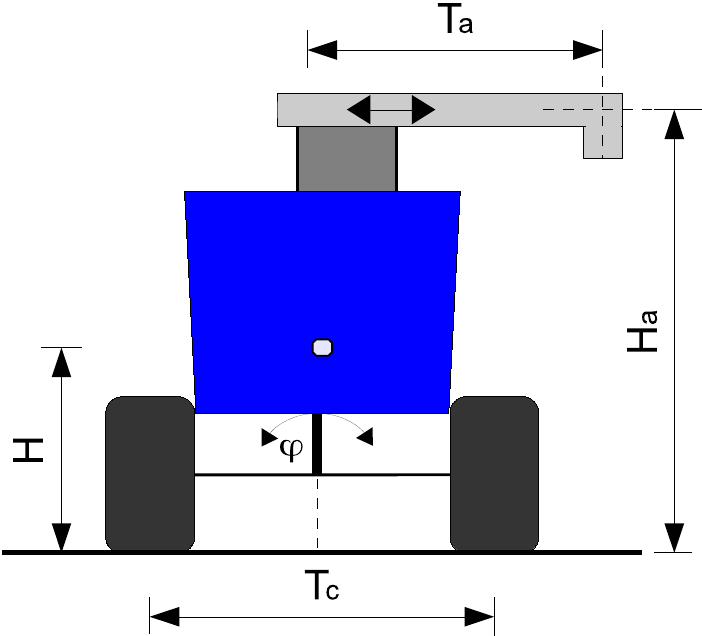}
			\caption{Forces in the Y-Z plane.}
			\label{subfig:vehicle_model_yz}
		 \end{subfigure}%
		\caption{Free-body diagrams of the robot in the x-y and y-z coordinate frames.}
		\label{fig:vehicle_model}
	\end{figure*}
	
	Based on the trigonometric relationships illustrated in Figure~\ref{subfig:vehicle_model_xy}, 
	the steer angle was transformed using the rotation matrix at the front of the left-hand side of Equation~\eqref{eq:transfrom_tire_force}~\cite{Bevly09}. 
	The robot utilises Trapezoid- steering, which produces equal steering angles on the right and left sides, making the transformation the same for both sides. 
	Thus,
	
	\begin{equation}
		\begin{bmatrix}
			 F_{xlf} \\
			 F_{ylf} \\
		\end{bmatrix} = 
		\begin{bmatrix}
			cos(\delta_f) & -sin(\delta_f) \\
			sin(\delta_f) & cos(\delta_f) \\
		\end{bmatrix}
	\begin{bmatrix}
			 F_{xw_{lf}} \\
			 F_{yw_{lf}} \\
		\end{bmatrix} 
	\label{eq:transfrom_tire_force}
	\end{equation}
	The CT side models the dynamic yaw~$\psi$, pitch~$\phi$, lateral~$u$, and longitudinal~$v$~motion responses of the robot vehicle. 
	The yaw motion is modelled by Equation~\eqref{eq:dd_yaw_calc}.
	Variables~$a$~and~$b$ are respectively the longitudinal distances from the front and rear wheels to the CG, 
	$I_{zz}$ is the yaw moment of inertia, and $T_c$  is the tyre base track width. Then,
	\begin{multline}
		I_{zz}\ddot{\phi}= a\left(F_{ylf}+F_{yrf}\right)-b\left(F_{ylr}+F_{yrr}\right)+\\
		\frac{T_c}{2}\left(F_{xlf}-F_{xrf}+F_{xlr}-F_{xrr}\right)
	\label{eq:dd_yaw_calc}
	\end{multline}
	Similarly, the robot roll motion is given by equation \eqref{eq:dd_roll_calc}, where $K_r$ and $C_r$ represent a combined rotational spring-damper 
	based on tyre values, with $\sum{F_{y}}$ being the sum of the forces in the longitudinal direction. 
	The variable  $H$ represents the current CG height above grounds, $M$ is the current total mass of robot $m$ and transported load, and $g$ is the Earth's gravity. 
	\begin{equation}
		I_{xx}\ddot{\psi}=-\left(H\sum{F_{y}}+(K_r-M g H)\psi+C_r\dot{\psi}\right)
	\label{eq:dd_roll_calc}
	\end{equation}
	The load shift from the right to the left of the robot is assumed to be proportional to the current pitch angle. 
	The longitudinal motion output of the robot is given by
	\begin{equation}
		\sum{F_{y}} = M\left(\dot{u}-v\dot{\phi}\right)
	\label{eq:d_u_calc}
	\end{equation}
	The lateral motion of the feeder robot is determined by the sum of the lateral forces, the roll acceleration, and the longitudinal speed, such that
	\begin{equation}
		\sum{F_{x}} =M\left(\dot{v}+u\dot{\phi}+H\ddot{\psi}\right)
	\label{eq:d_v_calc}
	\end{equation}
	
	\subsection{Sensor fusion}
	
	The concept behind sensor fusion is that more accurate estimates of a physical phenomenon can be obtained by combining different sensor-data sources~\cite{Hall&97}.
	The combined sensor data can better accommodate uncertainty and noise in measurements~\cite{Thrun01}.
	The sensor fusion solution adopted in this study uses an Extended kalman filter~(EKF)~\cite{Thrun&05} to estimate the current position of the robot.
	The process is represented by the following velocity motion model:
	\begin{multline}\label{eq:process_model}
	 f(\hat x_{k-1},\mu_{k},0 ) =
	 \begin{bmatrix}
	  x_{k} \\ 
	  y_{k} \\
	  \psi_{k} \\
	 \end{bmatrix} =
	  \begin{bmatrix}
	  x_{k-1}\\ 
	  y_{k-1}\\
	  \psi_{k-1} \\
	 \end{bmatrix}
	+\\
	  \begin{bmatrix}
	  -\frac{u_{e_{k}}}{\dot{\psi}_{s_{k}}}\left(sin(\psi_{k-1})-sin(\psi_{k-1}+\dot{\psi}_{s_{k}}T_k)\right)\\
	  \frac{u_{e_{k}}}{\dot{\psi}_{s_{k}}}\left(cos(\psi_{k-1})-cos(\psi_{k-1}+\dot{\psi}_{s_{k}}T_k)\right)\\
	  \dot{\psi}_{s_{k}}T_k\\
	 \end{bmatrix}
	\end{multline}
	The process input $\mu_{k}$ at interval $k$ in time is used to predict the next state and is based on the monitored variables $\dot{\psi}_{s}$ and estimated speed $u_{e}$ based on rotary encoder values.

	The EKF utilises an event-based correction stage that is dependent on the inputs from the vision system and RFID. 
	The vision system provides updates when the door poles of the entrance and exit are in view, and compares them against a pole landmark map. 
	The chosen landmark $\left(m_{x,j},m_{y,j}\right)$ are converted into polar coordinates ($r,\theta$) to allow for direct comparison with the sensor input:
	
	\begin{multline}\label{eq:mes_laser}
	h_{vision_{out}}(x_{k,j},0)
	= \begin{bmatrix}
	 r_{k,j} \\
 	\theta_{k,j} \\
	 \end{bmatrix} =\\
	\begin{bmatrix}
	 \sqrt{(m_{x,j}-x_k)^2+(m_{y,j}-y_k)^2} \\
 	arctan2(m_{y,j}-y_k, m_{x,j}-x_k)-\psi_k \\
	 \end{bmatrix}
	\end{multline}
	where $(m_{x,j}$, $m_{y,j})$ is the position of the door-pole in the local map and ($x_k$, $y_k$, $\psi_k$) is the estimated position of the robot. 

	When the robot is moving inside the feeding area the vision input can be used to update the vehicle orientation and distance to the side wall~\cite{Hansen&11}.
	\begin{equation}\label{eq:mes_rfid_out}
		h_{vision_{in}}(x_{k,j},0)=
		\begin{bmatrix}
		d_m\\
		\theta_m\\
		 \end{bmatrix}
		  =
		\begin{bmatrix}
		 \frac{A_my_k+B_m x_k+C_m}{\sqrt{A^2_m+B^2_m}}\\
		 arctan2(A_m, B_m)-\psi_k \\
		 \end{bmatrix}
	\end{equation}
	where $A_m$, $B_m$ and $C_m$  are the parameters for the general form of the line equation representing the mapped position of the side wall. 
	The sensory update does not provide the robot with information about its current position along the side wall and, therefore, position correction is needed.
	
	The positions of the RFID tags can also be seen as points along the sidewall ($m_{x,i},m_{y,i}$). 
	When the RFID tag reader first detects the tag we can use this to provide a position estimate ($\Delta x_{e,i},\Delta y_{e,i}$) 
	relative to this tag by combining the detection event with input from the vision sensor. 
	In these co-modelling scenarios, we assume it to be at the centre of the detection zone (i.e. at zero), 
	making the relative position measurement output correspond the intersection point between the line (sidewall) and ellipse (detection zone). 
	The relating landmark of the RFID tag is then:
	\begin{multline}\label{eq:mes_rfid_in}
	h_{rfid}(x_{k,j},0) =
	\begin{bmatrix}
		\Delta x_{k,i}\\
		\Delta y_{k,i}
	 \end{bmatrix}
	  =\\
	\begin{bmatrix}
		m_{x,i}cos(-\psi_k)-m_{y,i} sin(-\psi_k)-x_k\\
		m_{x,i}sin(-\psi_k)+m_{y,i}cos(-\psi_k)-y_k
	\end{bmatrix}
	\end{multline}
	
	The Jacobian matrices utilised in the EKF localisation method are not presented, but can be calculated based on equations \eqref{eq:process_model}, 
	\eqref{eq:mes_laser}, \eqref{eq:mes_rfid_in} and \eqref{eq:mes_rfid_out}.

	\section{Deployment}

		In this section, we  summarise the different modelling cases that have been deployed on agricultural automated or robotic ground-vehicle platforms.
		The development of models for the sake of modelling and simulation only is not the intended focus of this thesis. 
		Rather, the use of co-models to analyse and solve actual design cases in the agricultural domain and 
		the application of the acquired knowledge to ground-vehicle realisation (deployment) is the primary aim. 
		This approach is also in accordance with the extended agricultural development methodology presented in Section~\ref{sec:model_methodology}.
		
\begin{figure}[tbh]
		  \centering
		  \includegraphics[width=0.931\textwidth]{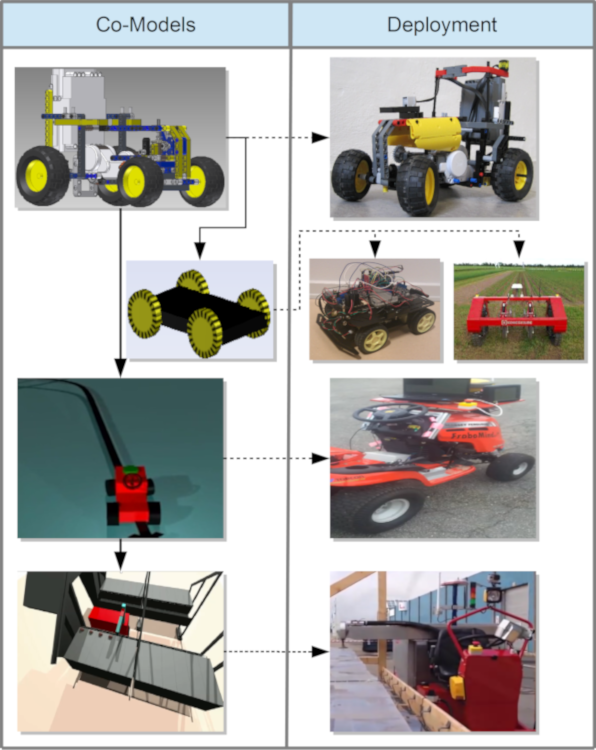}
		  \caption{Co-models that have been developed and deployed to actual ground-vehicle system platforms.}
		  \label{fig:deployed-models}
		  \vspace{-10pt}
		\end{figure}

		This chapter has primarily focused on cases involving front- or back-axle-steered ground vehicles, 
		but a solution for differentially steered vehicles has also been deployed. 
		In the majority of scenarios, DE components were moved from the Crescendo co-modelling environment 
		and developed into executable code for the different solutions illustrated in Figure~\ref{fig:deployed-models}. 
		The arrows between the co-models indicate how the previous co-models were utilised to develop the next generation.
		
		The LEGO case study was the first to be deployed onto an actual platform for testing and demonstration purposes~\cite{Christiansen&12a,Edwards&13}. 
		The LEGO tractor should be regarded as an intermediate step between modelling, simulation and 
		deployment to an agricultural ground-vehicle platform to provide gradual verification. 
		The four-wheeled differentially steered co-model was utilised in a master's thesis project and was deployed to a prototype platform~\cite{Jorgensen12}. 
		The same co-model was used in the modelling of noise encoder feedback on the KongsKilde Vibro Crop Robotti platform~\cite{Christiansen&14a,Green&14}. 
		Deployment of the ASuBot\footnote{An acronym for Aarhus and Southern Denmark University Robot.
		Source: \url{http://www.frobomind.org/index.php/FroboMind_Robot: ASuBot}} vehicle platform allowed us to confirm the functionality 
		of the path-following control system developed in the VDM formalism~\cite{Christiansen&13a},
		and actual verification allowed the path following functionality to be reused in the remaining presented co-models.
		
		The feeding system discussed above extended the DE model controller functionality to the operation of an animal feeding system for mink~\cite{Christiansen&15b,Christiansen&15a}. 
		The feeding operation was tested in a mink farm environment with a vehicle matching the co-model structure. 
		
	\section{Summary}
		This chapter has presented extended methodology guidelines to address the development of automated and robotic agricultural ground vehicles, 
		based on the state-of-the-art in this field, along with summaries of the three primary co-modelling case studies described in this thesis. 
		The deployment results show how co-modelling has been used in the development of a number of different automated and robotic agricultural ground-vehicle solutions. 
		The results indicate that co-modelling and co-simulation can strengthen the future development of automated and robotic solutions in the agricultural domain.

\chapter{Design Space Exploration}
\label{chap:dse}

\begin{figure}[tb]
	        \centering
			\includegraphics[width=0.81\textwidth]{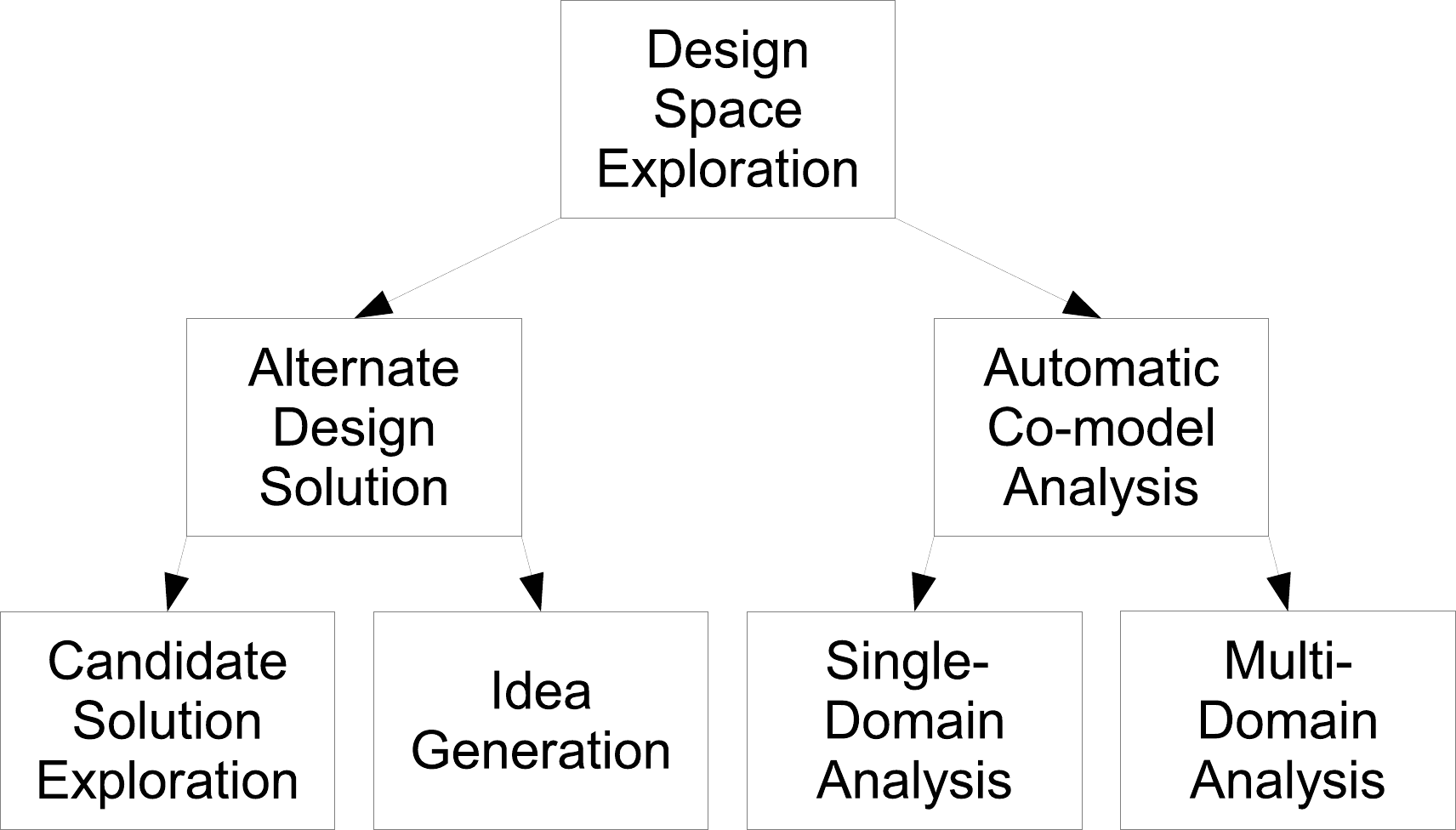}
			\caption{Overview of the different design space search approaches}
		 \label{fig:DSE_chapter_overview}
	\end{figure}

	This chapter presents the research and contributions of this thesis to the area of DSE in relation to co-modelling and co-simulation. 
	The chapter begins by introducing the basic concepts of DSE and their use in other projects. 
	Figure~\ref{fig:DSE_chapter_overview} provides an overview of the research conducted 
	using DSE and outlines the four contributions presented in this chapter. 
	This chapter discusses experimentation with alternate design solutions based on DSE, 
	in order to discover product solutions that can be implemented on actual platform realisations and 
	to provide insight into alternative solutions to a problem. 
	The chapter describes ACA of two case studies 
	for single- and multi-domain design problems. 
	This approach was used to find potential viable candidate solutions.
	
	The publications~\cite{Christiansen&14c,Christiansen&13a,Christiansen&15b,Christiansen&15a} are related to this chapter. 
	A total of four contributions are derived from these publications, which are presented in separate sections.
	
	\section{Design Space Exploration Concepts}
	\label{sec:con-dse}
	
	The aim of Design Space Exploration is to provide the project stakeholders with the ability to conduct 
	experiments in the search for solutions that can meet the system requirements within the defined design space. 
	The stakeholders manually define the design space for the system, from which they wish to determine viable candidate solutions, 
	using the co-simulation response surface of the design space~\cite{Gamble&14}. 
	The design alternatives in a design space can be a continuous parameter range, discrete value ranges, or distinct mode alternatives. 
	One could, for example, use the different path tracking-methods (modes) in Figure~\ref{fig:waypoint_follow} 
	and their configurations to define a design space for the operation of an automated vehicle in a specific set of scenarios. 
	The point is that the stakeholders must select which aspects of a co-model they find of interest and wish to explore.
	
	The motivation for using DSE in combination with co-simulation is to find candidate solutions for a specific aspect of a system, 
	such as the actuation, controller, or sensor operation of a robotic system~\cite{Ni&12}.
	The use of DSE relates to the  design problem illustrated in Figure~\ref{fig:grass_cutters}, since the method can be 
	used to allow the different stakeholders to compare candidates and to use the co-simulation response to select a solution. 
	Co-simulation can be used to explore alternate solutions for specific design aspects which would be too costly to test in the real world. 
	An informed discussion can then be conducted on which solution should be developed into an actual product and deployed into a platform solution, 
	by allowing the stakeholders to visualise the response of the system.

	One candidate solution might be deemed to be optimal, based on the output response of a specific co-model and a specific design space. 
	The intention in this thesis was to avoid focusing on identifying the best or optimal solution within a certain design space, 
	as factors not considered by the stakeholders might influence the outcome in the real world.
	One could still determine the minima or maxima responses of a defined design space, 
	but this does not mean that the stakeholders have found the best or optimal solution under all conditions.
	
	The design space is explored either automatically, manually, or using a combination mode, to determine viable candidate solutions. 
	The different overall design space search approaches are related to the DSE structure illustrated in Figure~\ref{fig:DSE_chapter_overview}. 
	Manually experimenting with a co-model can have two purposes: to aid the developer in better understanding 
	the co-model response and to determine if an alternative design solution the stakeholders have proposed is viable. 
	Research has shown that examples of how a desired system should perform are beneficial for the creative process~\cite{Herring&09}. 
	An idea developed through such a creative process could result in an alternative design solution outside the initially defined design space. 
	The potential for idea-generated candidate solutions outside the stakeholder's well-defined design space is 
	one of the reasons why we do not refer to the ''best'' or ''optimal''  'design solution for a system.
	
	The Crescendo technology provides an automatic functionality to search a defined design space called ACA. 
	The line-following robot in~\cite{Pierce&12a} was developed through the first ACA-based search for a candidate solution. 
	The intention was to determine a solution for the placement of the robot light sensors utilised for line-following. 
	A candidate solution based on preselected evaluation metrics was found by rerunning the same scenarios for different sensor-position candidates.
	The ACA functionality is designed to provide automatic search and optimisation methods in order to determine viable candidate solutions. 
	The idea is to provide a cost function~\cite{Chong&08} for the ACA functionality to categorise and compare candidate solutions, so as to automatically select the most promising candidates. 
	Automatic parameter sweeps using ACA are one such approach to finding viable candidate solutions, when little is known about the system configuration. 
	The parameter sweep search methods can be somewhat calculation intensive if a larger design space must be examined, making input from optimisation methods relevant. 
	
	Academic research has been conducted in the area of optimisation methods for multi-disciplinary systems~\cite{Deb07,Padmanabhan&03}. 
	Current publications concerning multi-disciplinary optimisation have focused on identifying solutions in cases where the system model 
	has been developed using a single tool and the design space is based on a bounded continuous parameter range. 
	When new automatic search methods is developed for ACA functionality, one should attempt to adopt optimisation methods that have been used for single- or multi-domain problems. 
	
	\section{Alternative Design Configurations}
	\label{sec:visualisation}

\begin{figure}[htb]
			\centering
			\includegraphics[width=0.82\textwidth]{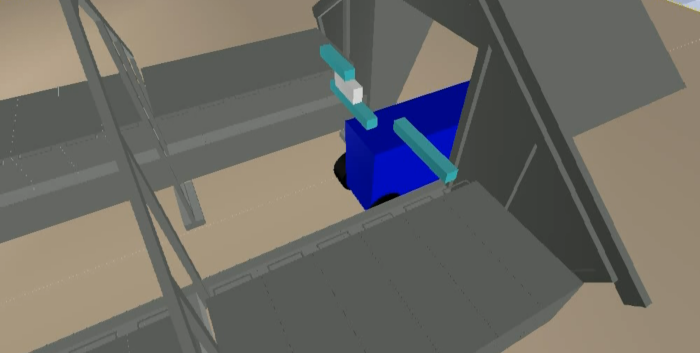}
			\caption{Exploring feeding-arm solutions for the mink robot.}
		        \label{fig:DSE_feed_v1}
		\end{figure}

		Manually experimenting with alternative solutions using co-simulation helps exemplify the type of system the developers intend to produce. 
		Visualisation of the co-simulation can be used to convey the type of system the developers have in mind to the other project stakeholders. 
		Visualisation also allows domain experts with knowledge in the project area to provide input on the  solutions and 
		to identify any faulty assumptions that have been made by the developers. 
		Figure~\ref{fig:DSE_feed_v1} illustrates such a visualisation of the first co-model version of the mink-feeding robot, 
		where the intention was to obtain input on how and where the mink-feeding robot should dispense fodder. 
		
		Feeding mink is a high-precision task compared to that of other domesticated animals used in livestock production.
		The farmer chooses the amount of fodder each cage receives, based on personal experience and knowledge concerning demands for mink gender, age, and species.
		Based on feedback from mink-farmers, the author determined that each mink cage would receive a portion of fodder in the range of 80--300 g.
		In all mink-feeding cases, the total weight of the vehicle changes gradually throughout the feeding process.
		With vehicle fodder tanks that can transport maximum loads in the range of 500--2500 kg, a machine would theoretically be able to feed 1500--30 000 cages.
		Automatically placing the fodder in these specific areas requires an on-board localisation system that can determine the current vehicle position. 
		Examples of ground-vehicle solutions for automated mink feeding can be found in Figures~\ref{fig:minkfeedingvehicle} and \ref{fig:barcode_tag_reader}.
		
		In the case illustrated in Figure~\ref{fig:DSE_feed_v1}, it was unclear whether the feeding should be performed on the upper or lower part of the mink cages on each side of the robot vehicle. 
		The intention was to have the capacity to switch feeding between the upper and lower parts of the mink cages without  moving the feeding-arm system by using two feeding outputs for one side. 
		The feedback from the stakeholders determined that food is placed at one level for an entire row of mink cages and that the cage heights can vary between rows. 
		In addition, the feedback from the stakeholders made it clear that analysis of the different solutions to this problem was required. 
		Therefore, a sixth contribution of this thesis is:
		
		\contribution[alternative]{contribution:visualisation}{Manual analysis of alternative design solutions using DSE for a robot where different candidates exist.}
		
		The co-modelling methodology extensions defined in Section~\ref{sec:model_methodology} were used to model the system solutions. 
		This co-modelling case also clarified the need for the ability to branch into multiple modelling cases when applying the co-modelling methodology extensions. 
		This branching ability was added to the co-modelling extensions, to facilitate the exploration of multiple solutions to a problem. 
		
		\subsection{Robotic mink-feeding-arm solutions} 
		Feeding mink is a high precision task compared to that of cows and pigs, because the normal feeding area has been empirically determined to be between 0.2-0.35 m for each cage. 
		Each mink cage must be dosed with a predetermined amount of fodder, which is placed on top of the cage. 
		A co-model was therefore designed for a robot system to dispense mink food along a row of cages at predetermined locations. 
		The robot co-model was evaluated according to the overall system performance demands for the different system configurations.
		
		\begin{figure}[bth]
			\centering
			\includegraphics[width=0.98\textwidth]{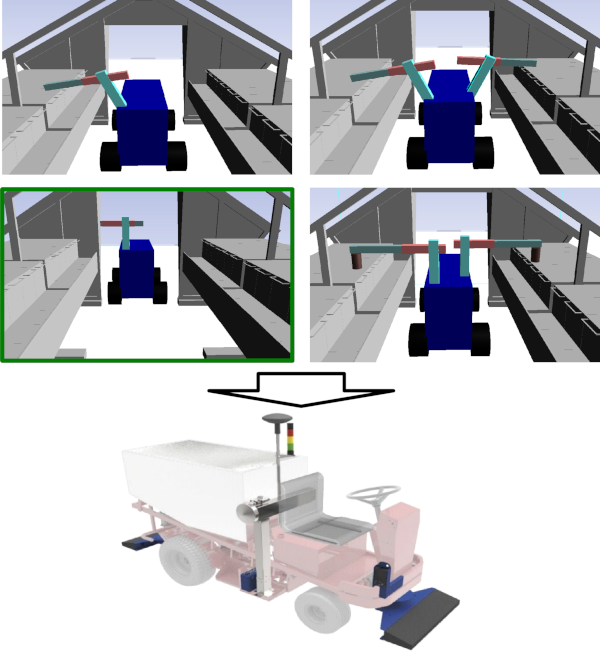}
			\caption{Feeding-arm candidate solutions experimentally examined to determine a viable candidate solution.}
		       \label{fig:DSE_feeding_arm}
		       \vspace{-10pt}
		\end{figure}
		
		We considered solutions with single- and double-sided feeding-arm outputs with two prismatic or revolute joints. 
		The goal of this analysis was to determine the most viable candidate for development into an actual system. 
		Double-sided feeding was considered based on the idea of better utilisation of the feeding robot. 
		Feeding on both sides would double the output placement of fodder at the same vehicle speed. 
		Further, shifting the arms by half a cage length would allow the same pump system to be used for both sides, while still permitting individual amounts of fodder  to be output.
		
		\subsection{Results}
		
		Each solution was modelled for evaluation based on the DSE co-simulation response. 
		The DSE functionality was used to evaluate each co-simulation robot system through collision checking and in relation to placement of the mink fodder. 
		Throughout the model development, we ran the same scenarios using DSE on the four different candidate solutions. 
		The DSE results were used to determine when the co-model was working as intended and to allow the project stakeholders to compare the solutions.

		In the final version, all four candidate solutions successfully fulfilled the stakeholder requirements.
		The four candidate solutions for the four different arm and feeding systems are illustrated in Figure~\ref{fig:DSE_feeding_arm}.
		The single-sided feeding-arm solution with prismatic joints was selected for feed-arm operation, 
		as the prismatic solution was deemed simpler to operate manually should this be necessary. 
		The single-side feeding-arm solution was chosen over a double-sided feeding-arm solution, 
		because the stakeholders deemed it to be a better first prototype design for the actual robot. 
		The double-sided solution can be easily applied as an upgrade of the same prototype platform at a later stage, 
		because no major software upgrades are required.

	\section{Derivation of a New Design Idea from the Creative Process}
	\label{sec:patent}
	
\begin{figure}[bth]%
		\begin{subfigure}[b]{0.495\textwidth}
                		\includegraphics[width=\textwidth]{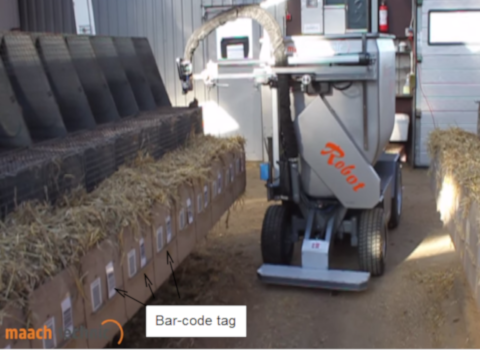}
                		\caption{Bar-code localisation solution\footnotemark.}
                		\label{fig:barcode_tag_reader}
        		\end{subfigure}
        		\hfill
        		\begin{subfigure}[b]{0.495\textwidth}
                		\includegraphics[width=\textwidth]{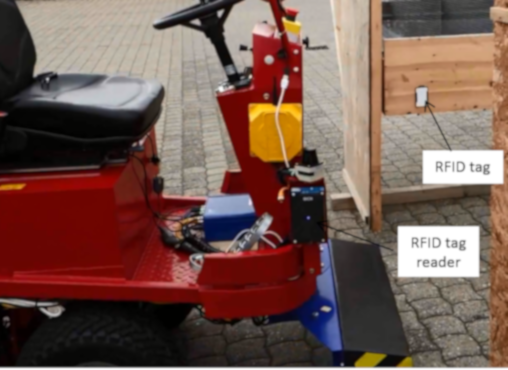}
                		\caption{RFID localisation solution.}
                		\label{fig:rfid_tag_reader}
        	\end{subfigure}
        	\caption{Example of different identification-tag-based localisation systems, which allow the robotic vehicle to automatically distribute fodder along the cage rows.}
        	\label{fig:tag_readers}
	\end{figure}

	Identification tags such as RFID have been used in the last decade to 
	provide local and global positioning information for a vehicle~\cite{Choi&09, Lin&07, Zhou&11b}.
	Examples of other identification tags that have been used for localisation are bar-codes, 
	Quick response (QR), or other visual identification code tags, such as those illustrated in Figure~\ref{fig:tag_readers}.
	In the illustrated cases, the designed vehicle was equipped with a tag reader at a known position on the vehicle.
	Identification tags with known positions were placed along the vehicle's path to provide fixed position references (landmarks). 
	Using an a priori map of the identification tag locations, the vehicle could obtain absolute positioning estimates in relation to its surroundings. 
	Positioning estimates from the identification tags were provided to the vehicle, when the tag reader was within the detection zone of each tag.
	
	\footnotetext[\value{myfootnote}]{Source: Image from Maach Technic video.}
	
	Sensory information from other sensor sources can be used to reduce the required number of identification tags.
	By combining the identification tag locations with other on-board positioning sensors, for example, 
	wheel rotary encoders and an IMU, the vehicle can continually update the current position estimate~\cite{Marin&13}. 
	The required distance between identification tags is dependent on the user-defined position accuracy and available data from other sensor sources. 
	
	\subsection{Idea motivating the invention}
	
	A vehicle with mink fodder transports a varying load over time, which affects the steering and operational performance.
	When a wheel rotary encoder is used to estimate the distance travelled, one normally assumes an a priori known effective wheel radius $R_{ee}$. 
	By measuring the number of wheel rotations using the wheel rotary encoders, the vehicle computer can provide an estimate of the distance travelled $d_e$, with
	\begin{equation}\label{eq:travel_dist}
	    d_e = 2\pi R_{ee} \frac{G_{k}}{G_{o}}
	\end{equation}
	where $G_{k}$ is the current count of the encoder and there are $G_{o}$ counts per revolution. 
	The wheel speed can be calculated using the estimate of the rotational speed
	\begin{equation}
	    u_e =  R_{ee} \omega_e \approx  R_{ee}  \frac{2\pi G_{k}}{G_{o}T_k} 
	\end{equation}
	where $\omega_e $ is the estimate of the wheel rotational speed and $T_k$ is the sample time.
	
	When operating a load-transporting vehicle, the effective radius of the wheel will vary dependent on the current load transported.
	Following equation~\ref{eq:travel_dist}, if the $R_{e}$ is compressed by 0.01 m compared to normal operation, 
	this corresponds to an estimated difference of $\approx$ 0.0628 m for each rotation, not accounting for other influencing factors. 
	The changing load conditions makes it relevant to provide a means of estimating $R_e$ online, i.e., during operation.
	Therefore, another contribution of this thesis is:
	
	\contribution[patent]{contribution:patent}{A method for online estimation of wheel and vehicle parameters for a load-carrying ground vehicle using two or more identification tag readers.}
	
\begin{figure}[tbh]
			\centering
			\includegraphics[width=0.75\textwidth]{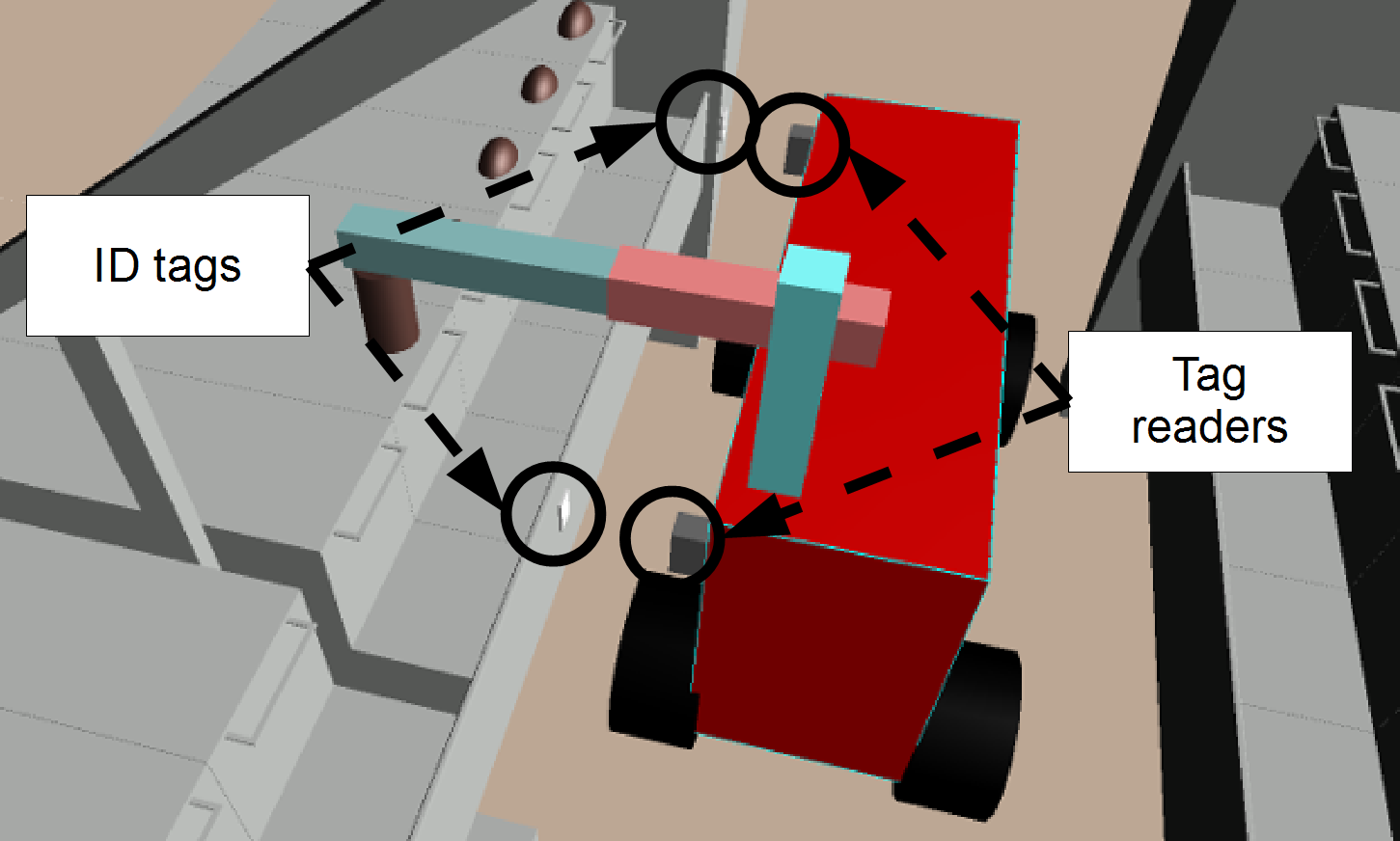}
			\caption{General invention concept. The vehicle uses two or more identification tag readers for the localisation process.}
			\label{fig:patent_idea}
		\end{figure}

	The patent application for this device is given in~\cite{Christiansen&14c}, while an introduction to the concept is given in this section.
	The concept was devised by working with virtual prototypes of the robotic mink-feeding vehicle using the co-modelling and co-simulation methodology extensions presented in Section~\ref{sec:model_methodology}. 
	The solution to the wheel rotary encoder problem was devised by working with the example from the previous section and by developing the virtual prototype solution illustrated in Figure~\ref{fig:patent_idea}.
		
	This invention is intended to improve localisation of load-changing automated ground-vehicles where the tyre experiences deformation, 
	e.g., in the case of pneumatic tyres, depending on the load distribution at a given instant.
	A vehicle for feeding mink could utilise this invention to improve localisation estimates.
	Both a fully automated version and a version driven by a human driver with automatic feeding could benefit from this invention.

	\subsection{Envisioned system configuration}
	
\begin{figure}[tbh]
			\centering
			\includegraphics[width=0.50\textwidth]{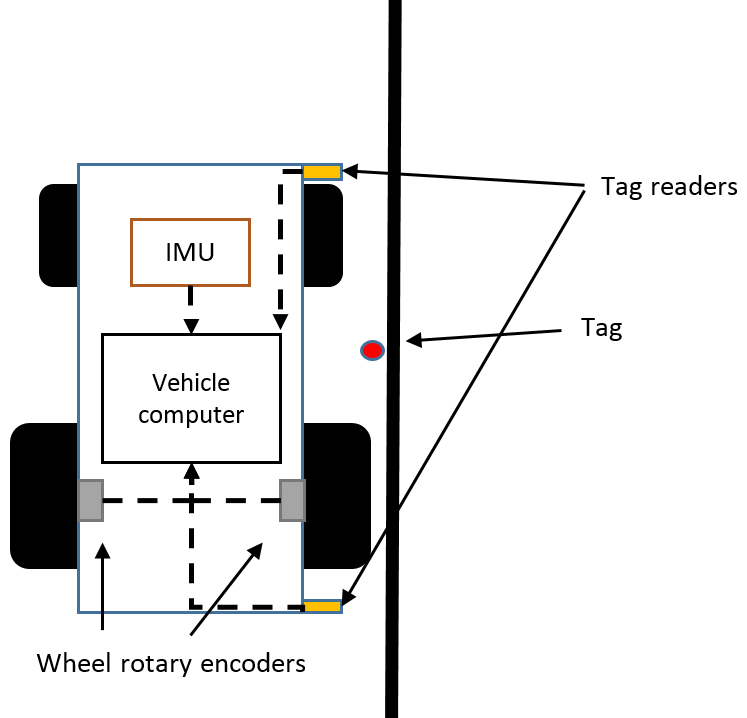}
			\caption{Sample configuration of envisioned invented system.}
			\label{fig:patent_concept_system}
		\end{figure}

	The envisioned  standard system illustrated in Figure~\ref{fig:patent_concept_system} is composed of a load-changing vehicle 
	with sensory input from two tag readers, wheel rotary encoders, and an IMU. 
	The design concept is to utilise the tag readers on the vehicle with a fixed distance in the known driving direction, 
	to provide a reference distance and time measurement so that online calibration of the wheel rotary encoder values and vehicle parameter estimation can be performed.
	Online calibration of the wheel rotary encoder values can be used to increase the distance between the identification tags. 
	By increasing the distance between the identification tags, the number of tags needed to cover the same area can be diminished.
	Each tag has the potential to be read by both tag readers, which in itself increases the number of position updates using the same number of tags.
	
	The envisioned system has two or more tag readers of the same or different types. 
	Both tag readers mounted on the vehicle must be able to read the same tag. 
	A tag could, for example contain both a bar-code string and RFID information, making it possible to utilise a combination of tag readers.
	
	The time required for the tag reader to pass the same identification tag can be used as a speed estimate source. 
	Compared to the IMU and the wheel rotary encoder sensor measurements, the tag-reader based vehicle speed estimates 
	are not based on a derived measurement in terms of wheel rotational speed or acceleration measured by the IMU. 
	The precision of the speed estimate can influence the placing of the load in the surroundings, as the systems may have a reaction time that affects the placement position.
	
	The combined information from the tag readers in combination with the IMU and the wheel rotary encoders can indicate whether each tag reader is working properly and whether the identification tags can be read. 
	If the vehicle passes a predefined number of tags but only one of the readers is detecting the tags, the operator will recieve a warning about the non-inputting tag reader. 
	Combined information from two tag readers can also be used to evaluate whether an identification tag is placed correctly according to the identification tag map. 
	In cases where both tag readers are unable to read an identification tag over multiple runs, the system could warn the operator that the identification tag needs replacement.
	
	\subsection{Detection method}

\begin{figure}[htb]%
	\centering
	\hfill
		\begin{subfigure}[b]{0.45\textwidth}
                		\includegraphics[width=\textwidth]{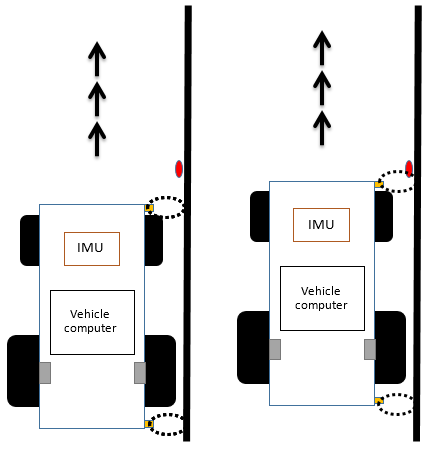}
                		\label{fig:patent_concept_in}
        		\end{subfigure}\hfill
        		\vline
        		\hspace{0.15cm}
        		\vline
        		\hfill
        		\begin{subfigure}[b]{0.45\textwidth}
                		\includegraphics[width=\textwidth]{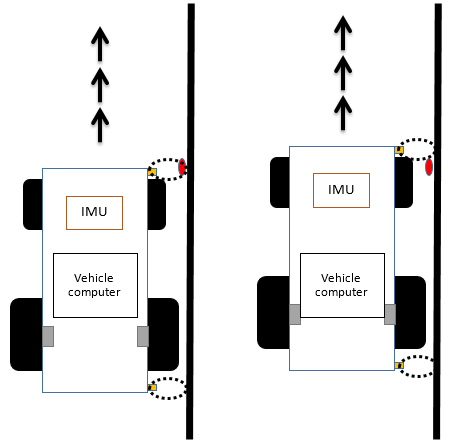}
                		\label{fig:patent_concept_out}
        	\end{subfigure}\hspace{0.25cm}
        	\caption{Method used in the calibration procedure.}
        	\label{fig:patent_concept}%
	\end{figure}

	Figure~\ref{fig:patent_concept} illustrates an example of tag detection using RFID readers. 
	The RFID reader has a detection zone in which it can receive identification information from an RFID tag, such as Identification data (ID). 
	The vehicle computer receives the RFID reader information at a specific periodic time interval.
	When there is no tag inside the RFID-reader receiving zone, the reader either transmits no data or no tag in range to the computer.
	When one of the RFID reader detection zones moves within range of the RFID tag, the computer logs a tag event.
	When a bar-code, QR, or visual tag reader is utilised, the principle remains the same in terms of detection edges, 
	even though the readers have different types of detection zones. 
	The next event occurs when the tag reader moves outside the logged tag's detection zone. 
	Both tag events are defined as specific events in time and used as time interval references when both readers have passed the tag. 
	
	The edges of the tag readers' receiving ranges can be perceived as known positions in the lateral direction of the vehicle movement.
	In terms of the vehicle movement direction, we denote the two readers as the front and rear tag-reader units.
	Using two of these points, we can define a length distance, with a total of four length distances $(d_{ii},d_{oo},d_{io},d_{oi})$ for the two tag readers.
	Here, $d_{ii}$ denotes the distance between the points for activation of the front and rear tag readers via an ingoing tag event.
	
	The movement direction determines when we should start the parameter estimation procedure. 
	Until the expected event has been triggered at the rear tag reader, the computer continues to log data from the wheel rotary encoders.
	The flow diagram for the detection procedure is documented in the patent application~\cite{Christiansen&15a} illustrated in internal Figures~6~and~7.
	When the vehicle has passed a single tag, it produces four encoder measurements and four time intervals in total, 
	which can be matched to the known length distances.

	\subsection{Parameter estimation processing example}
	
	Multiple methods can be used to estimate the relevant parameters related to the vehicle. 
	A number of these methods are mentioned below:
	\begin{itemize}
		\item Direct calculation for single sample based on equation~\ref{eq:travel_dist}
		\item Least squares estimation~\cite[Chapter 12]{Chong&08}
		\item Kalman filtering~\cite{Kalman60,Welch&06}
	\end{itemize}
	
	Based on equation~\ref{eq:travel_dist}, the least squares method can be implemented in order to estimate the current rolling wheel radius $R^{*}_{ee}$ for a single tag, where
	\begin{equation}\label{eq:leastsquares_roling}
	R_{ee}^{*} = \frac{2\pi (d_{ii}G_{ii_k}+d_{io}G_{io_k}+d_{oi}G_{oi_k}+d_{oo}G_{oo_k})}
	{(d_{ii}^2 + d_{io}^2 + d_{oi}^2 + d_{oo}^2) G_o}
	\end{equation}
	The intention here is to use the combined measurements to estimate $R^{*}_{ee}$.
	The above case is for a single encoder on a flat surface.  
	If two encoders are available, as depicted above, one can calculate the average the value for matching samples.

		\section{Exploring Controller Solutions}
		\label{sec:exp-dse}
		
		In industrial projects, components can be produced by different producers; this includes software components. 
		Parts of the software components comprising a system may be locked against modification by the external developers making the component black box~\cite{COMPASSD21.5}. 
		Locked software components pose a design and modification challenge for both farmers and external developers in the agriculture industry~\cite{Wiens15}. 
		Scenarios involving locked software components mean that it is important to explore whether alternative solutions in other parts of the system design space, might be modified to obtain the intended result. 
		Co-modelling can constitute a means of modelling these agriculture systems using approximations of the locked software components, 
		allowing developers to identify alternative solutions within the design space that they can manipulate.
		
		\subsection{Agricultural vehicles transporting loads}
		
		Load-carrying agricultural vehicles can experience load changes during operation. 
		Changes in load occur in operational tasks where animal food is dispensed, sprayer tanks are emptied, or operational implements change position over time. 
		Load changes affect the weight distribution of the vehicle and, consequently, the steering and driving performance. 
		The surface conditions the vehicle traverses also vary in response to the environment. 
		As a consequence, automated guidance controllers for such agricultural vehicles should have the ability to adapt to changes in load and surface conditions.
		
		Commercial  steering controller solutions can be locked, i.e., external developers cannot make direct changes to the software.
		Such locked software solutions render it necessary to explore solutions in other areas of the system when adjustments to the functionality need to be made.
		Therefore, another contribution of this thesis is:
		
		\contribution[DSEvehicle]{contribution:DSEvehicle}{Automated design space exploration to determine adaptive settings for a steering controller solution.}
	
		Here, the co-modelling methodology extensions defined in Section~\ref{sec:model_methodology} were used to model an auto-steering solution for an agricultural vehicle. 
		The auto-steering solution was based on an early version of method~1 described in Section~\ref{sec:case_study2}, 
		which ensures a one-to-one consistency between the co-model and the actual implementation. 
		A pre-modelled auto-steering solution was utilised instead of attempting to use measurements to develop an estimated model of a commercial auto-steering system. 
		Such a pre-modelled auto-steering solution ensures correct modelling of these system parts and allows the project to focus on the exploration of alternative system changes. 
		The auto-steering software is assumed to be a locked module, and the search for alternative solutions is therefore relevant. 
		The design concept here was to use an adaptive vehicle drive speed as the alternative solution, to facilitate auto-steering performance interaction with changing vehicle weight distribution and surface conditions.
		
		\subsection{Co-model setup}
		The co-modelled case for autonomous steering operation discussed here uses the ASuBot vehicle platform.
		The ASuBot vehicle used in the modelling is a Massey Ferguson MF 38-15SD garden tractor. 
		\begin{figure}[hbt]%
		\captionsetup[subfigure]{justification=justified,singlelinecheck=false}
		\centering
			\begin{subfigure}[t]{0.41\textwidth}
                		\includegraphics[width=\textwidth]{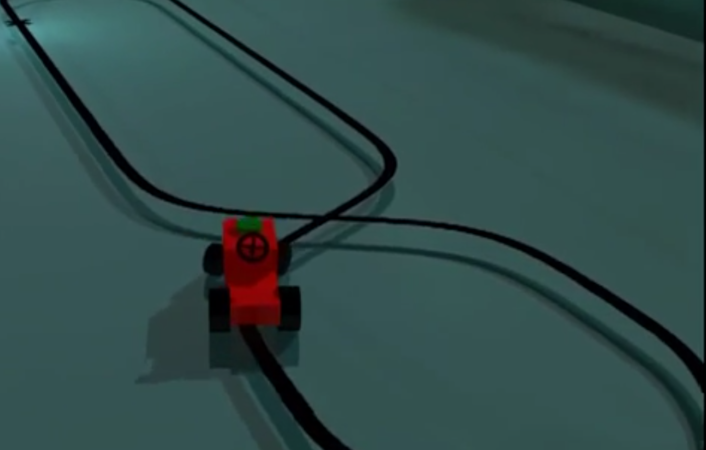}
                		\caption{Visualisation of the ASuBot co-simulation for the load-change test case.}
                		\label{fig:route_follow_sim}
        		\end{subfigure}
        		\hfill
        		\begin{subfigure}[t]{0.5\textwidth}
                		\includegraphics[width=\textwidth]{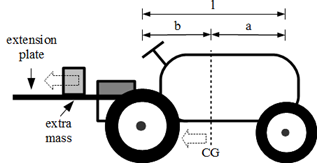}
                		\caption{Sketch of the garden tractor equipped with a load displacement mechanism controlling the load distribution between the front and rear wheels.}
                		\label{fig:load_change}
        		\end{subfigure}
        		\label{fig:load_change_dse_setup}
        		\caption{Virtual setup for the DSE search.}
		\end{figure}
		The garden tractor is also equipped with the mechanical load displacement mechanism illustrated in Figure~\ref{fig:load_change}, which enables experiments with the CG placement. 
		The co-model of the garden tractor is used to describe the different CG placement scenarios when the garden tractor is set to follow a pre-planned route. 
		The shift of the CG is intended to model load changing that affects the operation of the garden tractor. 
		The change in the load distribution is performed  in the co-model, by adjusting the values of $a$ and $b$ for the garden tractor, with
		\begin{equation}
	    		a =  a_{0}+\Delta cg
		\end{equation}
		\begin{equation}
			b =  L-(a_{0}+\Delta cg)
		\end{equation}
		where $a_{0}$ is the normal distance from the front wheel to the CG, $\Delta cg$ is the shift in the CG, and $L$ is the wheelbase.
		The current position of the CG determines the load ratio ($N_{f},N_{r}$) placed on the back and front wheels.
		For the front wheels, the load can be calculated from
		\begin{equation}
		\label{eq:front_load_calc}
		N_{f} = \frac{b}{L}N_{tot}
		\end{equation}
		where $N_{tot}$ is the total normal force.
		
		The co-model utilises the parameter $\mu$ in the CT model to describe the vehicle wheel-surface interaction in terms of wheel slip~(see Section~\ref{sec:case_study2} for a description of the differential equations).
		The change in surface condition is intended to co-simulate the garden tractor solution for different environments.
		
		The pre-planned route consists of a sequence of waypoints with relative Cartesian coordinates, which describe the path the vehicle must follow. 
		A complete route for an agricultural vehicle consists of a number of distinct segments with the possibility of repetition throughout the path. 
		All intended path scenarios are encompassed into the single looped path illustrated in Figure~\ref{fig:route_follow_sim}, so as to reduce the length of the test scenario. 
		The selected route is intended to represent a realistic scenario that the vehicle could encounter in the field. 
		The route for the project consists of straight lines in opposite parallel directions,
		with circle arc turns in the clockwise and anticlockwise directions and two lane changes in the opposite direction.

		\subsection{Automatic Co-model Analysis}
		
		The ACA implementation utilised in this project is used to select a viable candidate setup based on DSE. 
		The ACA is set to perform a DSE for a backwards shift in CG between 0 and $0.4$~m. 
		The ACA explores the drive speed $u$ in the range 1--2~m/s representing the intended operational vehicle speed area. 
		Note, that 1--2~m/s is the speed range at which the garden tractor is expected to operate within under normal conditions. 
		Throughout each co-simulation, the drive speed is kept constant, so the evaluation is not based on specific parts of the route;
		this allows an overall estimate of the steering performance for the route to be obtained.
		
		The ACA also explores tyre-surface friction $\mu$ coefficients between $0.55$ and $0.7$, 
		representing asphalt/concrete, soil, gravel, and sand conditions~\cite{Wong08}.
		A varity of surface conditions are used in order to determine their impact on the auto-steering operation performance. 
		In Crescendo, the ACA functionality is used to automatically run all the different surface conditions for the chosen drive speeds. 
		
		The DSE is used in a mode that combines automatic and manual execution. 
		In this case, the automatic mode is applied, which involves parameter sweeping of the complete drive speed range with a 0.2~m/s interval. 
		The new search drive speed range is manually selected when the border between the viable and unviable solutions is found. 
		The ACA process is then repeated with a 0.1~m/s step interval to increase the precession. 
		This process is intended to first provide a rough overall view and then to zoom in on the areas of relevance.  
		The process can be repeated for even smaller drive-speed step intervals, 
		but from experience, it was determined that 0.1~m/s was the controllable limit for this garden tractor. 
		
		To evaluate the results from the ACA runs, a cost function was used to compare the vehicle movement against the evaluation route. 
		Starting from the first waypoint in the route sequence, the direct distance between 
		the GNSS-receiver and the relative route, which is also known as XTE~\cite{DSFISODIS12188--2}, was determined. 
		The viable candidate settings were determined based on a maximum criterion of 0.3m XTE for each simulation.
		The evaluation criterion for this case study is the acceptable deviation from the route which was chosen as a sample value to demonstrate selection of viable solutions. 
		Note that, if the described ACA are to be rerun for a specific operational task intended for the garden tractor, the evaluation criteria value should be chosen to fit these demands.
		
		\subsection{Results}
		
		The ACA was run for the chosen design space to produce the output path for each co-simulation.
		\begin{figure}[htb]
	       	\centering
			\includegraphics[width=0.99\textwidth]{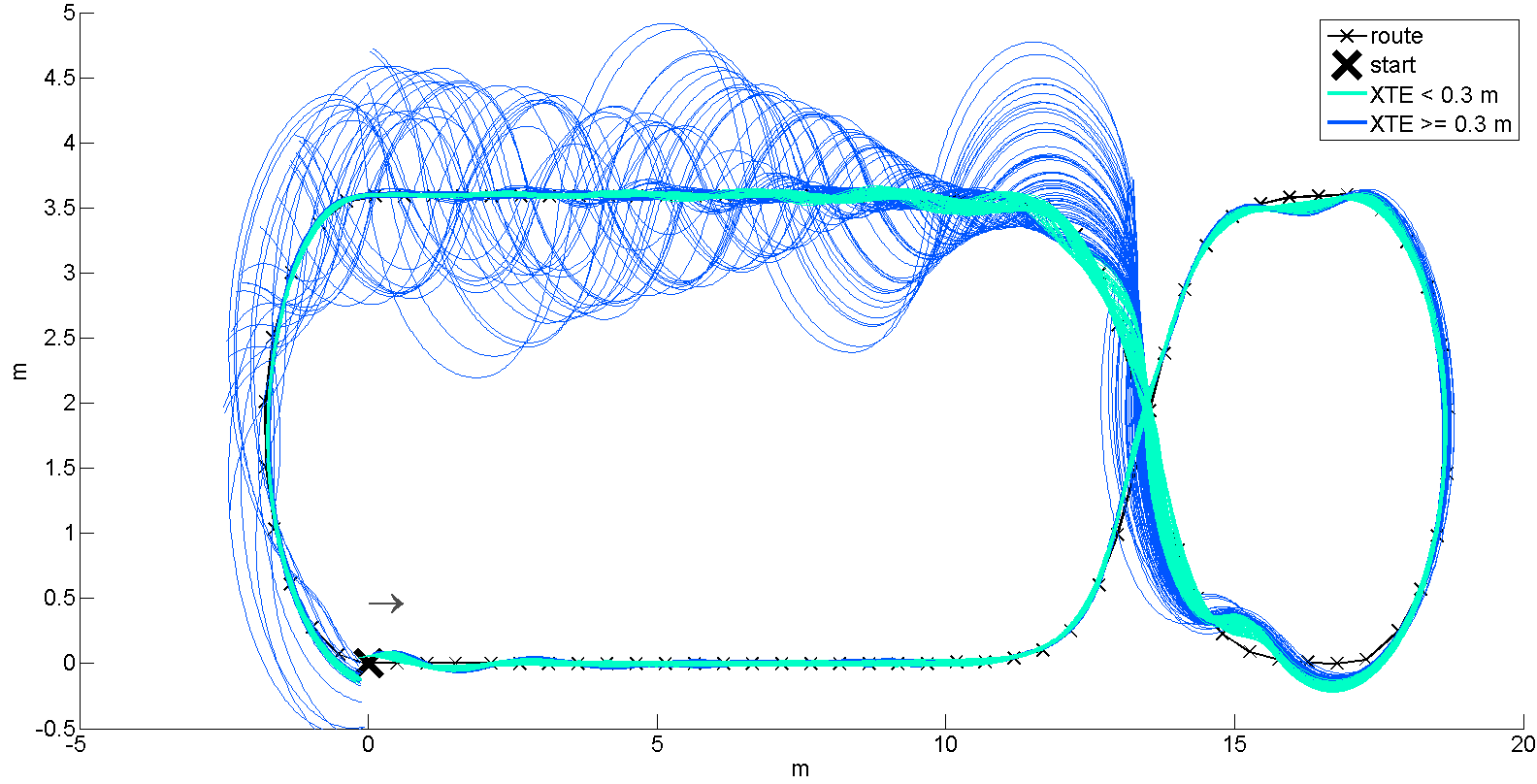}
			\caption{Simulated paths from ACA plotted in relation to the route.
			The simulation path results are coloured based on a 0.3m maximum XTE criterion.}
		 \label{fig:DSEpath}
		\end{figure}
		Each co-simulation was run until the vehicle had traversed the route or the controller was unable to steer towards any remaining route waypoints.
		The results are illustrated in Figure~\ref{fig:DSEpath} and \ref{fig:DSEspeedcon}, and the simulations are coloured based on the evaluation criteria. 
		The data in both Figure~\ref{fig:DSEpath} and \ref{fig:DSEspeedcon} are plotted for $\mu=0.55$, which represents the tyre-surface contact for wet soil.
		The automated vehicle may be unable to determine the specific surface it is traversing at any given given time;
		therefore, it is helpful to assume the vehicle is always moving on wet soil.
		
		\begin{figure}[hbt]
	        \centering
			\includegraphics[width=0.99\textwidth]{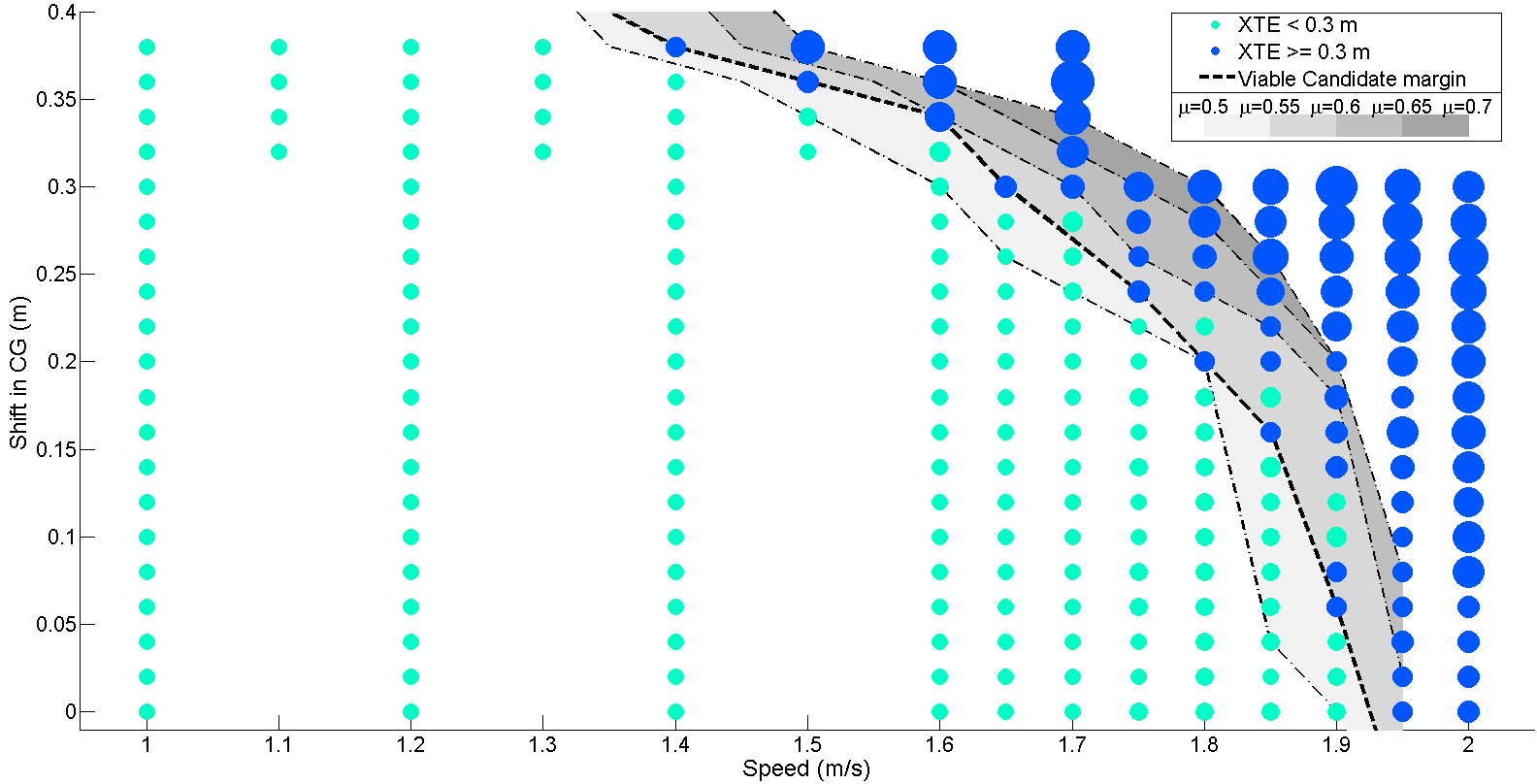}
			\caption{The full set of ACA results. Blue represents the non-viable speed setting candidates for $\mu=0.55$. 
			The safety margins are plotted for $\mu=$~0.5--0.7. 
			The $\mu=$ 0.5 scenario is plotted to illustrate the conditions outside the worst-case range.}
		 \label{fig:DSEspeedcon}
		\end{figure}
		
		The results from the ACA provide the developer with the ability to design an adaptive solution regulating the speed in relation to the current vehicle CG.
		The final runs of the  ACA produce a smaller subset of the total explored design space that indicates the margin between viable and non-viable control setting candidates.

		\subsection{Remarks on results}
		
		The results obtained here indicate that the simulation and DSE method can be used to improve the performance, 
		of classical guidance controller systems on load-changing vehicles.
		A small number of instances have been successfully tested on the actual garden tractor, but further testing is required in order to cover the full range of CG positions. 
		Future DSE analysis in this area should also incorporate a combination of different load distributions and load amounts, to encompass a broad range of scenarios.
	
		This DSE case study should be viewed as an approach to determining a solution for the circumvention of locked components of a software solution using co-modelling and co-simulation. 
		If developers had access to the locked components, they could simply implement a yaw-rate controller in the steering solution to compensate for the load-change problems. 

	\section{Obtaining a Multidisciplinary Solution Overview using DSE}
	\label{sec:over-dse}
	
	In some instances, candidate solution for the same design problem can be found in multiple disciplines. 
	An overview of these solutions should then be provided, so the stakeholders can make an informed choice between the different disciplinary solutions. 
	As an illustration of this type of analysis, the co-model in Section~\ref{sec:case_study3} was intended to give an overview of a multi-disciplinary engineering design problem.
	Here, we revisit the problem scenario from Section~\ref{sec:patent} for solutions to the tyre $R_e$ and transported load. 
	We assume that the invention solution has not been considered and other approaches are explored instead .
	Therefore, the final contribution of this thesis is:

	\contribution[DSEANA]{contribution:DSEANA}{A multi-domain design space exploration analysis has been illustrated, involving evaluation of solutions from different engineering disciplines for the same design problem.}
	
	These estimation problems require cross-disciplinary analyses, because multiple factors affect the outcomes and possible solutions can be found in different engineering disciplines. 
	Here, we demonstrate the approach to analysing the problem by modelling a load-carrying robot used for dispensing mink fodder at predetermined locations along rows of cages. 
	
	\subsection{System performance demands}

	The system performance demands define the task robot must conplete in order to be considered effective. 
	The system performance required by the project stakeholders includes the following:
	
	\begin{itemize}
		\item 	A maximum vehicle speed of 0.25 m/s (conforming to ISO-10218~\cite{ISO10218})
		\item 	No collisions with surroundings as illustrated in Figure~\ref{fig:system_setup}.
		\item 	The distance between the RFID tags $d_t$ should  be between 0.3 m and 20 m.
		\item	Feeding with a precision of $\pm$~0.08~m~inside the placement areas.
	\end{itemize}
	It should be noted that the performance requirements are non-domain-specific and focus on the overall performance of the robot. 
	Here, the maximum distance between the RFID tags represents the length of the feeding area and sets the limit for the minimum number of tags. 
	The lower limit for $d_t$ is chosen based on the mink cages length used in the co-modelling, resulting in one RFID tag for each cage.
	
	\subsection{Modelling cases}
	
		 The co-model describes the vehicle and its sensor, actuators, steering controller, feeding system, and sensor-fusion components.
		 The goal was to achieve the maximum possible distance between the RFID tags without compromising the pre-set system constraints.
		 The question here was whether the loading of the vehicle should be accounted for by reducing the maximum compression of the tyre, 
		 implementing a compensation method in the DE controller, or a combination of both. 
		 The following DE controller conditions were applied:

		\begin{description}
		\item[ $<$Static$>$] The estimated effective tyre radius was considered to be the mean of the values for the unloaded and fully loaded robot.
		This is based on the assumption that the mean value will produce the least overall error in the estimate.
	
		\item[$<$Pre-calibration$>$] A pre-measured estimate of the current rear tyre wheel radius in relation to the transported load is used in the DE part of the co-model. 
		The estimates of the effective radius were obtained through the MATLAB bridge and directly passed from 20-sim with an accuracy of~$\pm$0.001~m.
	
		\item[$<$Estimator$>$] The input data obtained by the vision sensor were used to estimate the current effective radius before entering the feeding area.
		This estimate was based on the distance travelled between the updates, with an accuracy of~-0.005~m.
		\end{description}
		
		Rather than simulating a single scenario, the test set shown in Table~\ref{tab:DSE_test_set} is designed to evaluate the expected min-mean-max operational values. 
		The DSE was used to evaluate the configuration solutions shown in Table~\ref{tab:DSE_test_set} in different development domains, so as to account for the load-carrying effects. 
		The operational values represent the expected range of transported load values and the surface-wheel and initial robot position conditions.
		The initial position was of interest in this case because a human operator may inaccurately place the robot at its starting point.
		The models of the tyre radius on the CT side were varied between low and fully loaded conditions. 	
		The tyre-surface friction was of interest here as the vehicle must stop in order to deploy the feeding arm before beginning the feeding process.
	
		\begin{table}[tbh]
			\centering
			\caption{Candidate solution sets used for the system evaluation and min-mean-max test set used for the DSE of the feeding robot.}
			\resizebox{\textwidth}{!}{%
			\begin{tabular}{|c|c||c|c|c|}
			  \hline
			  \multicolumn{2}{|c||}{\textbf{\textit{System configurations}}} & \multicolumn{3}{|c|}{\textbf{\textit{Min-mean-max test set}}} \\
			  \hline
			  \textbf{Rear tyre }  &  \textbf{Vehicle} & \textbf{Load mass}  &  \textbf{Surface-tyre} & \textbf{Initial Position}\\
			  \textbf{radius change} & \textbf{state estimate} & & $\mu$ friction & \textbf{$x_{init}, y_{init}, \psi_{init}$}\\  \hline
			  0.001~m & \textbf{$<$Static$>$} & 1\% (6~kg) & 0.3 & $x_{init} = \{$-0.5~m,0~m,0.5~m$\}$ \\
			  0.02~m & \textbf{$<$Pre-calibration$>$} & 50\% (300~kg) & 0.5 & $y_{init} =  \{$-0.1~m,0~m,0.1~m$\}$ \\
			  0.04~m & \textbf{$<$Estimator$>$} & 100\% (600~kg) & 0.7 & $\psi _{init}= \{-15^{\circ},0^{\circ},15^{\circ}\}$\\
			  \hline
			\end{tabular}}
			\label{tab:DSE_test_set}
		\end{table}

	\subsection{Automatic co-model analysis}
	
	To select the value of $d_t$ parameter that for the ACA co-simulation, an output cost-function is defined.
	The result of each co-simulation is evaluated based on the success-rate in placing the fodder at the correct positions between two tags.
	
	\begin{equation}
  		\label{eq:costfunc}
   		 f_{d_t} = -\frac{b_{suc}^2}{b_{tot}}\\
	\end{equation}
	where $ b_{tot}$ is total placement positions between two RFID-tags and $b_{suc}$ is the number of successfully fodder placements.
	The output of the cost-function ensures largest $d_t$ with the highest number of successful fodder placements is the minima for the searchable range. 
	In mathematics, by convention optimization problems are usually stated in terms of minimization, thus the minus sign.
	ACA uses the golden section search method~\cite[Chapter 7]{Chong&08} in combination with the cost-function in Equation~\eqref{eq:costfunc}, to determine the best candidate within the design space.
	To use golder-section search it is assumed that the cost-function is unimodal function, meaning that there is only a single local minimum.
	
	\subsection{Results}
\begin{figure}[th]
		\centering
         	  	\includegraphics[width=0.87\textwidth]{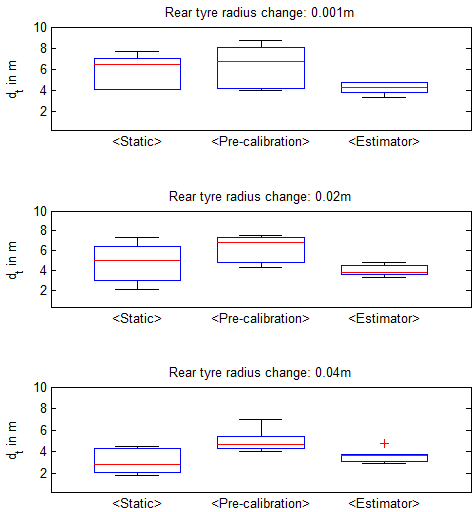}
         	  \caption{Result of the ACA run for the feeding farming co-model in terms of determined $d_t$, for the nine different system configurations in Table~\ref{tab:DSE_test_set}.}
           \label{fig:result}
	\end{figure}

	The result of the ACA is illustrated in Figure~\ref{fig:result} using boxplots.
	In each box-plot, the central line marks the median, the edges of the box are the 25th and 75th percentiles and the whiskers marks to the two most extreme data points. 
	Each system configurations box-plot, represents the determined max RFID $d_t$ distance values, for each instance in the min-mean-max test set from Table~\ref{tab:DSE_test_set}. 

	\subsection{Discussion of results}
	\label{sec:dis}
	
	The results provide an overview of the candidate system configurations based on the estimated RFID tag distance. 
	Developers can use the candidate overview to select configurations for testing on the an actual platform.
	The intention here is to provide the stakeholders in the project with an overview of the different candidate solutions. 

	From the box plots, it can be seen that $<$Pre-calibration$>$ method provides the best overall results for all tyre solutions. 
	This is to be expected since the value $R_{ee}$ used matches reality with a high degree of accuracy.
	The candidate with the best results in terms of largest overall $d_t$ for all co-simulation cases is not necessarily the one that will be chosen for implementation on the actual robot.
	Factors such as material, development, implementation, and maintenance costs affect the final configuration selection. 
	The 0.001~m tyre compression solution requires adjustment by an operator before start-up. 
	The pre-calibrated solution must be updated periodically to account for changes in the robot setup. 
	The vision solution is calibrated for a specific set of farm configurations and requires adjustments for new conditions. 
	Nevertheless, this overview provides a means of evaluating the external costs with respect to the expected 
	distance between the RFID tag  and affords a more educated configuration selection.

	Note, that the co-model can be reused to explore other aspects of this robotic system, such as the invention described in Section~\ref{sec:patent}.
	The time saved by the co-modelling and ACA could be invested in other areas of the project. 
	The overview obtained by ACA does not guarantee optimal solutions, 
	but it does facilitate the analysis of multiple candidate solutions.

\section{Summary}
	This chapter has presented four different search approaches to DSE using co-modelling and co-simulation. 
	The results show how DSE can be used for both automated and manual exploration of a design space. 
	The four different approaches illustrate that DSE can be used for virtual prototype development support and the determination of candidate solutions.

\chapter{Concluding Remarks}
\label{chap:con}
	This chapter summarises the results achieved in this thesis and presents the conclusions to the research hypothesis.
	In this thesis, a model-based approach to the development of automated and robotic agricultural ground vehicles was proposed.
	The hypothesis and objectives of the thesis defined in Chapter 1 are related to the chapters on automated 
	and robotic ground-vehicle co-modelling (Chapter 2) and Design Space Exploration (Chapter 3). 
	
	The purpose of this chapter is to evaluate the outcome of the thesis, 
	and to assess the extent to which the evaluation criteria, objectives, and hypothesis have been met.
	Section~\ref{sec:contributions} summarises the research contributions; this is
	followed by an evaluation of research contributions in Section~\ref{sec:evaluation}
	and an assessment of how the contributions have fulfilled 
	the PhD project hypothesis  which is given in Section~\ref{sec:hypothesis}.
	Finally, future work is described and presented in Section~\ref{sec:future}. 
	Some of these areas will be addressed in the follow-up research project INTO-CPS\footnote{The INTO-CPS project website can be found at: \url{http://into-cps.au.dk/}.}, 
	which will be conducted using industrial case studies.
	Other topics discussed in the future work section are also related to a possible future postdoctoral research project.
	
	\section{Research Contributions}
	\label{sec:contributions}
	
\begin{figure}[hbt]
	      \centering
			\includegraphics[width=0.855\textwidth]{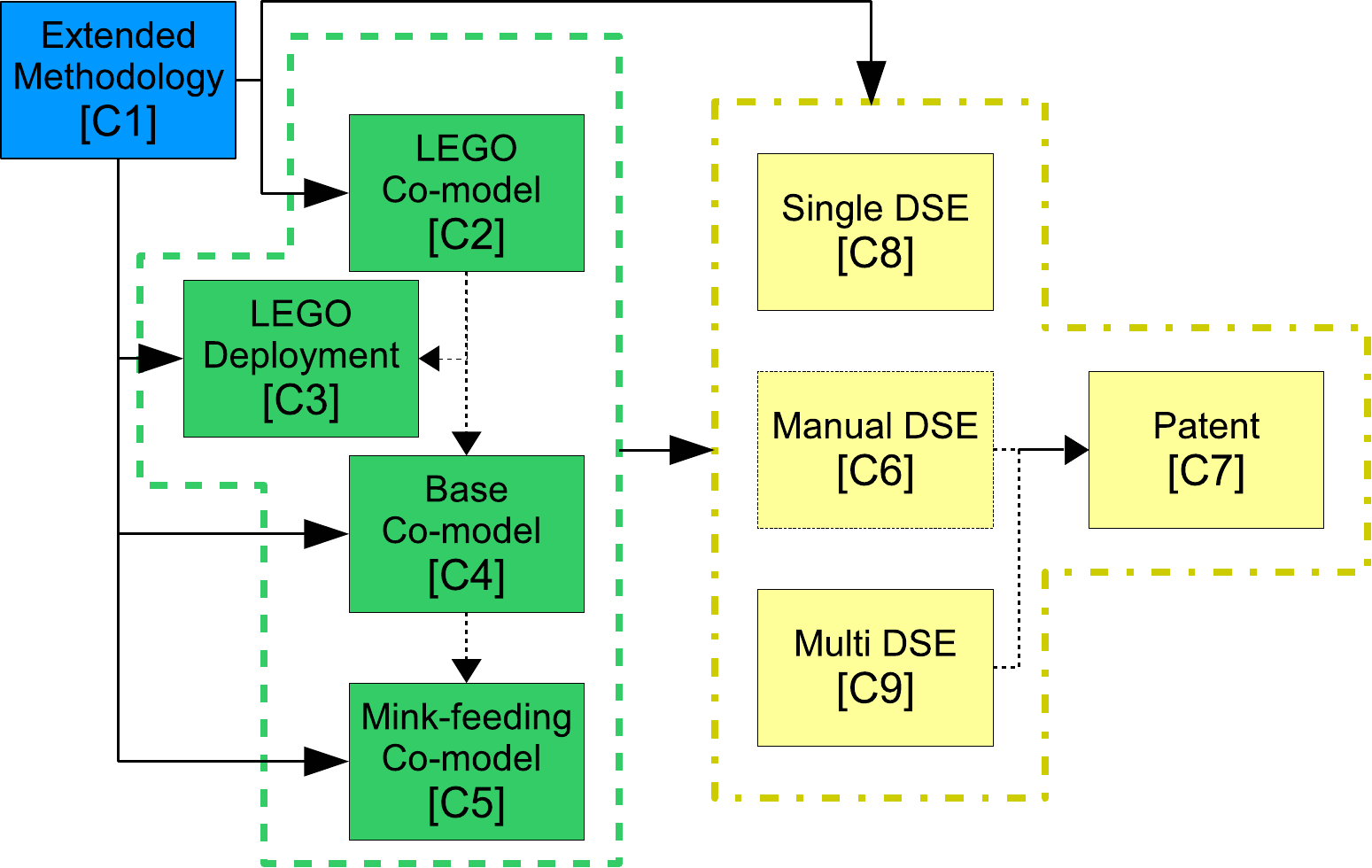}
			\caption{Overview of the research contributions presented in this PhD thesis 
				and how their categorisation in relation to each other.}
		\label{fig:Contribution-diagram}
		\end{figure}

	This PhD thesis has presented nine research contributions in the previous chapters. 
	The research contributions are collected into three different categories: 
	\textit{Methodological extensions}, \textit{Co-modelling of automated and robotic ground vehicles} 
	and \textit{Methods to perform Design Space Exploration}. 
	An overview of the contributions and the relations between them is provided in Figure~\ref{fig:Contribution-diagram}. 
	The contributions have been given individual short-form names for easier identification~(C1, C2, C3, etc.). 
	The methodological extensions represented using Contribution 1 (blue block) 
	have been applied in the modelling of different automated and robotics agricultural ground-vehicle case studies. 
	The contributions that are related to the implementation of co-modelling of various 
	automated and robotic agricultural ground-vehicle systems (using green blocks) are categorised by the green dashed border marking. 
	The green block contributions comprise 2 to 5. 
	The remaining contributions (6 to 9) which are related to DSE  (using yellow blocks), are categorised by the light yellow dot-dashed border marking. 
	Individual contributions that are interconnected are marked with dashed arrows, 
	to clarify their influences upon each other during their derivation. 

	The interconnection between the automated or robotic agricultural ground-vehicle 
	modelling cases illustrates the gradual expansion of the co-modelling of these systems. 
	The invention concept was derived through research related to contributions 7 and 9.
	
	\subsection{Methodological extensions}
	
	The literature survey of co-simulation in the agriculture domain presented in Section~\ref{sec:lit_model} 
	revealed the lack of a defined development methodology for this area of research. 
	The base development methodologies from \cite{Wolff13a,Fitzgerald&14c} focused on model-based development of embedded and cyber physical systems. 
	This constituted the first contribution, which was formulated as:
	
	\begin{framed}
	\printcontribution{1}
	\end{framed}
	
	Section~\ref{sec:model_methodology} introduced the extended model-based methodology for the development of automated and robotics agricultural 
	ground vehicles, which was developed based on the three presented case studies. 
	
	\subsection{Co-modelling of automated and robotic ground vehicles}
	
	The second group of contributions are related to modelling case studies, which have contributed to the definition of the 
	current extended development methodology for the model-based design of automated and robotic agricultural ground-vehicles. 
	The presented development methodology was gradually extended based on the experience 
	acquired from the co-modelling and co-simulation of these main case studies. 
	The LEGO tractor case study used for field coverage testing was used to refine the co-model development process and the gradual deployment of a co-model. 
	That case study led to the following contributions:

	\begin{framed}
	\printcontribution{2}
	\end{framed}

	\begin{framed}
	\printcontribution{3}
	\end{framed}

	An extended development methodology complemented by a base agricultural co-model was developed to support future model-based development. 
	This base automated agricultural vehicle co-model was presented in Section~\ref{sec:case_study2}, which described the aspects of both DE and CT modelling. 
	This contribution will potentially support future mobile development of solutions based on axle steered ground vehicles:

	\begin{framed}
	\printcontribution{4}
	\end{framed}

	We illustrated how the base co-model from contribution~4, can be extended for a specific purpose using the extended development methodology. 
	This case study involved development for a mink-feeding robot vehicle, and was used for sensor-fusion-based localisation and analysis of load-distribution dynamics. 
	The details of this study are described in Section~\ref{sec:case_study3}.
	The contribution of this case study is summarised as:
	
	\begin{framed}
	\printcontribution{5}
	\end{framed}
	
	\subsection{Methods to perform Design Space Exploration}
	
	DSE was applied in two agricultural cases, resulting in four contributions to this research domain. 
	The first two contributions relate to general development using the extended development methodology 
	and the potential results this can produce, as presented in Sections~\ref{sec:visualisation}~and~\ref{sec:patent}:
	\begin{framed}
	\printcontribution{6}
	\end{framed}
	\begin{framed}
	\printcontribution{7}
	\end{framed}
	To explore candidate solutions in a design space in a more automated manner using co-modelling and co-simulation, 
	the ACA functionality in Crescendo was used and extended, yielding the two contributions described in Sections~\ref{sec:exp-dse}~and~\ref{sec:over-dse}:
	\begin{framed}
	\printcontribution{8}
	\end{framed}
	\vspace{-5pt}
	\begin{framed}
	\printcontribution{9}
	\end{framed}

	Three of the contributions [C6, C7, C9] are related to different aspects of the system used in 
	the mink-feeding agricultural case study. 
	Throughout this PhD thesis, the extended development  methodology has been used to develop these co-models for the performance of DSE.
	The robotic mink-feeding case study is a case that is currently being implemented in the industry for commercial use. 
	The remaining contribution [C8] focuses on making changes to already implemented solutions in the agricultural industry and is of a more general nature.
	This contribution illustrates how DSE and ACA can be used to address development problems where the existing system contain locked software components.
	
	\section{Evaluation of Contributions}
	\label{sec:evaluation}
	
	In this section, the contributions described in Chapters~\ref{chap:agro} and~\ref{chap:dse} are evaluated with respect to the evaluation criteria listed in Section~\ref{sec:evalcrit}. 
	Evaluation of the industrial adoption of  the work performed in this thesis is outside the PhD research scope; 
	however, contributions [C4, C5, C6, and C7], developed during this PhD project,
	have been adopted in one form or another by industrial partners during the course of this research.

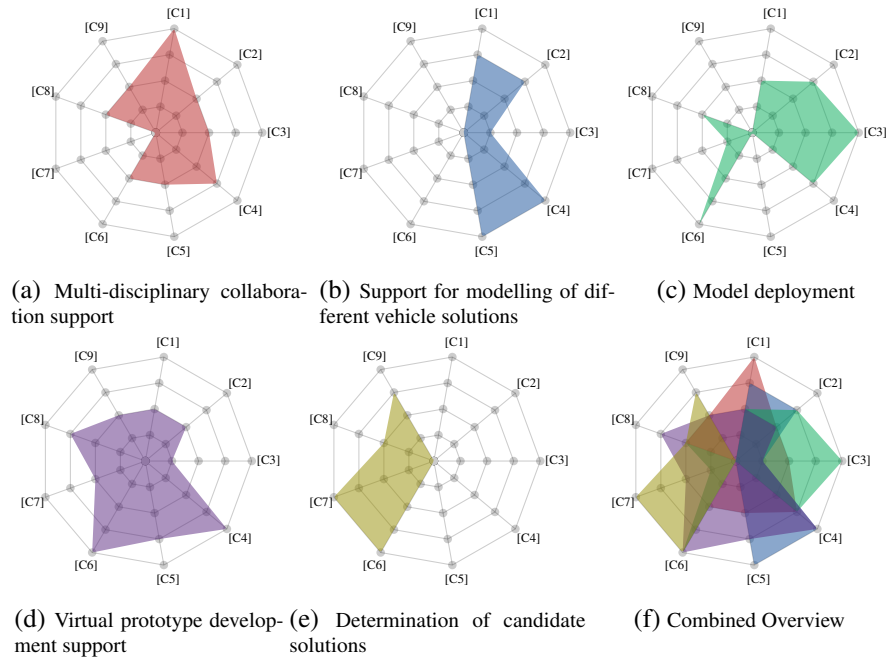
\begin{figure}[tb]
		\centering
%

\usetikzlibrary{shapes}
\usetikzlibrary{positioning,fit,calc}

\ifdefined\cref\else\def\cref#1{[C1]}\fi
\newcommand{\D}{9} 
\newcommand{\U}{4} 

\newdimen\R 
\R=3.5cm 
\newdimen\L 
\L=4cm

\newcommand{\A}{-360/\D} 
\newcommand{\AO}{\A + 120} 
 
\def\labelformat#1{\tiny #1}

\def\makeweb{
\tikzset{spiderwebpath/.style={opacity=0.2}}
\tikzset{measurepath/.style={ultra thick,opacity=0.5,rounded corners=1pt}}
\tikzset{mcol1/.style={color={rgb:red,237;green,45;blue,46}}}
\tikzset{mcol2/.style={color={rgb:red,0;green,140;blue,71}}}
\tikzset{mcol3/.style={color={rgb:red,24;green,89;blue,169}}}
\tikzset{mcol4/.style={color={rgb:red,102;green,44;blue,145}}}
\tikzset{mcol5/.style={color=olive}}
  \path (0:0cm) coordinate (O); 

  \foreach \X in {1,...,\D}{
    \draw[spiderwebpath] (\X*\A:0) -- (\X*\A:\R);
  }

  \foreach \Y in {0,...,\U}{
    \foreach \X in {1,...,\D}{
      \path (\X*\AO :\Y*\R/\U) coordinate (D\X-\Y);
      \fill[spiderwebpath] (D\X-\Y) circle (4pt);
    }

    \draw [spiderwebpath] (0:\Y*\R/\U) \foreach \X in {1,...,\D}{
        -- (\X*\A :\Y*\R/\U)
    } -- cycle;
  }
     \fill[white] (0,0) circle (4.03pt);
     \fill[spiderwebpath] (0,0) circle (4pt);

  
  \path (1*\AO :\L) node (L1) {\labelformat{ \cref{contribution:methodology}}};
  \path (2*\AO :\L) node (L2) {\labelformat{ \cref{contribution:legosim}}};
  \path (3*\AO :\L) node (L3) {\labelformat{ \cref{contribution:legodeploy}}};
  \path (4*\AO :\L) node (L4) {\labelformat{ \cref{contribution:modelcase2}}};
  \path (5*\AO :\L) node (L5) {\labelformat{ \cref{contribution:feeding}}};
  \path (6*\AO :\L) node (L8) {\labelformat{ \cref{contribution:visualisation}}};
  \path (7*\AO :\L) node (L6) {\labelformat{ \cref{contribution:patent}}};
  \path (8*\AO :\L) node (L9) {\labelformat{ \cref{contribution:DSEvehicle}}};
  \path (9*\AO :\L) node (L7) {\labelformat{ \cref{contribution:DSEANA}}};
}

   \def\labelformat#1{\large #1}
   \def\subfixspiderscale{0.4}

\def\makeContributionCompareWeb{
\begin{subfigure}[t]{0.33\textwidth}
\centering
\begin{tikzpicture}[scale=\subfixspiderscale, transform shape]
\makeweb
  \fill [mcol1,measurepath]
    (D1-4) -- 
    (D2-2) --
    (D3-2) --
    (D4-3) --
    (D5-2) --
    (D6-2) --
    (D7-0) -- 
    (D8-2) -- 
    (D9-2) --  
   cycle;
\end{tikzpicture}
\caption{\scriptsize Multi-disciplinary collaboration support}
\label{fig:multidisciplinary}
\end{subfigure}~
\begin{subfigure}[t]{0.33\textwidth}
\centering
\begin{tikzpicture}[scale=\subfixspiderscale, transform shape]
\makeweb
  \fill [mcol3,measurepath]
    (D1-3) --
    (D2-3) --
    (D3-1) --
    (D4-4) --
    (D5-4) -- 
    (D6-0) --
    (D7-0) -- 
    (D8-0) -- 
    (D9-0) -- 
   cycle;
\end{tikzpicture}
\caption{\scriptsize Support for modelling of different vehicle solutions}
\label{fig:contrib:modelling}
\end{subfigure}
\begin{subfigure}[t]{0.3\textwidth}
\centering
\begin{tikzpicture}[scale=\subfixspiderscale, transform shape]
\makeweb
  \fill [mcol2,measurepath]
    (D1-2) --
    (D2-3) --
    (D3-4) --
    (D4-3) --
    (D5-0) --
    (D6-4) --
    (D7-1) -- 
    (D8-2) -- 
    (D9-0) -- 
    cycle;
\end{tikzpicture}
\caption{\scriptsize Model deployment}
\label{fig:deployment}
\end{subfigure}
\begin{subfigure}[t]{0.3\textwidth}
\centering
\begin{tikzpicture}[scale=\subfixspiderscale, transform shape]
\makeweb
  \fill [mcol4,measurepath]
    (D1-2) --
    (D2-2) --
    (D3-1) --
    (D4-4) --
    (D5-3) -- 
    (D6-4) --
    (D7-2) -- 
    (D8-3) -- 
    (D9-2) -- 
   cycle;
\end{tikzpicture}
\caption{\scriptsize Virtual prototype development support}
\label{fig:contrib:support}
\end{subfigure}
\begin{subfigure}[t]{0.33\textwidth}
\centering
\begin{tikzpicture}[scale=\subfixspiderscale, transform shape]
\makeweb
  \fill [mcol5,measurepath]
    (D1-0) --
    (D2-0) --
    (D3-0) --
    (D4-0) --
    (D5-0) --
    (D6-4) --
    (D7-4) -- 
    (D8-2) -- 
    (D9-3) -- 
   cycle;
\end{tikzpicture}
\caption{\scriptsize Determination of candidate solutions}
\label{fig:contrib:solution}
\end{subfigure}
%
%
%
\begin{subfigure}[t]{0.33\textwidth}
\centering
\begin{tikzpicture}[scale=\subfixspiderscale, transform shape]
\makeweb
 \fill [mcol1,measurepath]
    (D1-4) -- 
    (D2-2) --
    (D3-2) --
    (D4-3) --
    (D5-2) --
    (D6-2) --
    (D7-0) -- 
    (D8-2) -- 
    (D9-2) --  
   cycle;
  \fill [mcol3,measurepath]
    (D1-3) --
    (D2-3) --
    (D3-1) --
    (D4-4) --
    (D5-4) -- 
    (D6-0) --
    (D7-0) -- 
    (D8-0) -- 
    (D9-0) -- 
   cycle;
  \fill [mcol2,measurepath]
    (D1-2) --
    (D2-3) --
    (D3-4) --
    (D4-3) --
    (D5-0) --
    (D6-4) --
    (D7-1) -- 
    (D8-2) -- 
    (D9-0) -- 
    cycle;
  \fill [mcol4,measurepath]
    (D1-2) --
    (D2-2) --
    (D3-1) --
    (D4-4) --
    (D5-3) -- 
    (D6-4) --
    (D7-2) -- 
    (D8-3) -- 
    (D9-2) -- 
   cycle;
  \fill [mcol5,measurepath]
    (D1-0) --
    (D2-0) --
    (D3-0) --
    (D4-0) --
    (D5-0) --
    (D6-4) --
    (D7-4) -- 
    (D8-2) -- 
    (D9-3) -- 
   cycle;
\end{tikzpicture}
\caption{\scriptsize Combined Overview}
\label{fig:contrib:combined-overview}
\end{subfigure}
}

\def\makeHypothesisCompareWeb{
\begin{subfigure}[t]{0.49\textwidth}
\centering
\begin{tikzpicture}[scale=0.6, transform shape]
\makeweb
  \fill [mcol1,measurepath]
    (D1-4) -- 
    (D2-2) --
    (D3-2) --
    (D4-3) --
    (D5-2) --
    (D6-2) --
    (D7-0) -- 
    (D8-2) -- 
    (D9-2) --  
   cycle;
  \fill [mcol3,measurepath]
    (D1-3) --
    (D2-3) --
    (D3-1) --
    (D4-4) --
    (D5-4) -- 
    (D6-0) --
    (D7-0) -- 
    (D8-0) -- 
    (D9-0) -- 
   cycle;
  \fill [mcol2,measurepath]
    (D1-2) --
    (D2-3) --
    (D3-4) --
    (D4-3) --
    (D5-0) --
    (D6-4) --
    (D7-1) -- 
    (D8-2) -- 
    (D9-0) -- 
    cycle;
\end{tikzpicture}
\caption{\scriptsize Hypothesis - part 1}
\label{fig:hypo1_eval}
\end{subfigure}
\begin{subfigure}[t]{0.49\textwidth}
\centering
\begin{tikzpicture}[scale=0.6, transform shape]
\makeweb
    \fill [mcol2,measurepath]
    (D1-2) --
    (D2-3) --
    (D3-4) --
    (D4-3) --
    (D5-0) --
    (D6-4) --
    (D7-1) -- 
    (D8-2) -- 
    (D9-0) -- 
    cycle;
  \fill [mcol4,measurepath]
    (D1-2) --
    (D2-2) --
    (D3-1) --
    (D4-4) --
    (D5-3) -- 
    (D6-4) --
    (D7-2) -- 
    (D8-3) -- 
    (D9-2) -- 
   cycle;
  \fill [mcol5,measurepath]
    (D1-0) --
    (D2-0) --
    (D3-0) --
    (D4-0) --
    (D5-0) --
    (D6-4) --
    (D7-4) -- 
    (D8-2) -- 
    (D9-3) -- 
   cycle;
\end{tikzpicture}
\caption{\scriptsize Hypothesis - part 2}
\label{fig:hypo2_eval}
\end{subfigure}
}
		\makeContributionCompareWeb
		\caption{Relation between contributions and evaluation criteria.}
		\label{fig:contributions}
		\vspace{-8pt}
	\end{figure}

	This evaluation is performed in terms of the different dimensions introduced in Section~\ref{sec:evalcrit}, and visually presented in Figure~\ref{fig:contributions}. 
	The figure illustrates an informal ranking of the contributions providing an overview of how the individual contributions fulfill each criteria. 
	The 0--4 scale used in the figure indicates the extent to which the contributions fulfil each of the evaluation criteria considered in the subfigures.
	The closer the shading comes to the edge of the spider-web, the greater the fulfilment of the given criterion.
	
	An evaluation of all contributions is given below, with respect to the evaluation criteria and the individual gradings.
	Subfigure~\ref{fig:contributions}f illustrates the extent to which all the contributions add to the overall fulfilment of the evaluation criteria.
	
	\subsection{Multi-disciplinary collaboration support}
	To assist stakeholders addressing automated or robotic agricultural ground-vehicle multi-disciplinary 
	design and to facilitate intercommunication between the engineering disciplines, 
	an extended development methodology and a number of development cases have been created. 
	This methodology, along with its application in the case studies and the analysis of co-models supports a common 
	understanding between the disciplines and provides stakeholders with the ability to grasp concepts that are not inherent to their respective disciplines. 
	Contribution 1, which covers the extended development methodology, is ranked as fully satisfying the multi-disciplinary collaboration support criterion, 
	as it can be used for any vehicle type in the agricultural domain. 
	The extended development methodology provide support for development in all phases of the project, from initiation to model deployment. 
	Both contributions 2 and 3 support collaboration in combination with the extended development methodology, for projects on automated field coverage. 
	
	The base co-model for front- and back-axle-steered vehicles (contribution~4), in combination with the development methodology guidelines, 
	provides general coverage support for development for projects based on such vehicle solutions. 
	The extension made using contribution 5 is the more specialised case involving the mink-feeding system, 
	but some aspects could be adapted to provide support for other related projects. 
	Further, the three contributions related to the DSE part (contributions 6, 8, and 9) of this PhD thesis provide multi-disciplinary support 
	by giving an overview of the stakeholders' current design space. 
	Note that contribution 8 is given a low ranking here since the determined solution requires further testing establish its the full validity.
	
	\subsection{Model Deployment}
	
	Model deployment grades the contributions in terms of deployment of the components and results in the real-world systems. 
	The intent is to verify that a solution developed using modelling and co-simulation applies to an actual system setting. 
	The extended development methodology (contribution 1) has been awarded an impact level of two,
	as this contribution provides the guidelines for the deployment of co-model components.  
	Both contributions 2 and 3 relate to the LEGO micro-tractor case used for testing field coverage operation. 
	Contribution 3 has been awarded a higher ranking than Contribution 2 because full deployment to the application area was achieved here. 
	The path-tracking methods from contribution 4 have been implemented on the AsuBot vehicle and the FixRobo mink-feeding robot from Conpleks Innovation. 
	However, contribution 4 must also be deployed for a back-axle-steered vehicle to receive the maximum grading.

	Full coverage is achieved by contribution 6, where the robotic mink-feeding arm was implemented in a real-vehicle solution. 
	Other feeding arm solutions from the DSE could also be deployed to the FixRobo vehicle solution, 
	but the deployment procedure would be identical and would not involve new aspects of the system. 
	The patent application of contribution 7, is awarded a ranking of one, as this solution was acquired by 
	Conpleks Innovation, but still requires implementation in a product solution. 
	
	\subsection{Determination of candidate solutions}
	
	Determining a candidate solution is related to the DSE and the use of co-models to determine solutions that can solve a design case.
	The intention is to determine prototype and parameter solutions for the automated or robotic agricultural ground-vehicles, on which the developers are working.
	Contributions 6 and 7 have the highest ranking in this area, as they provided a solution that could be ported to an actual industrial case. 
	Contribution 8 has been awarded an average ranking for this evaluation criterion as the DE model has been deployed to the vehicle, 
	but further testing needs to be performed to validate the solution fully. 
	As regards contribution 9, we obtain an overview of the solutions only, and it is the task of the stakeholders to select the candidate solution they wish to implement. 
	
	\subsection{Support for modelling of different vehicle solutions}
	
	Contribution 1 can be used when developing co-models of different vehicle types that can be used to develop automated and robotic ground-vehicle solutions. 
	Contributions 2 and 3 are related to co-models of vehicle solutions used for field coverage in the agricultural industry. 
	Contribution 4 is a base vehicle model for front- and back-axle-steered ground-vehicle solutions, 
	which can be adapted for new model-based development cases in combination with the extended development methodology. 
	Contribution 5 concerns specialised co-modelling of a mink-feeding robotic ground vehicle to support system development.
	This mink-feeding robot co-model models the full operational aspects of the vehicle when it is operating within the mink farm.
	
	\subsection{Virtual prototype development support}
	
	 Virtual prototyping is the use of co-modelling, co-simulation, and visualisation of automated or robotic agricultural ground-vehicle solutions. 
	 To develop these virtual prototypes, we use the extended development methodology that comprises contribution 1, which must be combined with a concrete case study. 
	 The LEGO tractor co-model in contributions 2 and 3 could be used to explore alternative mechanical solutions for LEGO-based systems. 
	 However, it is concluded that directly reassembling the LEGO tractor is more appropriate and is, in itself, a form of vehicle solution prototype. 
	 
	 The base agricultural co-model in contribution 4 can be used to develop these prototypes conjunction with the extended development methodology. 
	 The base co-model has been used for virtual prototyping in contribution 6 and  for the evaluation of the feeding arm and controller solutions respectively. 
	 Contribution 5 can primarily be used for virtual prototyping of the robotic mink-feeding systems and has been used for contributions 7 and 9.
	
	\section{PhD Project Hypothesis Validation}
	\label{sec:hypothesis}
	
	During the course of this PhD project, the different aspects of the hypothesis presented in 
	Section~\ref{sec:research_objective} have been addressed using the contributions from the project. 
	The hypothesis is comprised of two parts that have been developed into the evaluation criteria presented in Section~\ref{sec:evalcrit}. 
	For completeness, the PhD project hypothesis is restated below:
	
	\begin{itemize}
		\item \textbf{Collaborative models can support multidisciplinary collaboration and system development.}
		\item \textbf{A collaborative model of a robotic or automated agricultural ground-vehicle can be utilised to explore alternative design configurations.}
	\end{itemize}
	The first part of the hypothesis is covered by the contributions illustrated in Figure~\ref{fig:hypo1_eval}, 
	using the evaluation criteria from Subfigures~\ref{fig:multidisciplinary},~\ref{fig:contrib:modelling}, and~\ref{fig:deployment}. 
	Together, the evaluation results illustrated in Figure~\ref{fig:hypo1_eval} are sufficient to validate the first part of the PhD project hypothesis.
	\begin{figure}[hbt]
		\centering
%

\usetikzlibrary{shapes}
\usetikzlibrary{positioning,fit,calc}

\ifdefined\cref\else\def\cref#1{[C1]}\fi
\newcommand{\D}{9} 
\newcommand{\U}{4} 

\newdimen\R 
\R=3.5cm 
\newdimen\L 
\L=4cm

\newcommand{\A}{-360/\D} 
\newcommand{\AO}{\A + 120} 
 
\def\labelformat#1{\tiny #1}

\def\makeweb{
\tikzset{spiderwebpath/.style={opacity=0.2}}
\tikzset{measurepath/.style={ultra thick,opacity=0.5,rounded corners=1pt}}
\tikzset{mcol1/.style={color={rgb:red,237;green,45;blue,46}}}
\tikzset{mcol2/.style={color={rgb:red,0;green,140;blue,71}}}
\tikzset{mcol3/.style={color={rgb:red,24;green,89;blue,169}}}
\tikzset{mcol4/.style={color={rgb:red,102;green,44;blue,145}}}
\tikzset{mcol5/.style={color=olive}}
  \path (0:0cm) coordinate (O); 

  \foreach \X in {1,...,\D}{
    \draw[spiderwebpath] (\X*\A:0) -- (\X*\A:\R);
  }

  \foreach \Y in {0,...,\U}{
    \foreach \X in {1,...,\D}{
      \path (\X*\AO :\Y*\R/\U) coordinate (D\X-\Y);
      \fill[spiderwebpath] (D\X-\Y) circle (4pt);
    }

    \draw [spiderwebpath] (0:\Y*\R/\U) \foreach \X in {1,...,\D}{
        -- (\X*\A :\Y*\R/\U)
    } -- cycle;
  }
     \fill[white] (0,0) circle (4.03pt);
     \fill[spiderwebpath] (0,0) circle (4pt);

  
  \path (1*\AO :\L) node (L1) {\labelformat{ \cref{contribution:methodology}}};
  \path (2*\AO :\L) node (L2) {\labelformat{ \cref{contribution:legosim}}};
  \path (3*\AO :\L) node (L3) {\labelformat{ \cref{contribution:legodeploy}}};
  \path (4*\AO :\L) node (L4) {\labelformat{ \cref{contribution:modelcase2}}};
  \path (5*\AO :\L) node (L5) {\labelformat{ \cref{contribution:feeding}}};
  \path (6*\AO :\L) node (L8) {\labelformat{ \cref{contribution:visualisation}}};
  \path (7*\AO :\L) node (L6) {\labelformat{ \cref{contribution:patent}}};
  \path (8*\AO :\L) node (L9) {\labelformat{ \cref{contribution:DSEvehicle}}};
  \path (9*\AO :\L) node (L7) {\labelformat{ \cref{contribution:DSEANA}}};
}

   \def\labelformat#1{\large #1}
   \def\subfixspiderscale{0.4}

\def\makeContributionCompareWeb{
\begin{subfigure}[t]{0.33\textwidth}
\centering
\begin{tikzpicture}[scale=\subfixspiderscale, transform shape]
\makeweb
  \fill [mcol1,measurepath]
    (D1-4) -- 
    (D2-2) --
    (D3-2) --
    (D4-3) --
    (D5-2) --
    (D6-2) --
    (D7-0) -- 
    (D8-2) -- 
    (D9-2) --  
   cycle;
\end{tikzpicture}
\caption{\scriptsize Multi-disciplinary collaboration support}
\label{fig:multidisciplinary}
\end{subfigure}~
\begin{subfigure}[t]{0.33\textwidth}
\centering
\begin{tikzpicture}[scale=\subfixspiderscale, transform shape]
\makeweb
  \fill [mcol3,measurepath]
    (D1-3) --
    (D2-3) --
    (D3-1) --
    (D4-4) --
    (D5-4) -- 
    (D6-0) --
    (D7-0) -- 
    (D8-0) -- 
    (D9-0) -- 
   cycle;
\end{tikzpicture}
\caption{\scriptsize Support for modelling of different vehicle solutions}
\label{fig:contrib:modelling}
\end{subfigure}
\begin{subfigure}[t]{0.3\textwidth}
\centering
\begin{tikzpicture}[scale=\subfixspiderscale, transform shape]
\makeweb
  \fill [mcol2,measurepath]
    (D1-2) --
    (D2-3) --
    (D3-4) --
    (D4-3) --
    (D5-0) --
    (D6-4) --
    (D7-1) -- 
    (D8-2) -- 
    (D9-0) -- 
    cycle;
\end{tikzpicture}
\caption{\scriptsize Model deployment}
\label{fig:deployment}
\end{subfigure}
\begin{subfigure}[t]{0.3\textwidth}
\centering
\begin{tikzpicture}[scale=\subfixspiderscale, transform shape]
\makeweb
  \fill [mcol4,measurepath]
    (D1-2) --
    (D2-2) --
    (D3-1) --
    (D4-4) --
    (D5-3) -- 
    (D6-4) --
    (D7-2) -- 
    (D8-3) -- 
    (D9-2) -- 
   cycle;
\end{tikzpicture}
\caption{\scriptsize Virtual prototype development support}
\label{fig:contrib:support}
\end{subfigure}
\begin{subfigure}[t]{0.33\textwidth}
\centering
\begin{tikzpicture}[scale=\subfixspiderscale, transform shape]
\makeweb
  \fill [mcol5,measurepath]
    (D1-0) --
    (D2-0) --
    (D3-0) --
    (D4-0) --
    (D5-0) --
    (D6-4) --
    (D7-4) -- 
    (D8-2) -- 
    (D9-3) -- 
   cycle;
\end{tikzpicture}
\caption{\scriptsize Determination of candidate solutions}
\label{fig:contrib:solution}
\end{subfigure}
%
%
%
\begin{subfigure}[t]{0.33\textwidth}
\centering
\begin{tikzpicture}[scale=\subfixspiderscale, transform shape]
\makeweb
 \fill [mcol1,measurepath]
    (D1-4) -- 
    (D2-2) --
    (D3-2) --
    (D4-3) --
    (D5-2) --
    (D6-2) --
    (D7-0) -- 
    (D8-2) -- 
    (D9-2) --  
   cycle;
  \fill [mcol3,measurepath]
    (D1-3) --
    (D2-3) --
    (D3-1) --
    (D4-4) --
    (D5-4) -- 
    (D6-0) --
    (D7-0) -- 
    (D8-0) -- 
    (D9-0) -- 
   cycle;
  \fill [mcol2,measurepath]
    (D1-2) --
    (D2-3) --
    (D3-4) --
    (D4-3) --
    (D5-0) --
    (D6-4) --
    (D7-1) -- 
    (D8-2) -- 
    (D9-0) -- 
    cycle;
  \fill [mcol4,measurepath]
    (D1-2) --
    (D2-2) --
    (D3-1) --
    (D4-4) --
    (D5-3) -- 
    (D6-4) --
    (D7-2) -- 
    (D8-3) -- 
    (D9-2) -- 
   cycle;
  \fill [mcol5,measurepath]
    (D1-0) --
    (D2-0) --
    (D3-0) --
    (D4-0) --
    (D5-0) --
    (D6-4) --
    (D7-4) -- 
    (D8-2) -- 
    (D9-3) -- 
   cycle;
\end{tikzpicture}
\caption{\scriptsize Combined Overview}
\label{fig:contrib:combined-overview}
\end{subfigure}
}

\def\makeHypothesisCompareWeb{
\begin{subfigure}[t]{0.49\textwidth}
\centering
\begin{tikzpicture}[scale=0.6, transform shape]
\makeweb
  \fill [mcol1,measurepath]
    (D1-4) -- 
    (D2-2) --
    (D3-2) --
    (D4-3) --
    (D5-2) --
    (D6-2) --
    (D7-0) -- 
    (D8-2) -- 
    (D9-2) --  
   cycle;
  \fill [mcol3,measurepath]
    (D1-3) --
    (D2-3) --
    (D3-1) --
    (D4-4) --
    (D5-4) -- 
    (D6-0) --
    (D7-0) -- 
    (D8-0) -- 
    (D9-0) -- 
   cycle;
  \fill [mcol2,measurepath]
    (D1-2) --
    (D2-3) --
    (D3-4) --
    (D4-3) --
    (D5-0) --
    (D6-4) --
    (D7-1) -- 
    (D8-2) -- 
    (D9-0) -- 
    cycle;
\end{tikzpicture}
\caption{\scriptsize Hypothesis - part 1}
\label{fig:hypo1_eval}
\end{subfigure}
\begin{subfigure}[t]{0.49\textwidth}
\centering
\begin{tikzpicture}[scale=0.6, transform shape]
\makeweb
    \fill [mcol2,measurepath]
    (D1-2) --
    (D2-3) --
    (D3-4) --
    (D4-3) --
    (D5-0) --
    (D6-4) --
    (D7-1) -- 
    (D8-2) -- 
    (D9-0) -- 
    cycle;
  \fill [mcol4,measurepath]
    (D1-2) --
    (D2-2) --
    (D3-1) --
    (D4-4) --
    (D5-3) -- 
    (D6-4) --
    (D7-2) -- 
    (D8-3) -- 
    (D9-2) -- 
   cycle;
  \fill [mcol5,measurepath]
    (D1-0) --
    (D2-0) --
    (D3-0) --
    (D4-0) --
    (D5-0) --
    (D6-4) --
    (D7-4) -- 
    (D8-2) -- 
    (D9-3) -- 
   cycle;
\end{tikzpicture}
\caption{\scriptsize Hypothesis - part 2}
\label{fig:hypo2_eval}
\end{subfigure}
}
		\makeHypothesisCompareWeb
		\caption{Relation between contributions, evaluation criteria, and PhD project hypothesis.}
		\label{fig:hypothesis_eval}
	\end{figure}
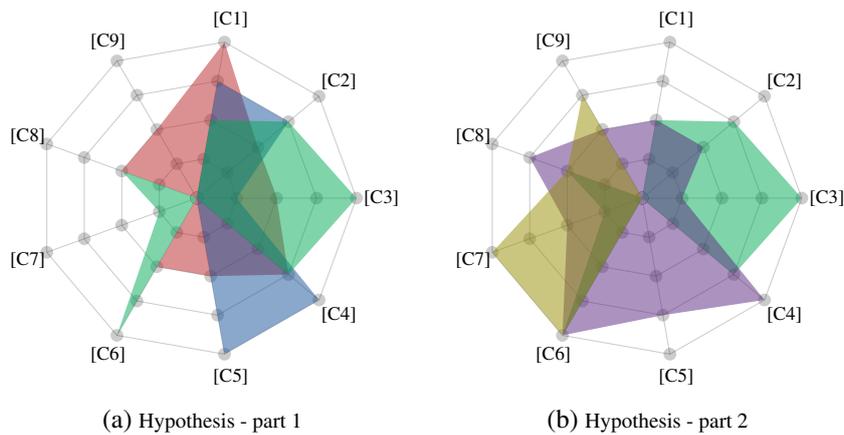
	
	The second part of the PhD project hypothesis is covered by the contributions illustrated in Figure~\ref{fig:hypo2_eval}, 
	using the evaluation criteria from Subfigures~\ref{fig:deployment},~\ref{fig:contrib:support},~and~\ref{fig:contrib:solution}. 
	Coverage of the evaluation criteria based on the project contributions validates the second part of the PhD project hypothesis.
	The conclusion is that the project hypothesis have been fulfilled and overall progress have been made in the research area.
	The results achieved  in this PhD thesis can be used in the future development of automated and robotic agricultural ground-vehicle solutions.

	\section{Future Work}
		\label{sec:future}
		
		This section primarily focuses on a future industrial postdoctoral project in collaboration with the company Conpleks Innovation, and is intended to extend the work conducted in this PhD  thesis. 
		We begin this section by presenting the general focus of the intended postdoctoral project, and describe how the work conducted in this thesis can be incorporated in synergistic manner . 
		We then describe how base co-models  can be developed for additional vehicle types, and how they can be converted into an automated model generation process. The section concludes by presenting some future development cases.

		\subsection{Postdoctoral project focus}
		The intention is to use well-known robotic development technologies, such as ROS, Gazebo, and a bullet physics engine, as part of the model-based development of robotic ground-vehicles. 
		In existing robot simulation platforms, such as Gazebo, the developer must create a model of the robotic vehicle in terms of its physical layout, its mechanical solutions, and actuators and sensors. 
		Even the reuse of parts of previous robotic vehicle models requires the developer to have a well-founded understanding of the inner workings of the simulation platforms. 
		
		Future work in the postdoctoral project should establish libraries specifically dedicated to the development of agricultural and other field robots, targeting a “plug-and-play” approach whenever a new robot vehicle solution is to be developed. 
		Since the INTO-CPS project will be using the FMI interface in the research endeavours with co-modelling and co-simulation, 
		it is expected that the FMI interface should the used to implement the development of the aforementioned robotics technologies into a co-modelling and co-simulation environment. 
		This implementation of the FMI will not be part of the postdoctoral research, where the focus will be the development of model-based libraries for agricultural robots.

%
%
		\subsection{Extending the range of base co-models}
		
		The  PhD project has mainly created a basis model vehicle steered using the front or back axle. 
		The  modelling capabilities should be extended to support differential and four-wheeled steering vehicle solutions. 
		Different pulled implement solutions, like sprayer trailers and crop tilling units, are also part of the agricultural operation, 
		and should be part of future co-modelled solutions for providing project developmental support. 
		These new base models would also broaden the support for the extended modelling methodology presented in  Section~\ref{sec:model_methodology} in this thesis. 
		In essence, the base vehicle models should be included as components in these model-based libraries for agricultural robots. 
	
		Automating the model development process, based on these four base vehicle models, would allow production of a specialised modelling solution, 
		where significant parts of the model could be auto-generated, allowing the developer to manually implement only the parts that are specially developed for their specific applications . 
		Over time, this auto-generated library could gradually be enriched by encompassing more case studies that have been developed based on the extended development methodology. 
		Auto-generated models would also allow project developers in other disciplines with little experience in modelling and simulation to develop the ability to create full models, 
		with little or no support from experienced developers in this area, thus achieving the intended “plug-and-play” capabilities.
		

		\subsection{Development cases}
		This subsection provides some directions on future work in the form of examples of future development cases.
		A number of these development cases are intended to be part of a future postdoctoral project. 
		The development cases represent different areas of agricultural operations that would benefit from co-model-based development. 
		These examples are also intended to allow the reader to “get new ideas” on what co-modelling can accomplish in terms of development.


		\subsubsection{Mink-feeding robot ground-vehicle}
		
		The design and co-modelling of solutions for a mink-feeding robot comprises a significant part of the contributions, which has been produced  in this PhD thesis. 
		Making the currently developed FixRobo robot solution by Conpleks Innovation into a completed commercial product 
		that can be shipped to the market would strengthen the extended development methodology presented in this thesis.
		This would prove that this methodology has the potential to support development of automated and robotic agricultural ground-vehicles from the conception of the idea to a final product. 
		Contribution 7 in this thesis that resulted in a patent application submitted last year, still needs to be deployed into a ground-vehicle realisation. 
		The intention here is to deploy this effort in a possible future postdoctoral project.
		

		\subsubsection{Dvorak Spider slope mower}
		
		This machine is already in the market, but it is delivered with a manually operated  remote-control system only that is supplied by another vendor. 
		A joint co-operation with this company is currently being planned by Conpleks Innovation with the expectation for the addition of the required robot functionality. 
		The interesting issue in relation to the suggested postdoctoral project is that the current controller is not sufficiently fault tolerant. 
		\begin{figure}[htbp]
			\centering
				\includegraphics[width=0.42\textwidth]{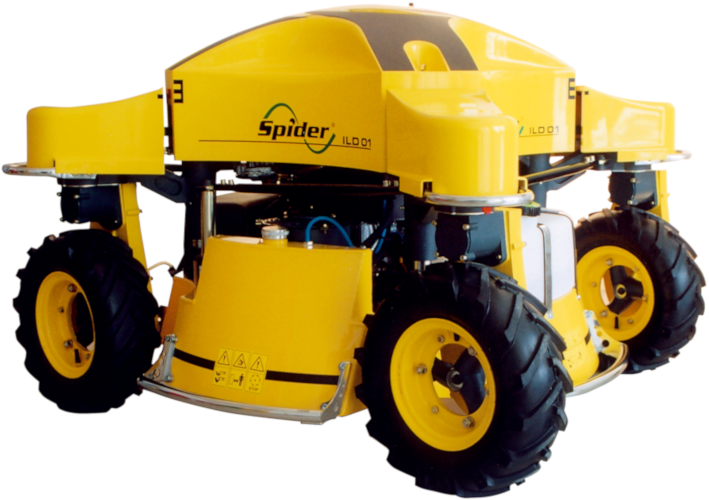}
			\caption{Dvorak Spider slope mower$^{2}$}
			\label{fig:Dvorak_Spider}
		\end{figure}
		Therefore, it is envisaged that the generic strategies for adding fault tolerance can be applied to this case study, and as a consequence, a better product can be produced. 
		In addition, it is envisaged that DSE and adaptive control can be valuable for this case.
		\setcounter{myfootnote}{\value{footnote}}
		\addtocounter{myfootnote}{1}
		\footnotetext[\value{myfootnote}]{Source: Picture of a DVORAK - machine division vehicle solution, DVORAK.}

		\subsubsection{The new Kongskilde Robotti}
		
		This robot will be modelled by the company Kongskilde (who is one of the partners in the INTO-CPS project) as a co-modelled robotic vehicle. 
		The extended development guidelines could be used here to co-model the new robotti vehicle. 
		The task assigned to Conpleks Innovations is to develop the controller that is going to be used for this vehicle solution. 
		Correspondingly, this could constitute part of the postdoctoral work. 
		\begin{figure}[htbp]
			\centering
				\includegraphics[width=0.5\textwidth]{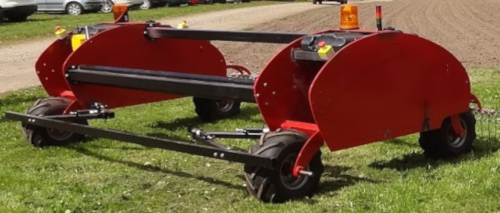}
			\caption{The new Kongskilde Robotti$^{3}$}
			\label{fig:new_robotti}
		\end{figure}
		Herein, DSE is envisaged to be valuable in order to enable the exploration of alternative positioning technologies to determine if 
		cheaper sensor solutions used to combine information on the current field task are able to elicit sufficient precision.
		\addtocounter{myfootnote}{1}
		\footnotetext[\value{myfootnote}]{Source: Picture of a Kongskilde Industries A/S vehicle solution, Kongskilde Industries.}
		

		\subsubsection{Analysing online adaptive path planners}
		
		Automating a mobile vehicle intended to navigate in rows of orchards is a feature relevant for plant nursing and tree cultivation. 
		The ability to navigate reliably is depended upon the vehicle’s capability to know its position and orientation relative to the trees in the orchards. 
		Automated mobile vehicles would need a map of the orchard for navigation. 
		Navigation and localisation in orchard-like environments has mainly been based on the a priori  knowledge on relevant objects in the region.
		
		\begin{figure}[htb]
		  \centering
		  \includegraphics[width=0.75\textwidth]{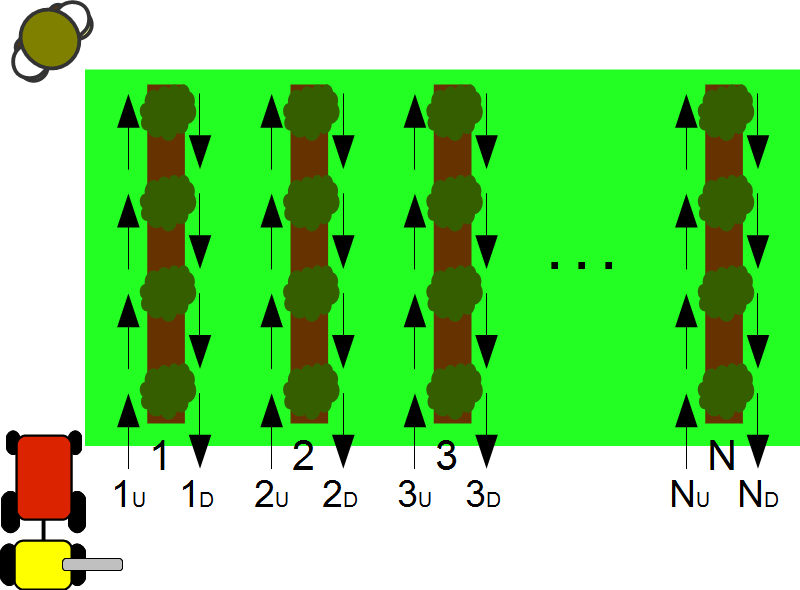}
		  \caption{The robot vehicle intended to perform plant nursing. 
		  The schematic shows the N rows of orchard trees and intermittent lanes where the robot must move up and down.
		   Shown also, is the pedestrian worker working in the field that the robot must avoid. }
	 	\label{fig:ochardplan}
		\end{figure}
		
		Utilising prior knowledge about the structure of an orchard does not account for changes in the existence of objects and their placement. 
		The operations of human workers in the same environment could be in conflict with robot operations, thereby creating situations where robot operations are halted or become hazardous. 
		Adapting online the planned route that the robotic vehicle must follow in order to account for any human worker in the field would ensure continuous and uninterrupted operation of the robot. 
		Co-modelling and co-simulation could be used here to create these hazardous scenarios and test different strategies to handle the problems online as they arise. 
		This type of hazardous scenario is not only related to orchards, and a similar case can be found for mink farms where multiple farm houses are placed in rows.
		
%
		
		\subsubsection{Tractor-implement solution for crop cleaning}
		
\begin{figure}[htb]%
			\begin{subfigure}[t]{0.49\textwidth}
				\includegraphics[width=\textwidth]{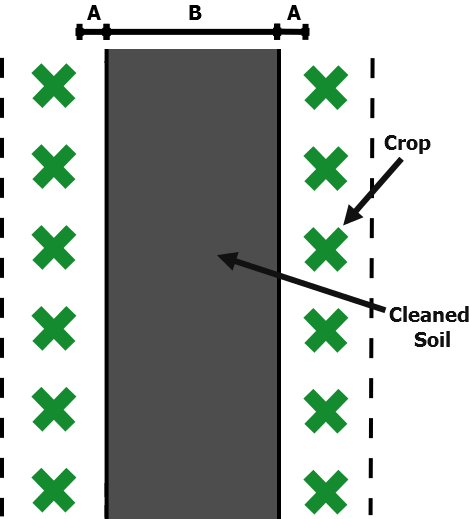}
				\caption{Conventional row-crop cleaning with a safety distance $A$}
				\label{fig:conventionalcropcleaning}
			\end{subfigure}\hfill
			\vline\hfill
			\begin{subfigure}[t]{0.459\textwidth}
				\includegraphics[width=\textwidth]{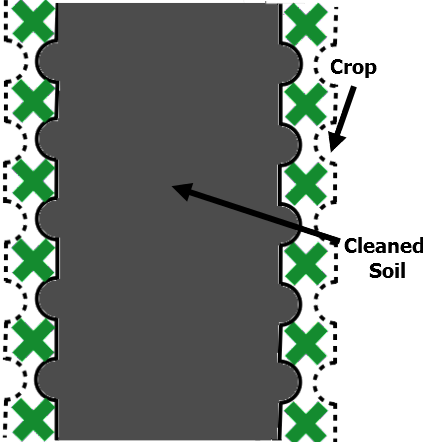}
				\caption{Intended intercrop cleaning method based on the Kongskilde implementation.}
				\label{fig:motivationtinesideshift}
			\end{subfigure}
		\end{figure}

		Current weed control products on the commercial market focuses on weed removal between the crop rows using a solution that is entirely mechanical. 
		This leaves a significant amount of weeds inside the rows between the crops, since a safety distance is chosen for crops to prevent the product from harming the crop~(see Figure~\ref{fig:conventionalcropcleaning}). 
		If the crop could be cleaned as illustrated in Figure~\ref{fig:motivationtinesideshift}, or in a similar manner, a higher degree of removal can be achieved.
		
\begin{figure}[htb]
			\centering
				\includegraphics[width=0.55\textwidth]{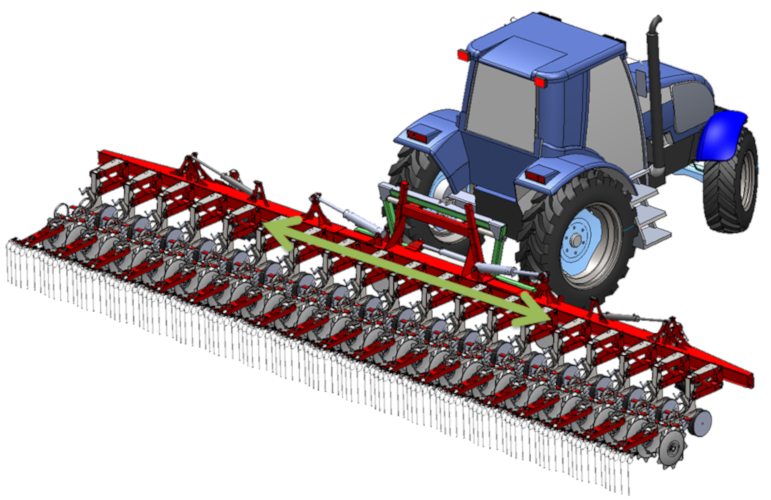}
				\caption{Envisioned implement solution from the KongsKilde innovation group.
				Intended boom side shifting operation is also shown'}
			\label{fig:BoomSideShift}
		\end{figure}

		At this present time, this system is being implemented as a new, stand-alone, row crop cleaner system in Kongskilde Industries using camera input and vision recognition. 
		The specific product, the tractor, and the direction of the boom side shifting, are shown in Figure~\ref{fig:BoomSideShift}.
		The side shifting of the boom is used to make overall corrections on the position of the boom, with respect to the position of the crops in the field and the position of the tractor. 
		This compensation is needed since differences between GPS logged tracks and the actual crop positions exist and vary.
		Co-modelling, co-simulation, and DSE, could be used here to explore the different field scenarios and weed removal strategies for automated operation.
		

\bibliographystyle{plain}
\bibliography{dan}

\part{Publications}\label{part2}

\ifdefined\includePublications\def\isaspublished{The content of this chapter is as published.}\else\def\isaspublished{The content of this chapter is as published.}
\fi
\def\framedpublications#1{
\begin{framed}\vspace{-.8em}
\begin{itemize}
\item[\cite{#1}] \bibentry{#1}
\end{itemize}\vspace{-.8em}
\end{framed}}
\setcounter{chapter}{0}

\chapter[Towards a Methodology for Modelling and Validation]{TOWARDS A METHODOLOGY FOR MODELLING AND VALIDATION OF AN AGRICULTURAL VEHICLES DYNAMICS AND CONTROL}
The paper presented in this chapter is a peer-reviewed conference paper and has been presented at IMAACA 2012.

\framedpublications{Christiansen&12a}

\includepdf[pages=-,scale=0.658,templatesize={137mm}{260mm},noautoscale=true,offset=0 0,pagecommand={\thispagestyle{headings}}]{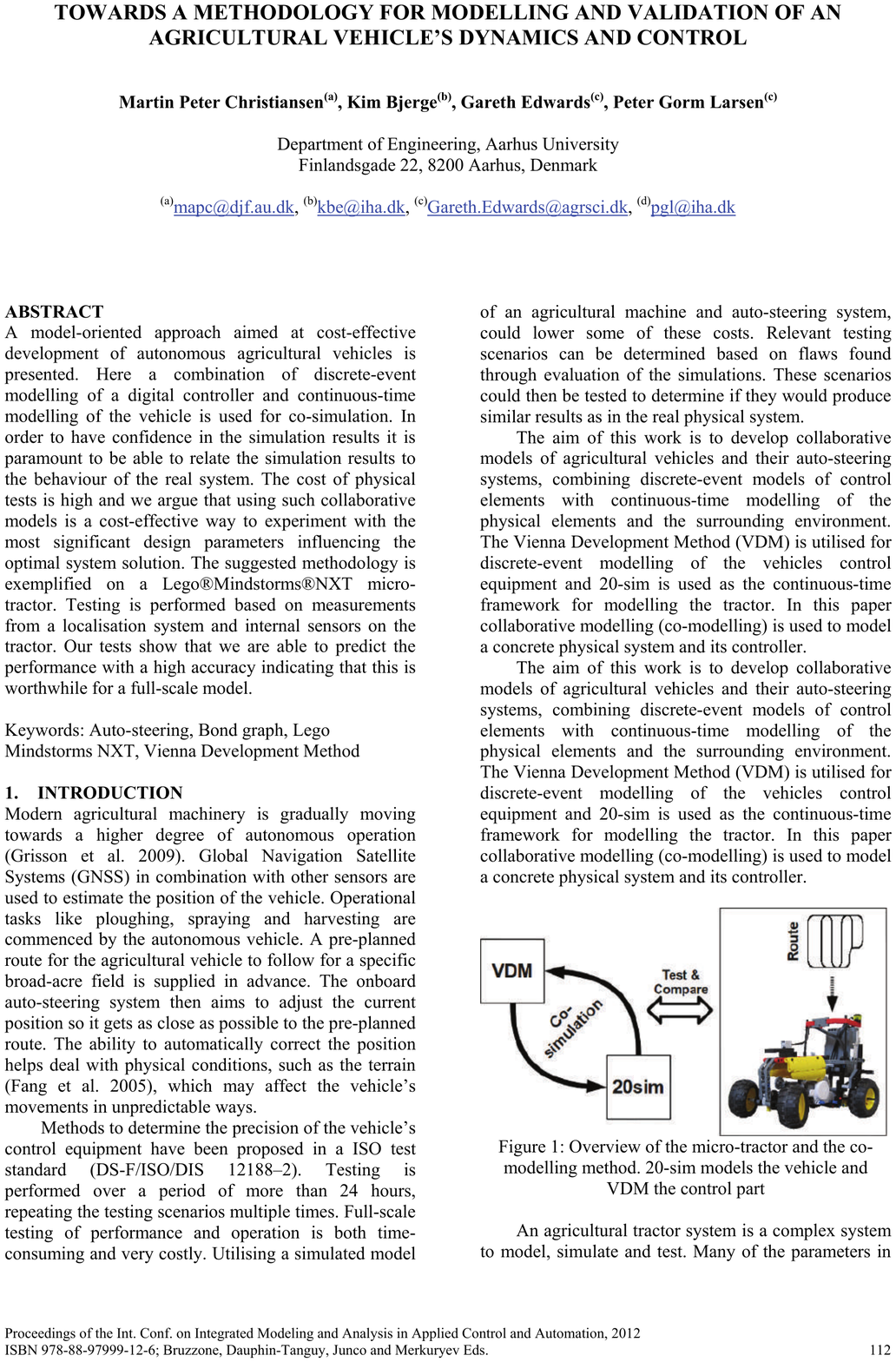}

\chapter[Adaptive Controller Settings for a Load-carrying Vehicle]{Collaborative Model Based Development of Adaptive Controller Settings for a Load-carrying Vehicle with Changing Loads}
The paper presented in this chapter has been presented at CIOSTA (Commission Internationale de l’Organisation Scientifique du Travail en Agriculture).

\framedpublications{Christiansen&13a}

\includepdf[pages=-,scale=0.68,templatesize={137mm}{260mm},noautoscale=true,offset=0 0,pagecommand={\thispagestyle{headings}}]{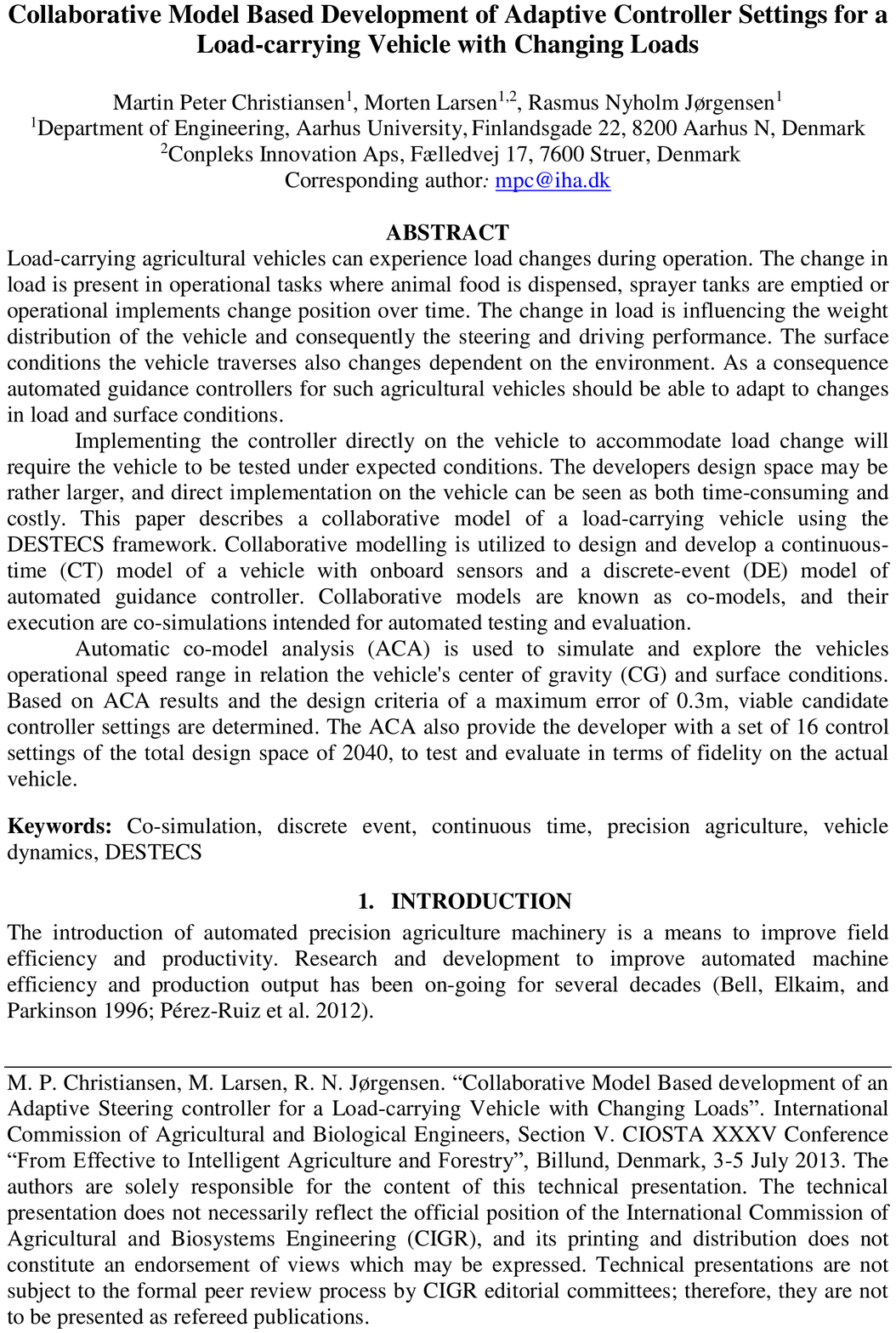}

\chapter[Planned Field Operations using LEGO Mindstorms NXT]{A Test Platform for Planned Field Operations Using LEGO Mindstorms NXT}
The paper presented in this chapter has been published in MDPI Robotics).

\framedpublications{Edwards&13}

\includepdf[pages=-,scale=0.658,templatesize={137mm}{260mm},noautoscale=true,offset=0 0,pagecommand={\thispagestyle{headings}}]{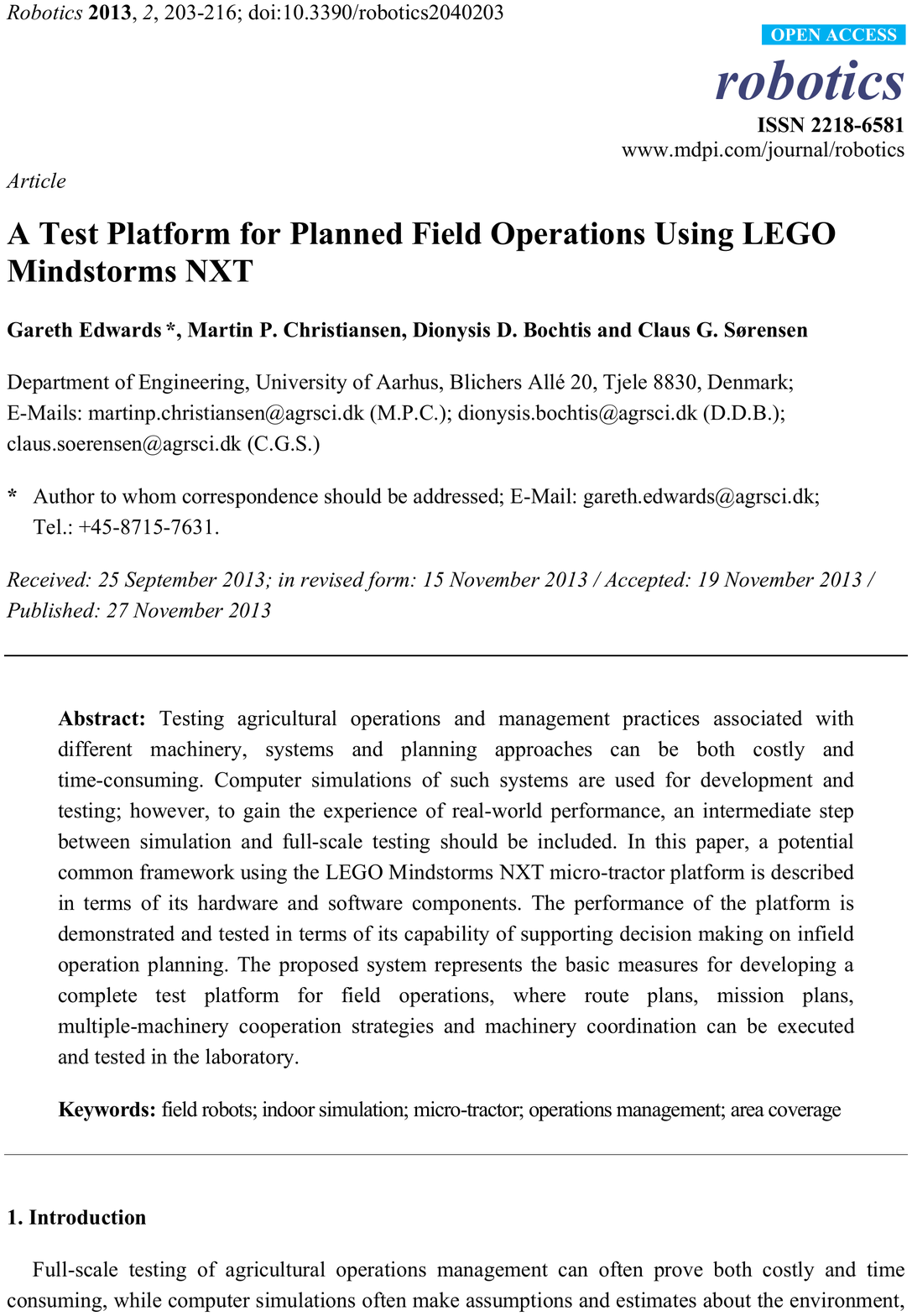}

\chapter[Patent Application]{Method for recording and predicting position data for a self-propelled 
wheeled vehicle, and delivery or pick up system comprising a selfpropelled, self-guided wheeled vehicle}
This patent application was submitted to the Danish Patent and Trademark Office 19 of December 2014).

\framedpublications{Christiansen&14c}

\includepdf[pages=-,scale=0.645,templatesize={137mm}{260mm},noautoscale=true,offset=0 0,pagecommand={\thispagestyle{headings}}]{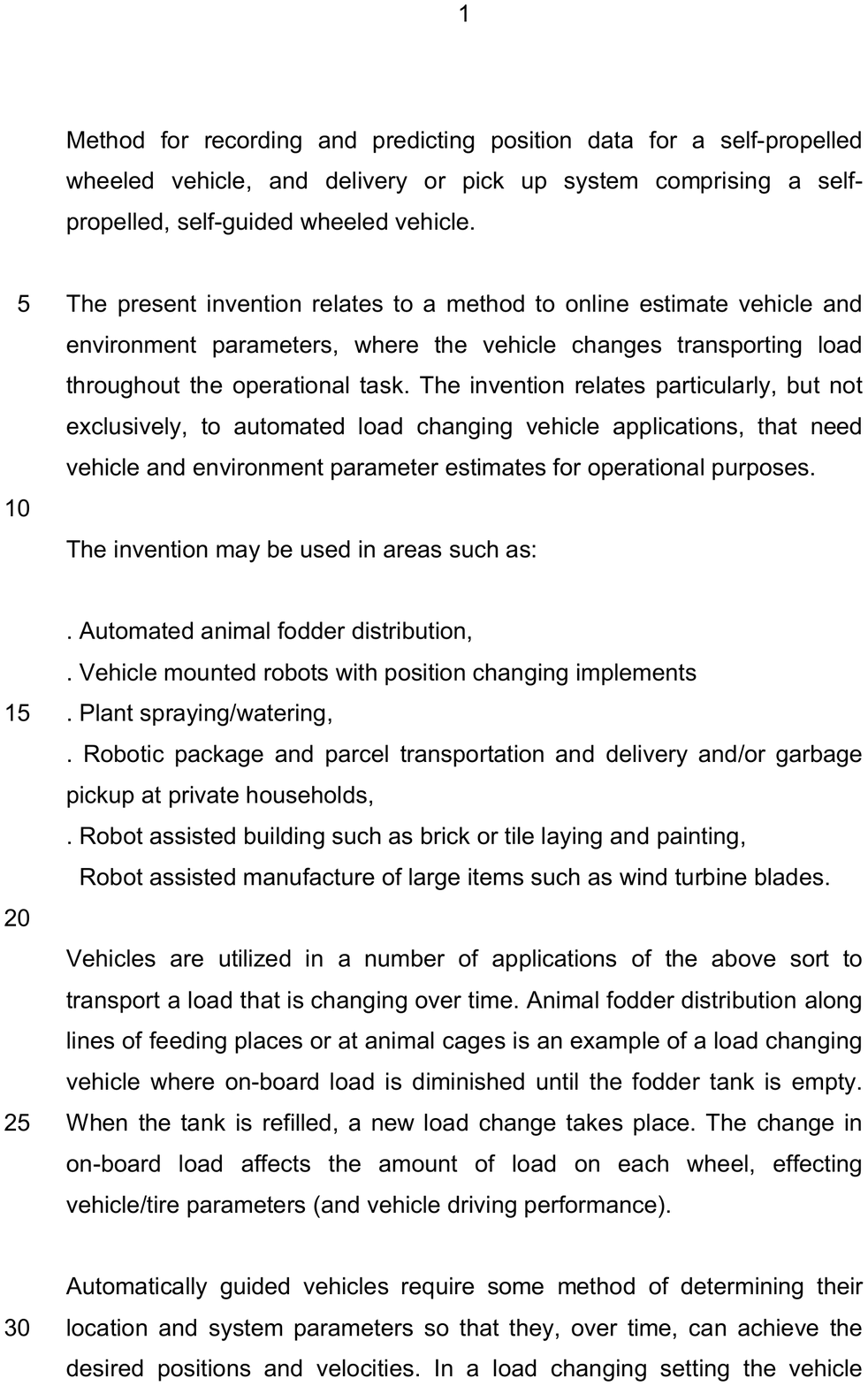}

\chapter[Candidate Overview using Co-model Driven Development]{Agricultural Robotic Candidate Overview using Co-model Driven Development}
The paper presented in this chapter has been submitted to IROS (International Conference on Intelligent Robots and Systems)

\framedpublications{Christiansen&15a}


\includepdf[pages=-,scale=0.73,templatesize={137mm}{260mm},noautoscale=true,offset=0 0,pagecommand={\thispagestyle{headings}}]{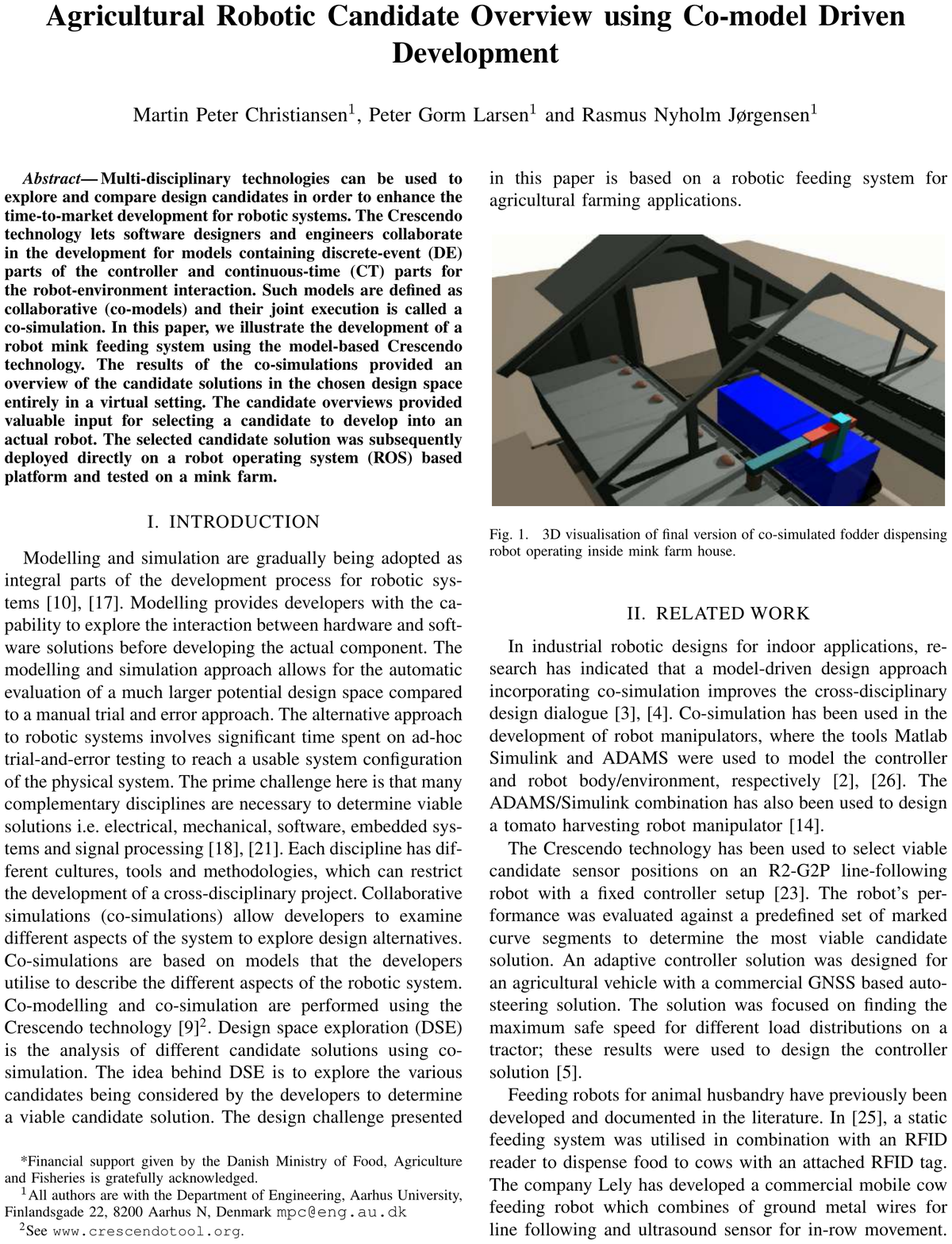}

\chapter[Multidisciplinary Robotic Design Space Exploration]{Robotic design choice overview using Co-simulation and Design Space Exploration}
The paper presented in this chapter has been submitted to MDPI Robotics

\framedpublications{Christiansen&15b}


\includepdf[pages=-,scale=0.625,templatesize={137mm}{260mm},noautoscale=true,offset=0 0,pagecommand={\thispagestyle{headings}}]{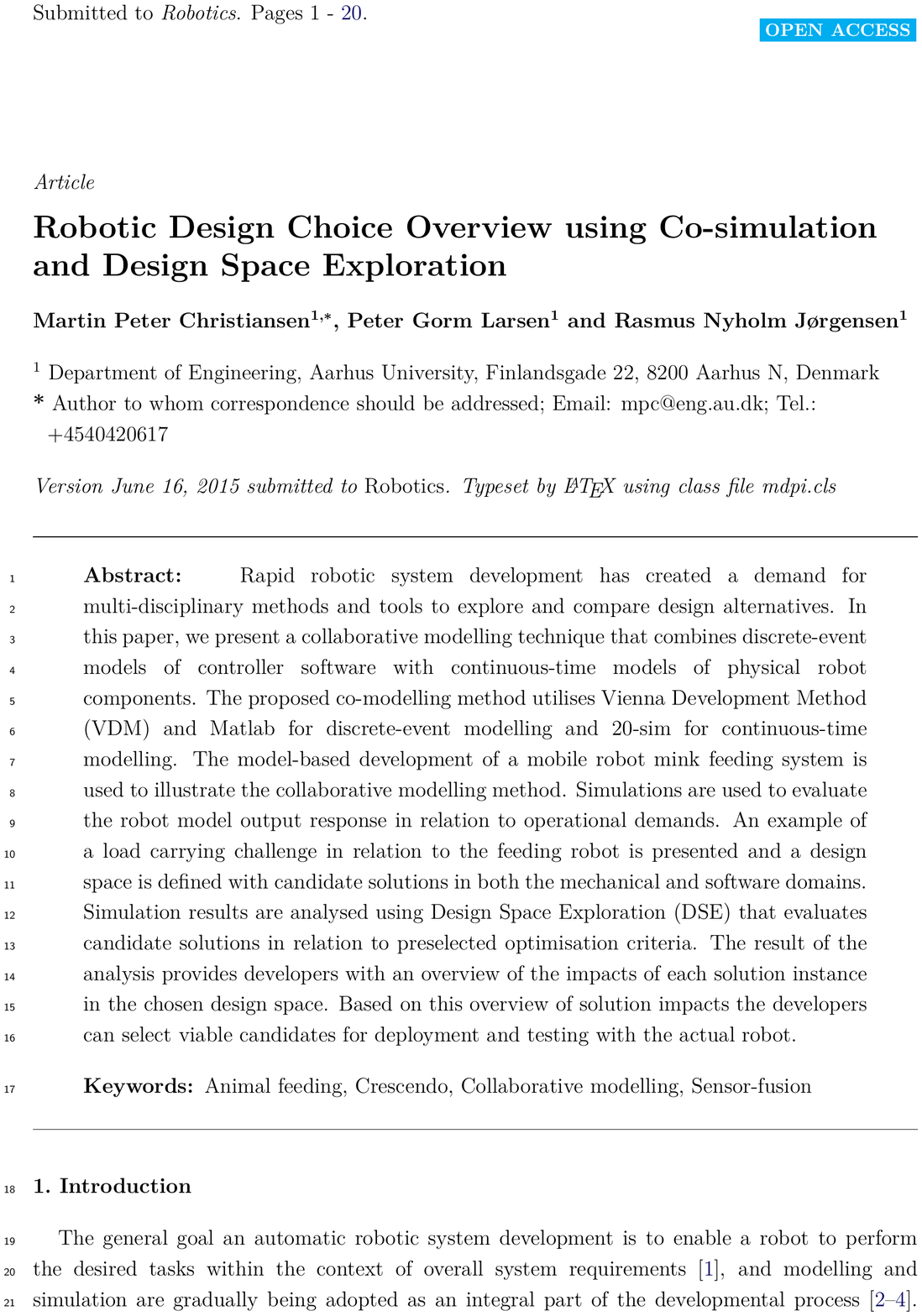}

\end{document}